\documentclass[10pt,journal,compsoc]{IEEEtran}

\ifCLASSOPTIONcompsoc
  \usepackage[nocompress]{cite}
\else
  \usepackage{cite}
\fi
\ifCLASSINFOpdf
\else
\fi
\usepackage{epsfig}
\usepackage{graphicx}
\usepackage{amsmath}
\usepackage{amssymb}
\usepackage{enumitem}
\usepackage{tablefootnote}
\usepackage{multirow}
\usepackage{capt-of}
\usepackage{fixltx2e}
\usepackage{floatrow}
\usepackage{pifont}
\newcommand{\cmark}{\text{\ding{51}}}
\newcommand{\xmark}{\text{\ding{55}}}

\usepackage[sort, numbers]{natbib}
\usepackage{enumitem}
\usepackage{tablefootnote}
\usepackage{multirow}
\usepackage{capt-of}
\usepackage{fixltx2e}
\usepackage{floatrow}
\usepackage[pagebackref=true,breaklinks=true,letterpaper=true,colorlinks,bookmarks=false]{hyperref}
\usepackage[table,xcdraw]{xcolor}
\usepackage[normalem]{ulem}
\useunder{\uline}{\ul}{}

\newcommand{\x}{\times} 
\newcommand{\etc}{\textit{etc}} 
\newcommand{\ie}{\textit{i.e.,}} 
\newcommand{\etal}{\textit{et al.}}

\newcommand{\sub}{\textsubscript} 
\newcommand{\UCF}{UCF\textunderscore CROWD\textunderscore 50 }
\newcommand{\UREB}{\textit{U-REB}} 
\newcommand{\UREBC}{\textit{U-REBC}} 

\newcommand{\Yh}{\hat{Y}} 

\begin{document}
%
\title{JHU-CROWD++: Large-Scale Crowd Counting Dataset and A Benchmark Method}
%
%
%
%

\author{Vishwanath A. Sindagi,~\IEEEmembership{Student Member,~IEEE,}
        Rajeev Yasarla,~\IEEEmembership{Student Member,~IEEE,}
        and~Vishal M. Patel,~\IEEEmembership{Senior~Member,~IEEE}
\IEEEcompsocitemizethanks{\IEEEcompsocthanksitem V. A. Sindagi, R. Yasarla and V. M. Patel are with the Department of Electrical and Computer Engineering, Johns Hopkins University, Baltimore, MD, 21218.\protect\\
E-mail: (vishwanathsindagi@jhu.edu, ryasarla19@jhu.edu, vpatel36@jhu.edu)
}
\thanks{Manuscript received April 19, 2005; revised August 26, 2015.}}

%
%

\markboth{Journal of \LaTeX\ Class Files,~Vol.~14, No.~8, August~2015}%
{Shell \MakeLowercase{\textit{et al.}}: Bare Advanced Demo of IEEEtran.cls for IEEE Computer Society Journals}
%



\IEEEtitleabstractindextext{%
\begin{abstract}
	

We introduce a new large scale  unconstrained crowd counting dataset (JHU-CROWD++) that contains  ``4,372" images with ``1.51 million" annotations. In comparison to existing datasets, the proposed dataset is collected under a variety of  diverse scenarios and environmental conditions. Specifically, the dataset includes several images with weather-based degradations and illumination variations, making it a very challenging dataset. Additionally, the dataset consists of a rich set of  annotations at both image-level and head-level.  Several recent methods are evaluated and compared on this dataset. The dataset can be downloaded from \href{http://www.crowd-counting.com}{\textit{http://www.crowd-counting.com}}.

Furthermore, we propose a novel crowd counting network that progressively generates crowd density maps via residual error estimation. The proposed method uses VGG16 as the backbone network and employs density map generated by the final layer as a coarse prediction to refine and generate finer density maps in a progressive fashion using residual learning. Additionally, the residual learning is guided by an uncertainty-based confidence weighting mechanism that permits the flow of only high-confidence residuals in the refinement path. The proposed Confidence Guided Deep Residual Counting Network (CG-DRCN) is evaluated on recent complex datasets, and it achieves significant improvements in errors. 
\end{abstract}

}

\maketitle

\IEEEdisplaynontitleabstractindextext

%
\IEEEpeerreviewmaketitle

\ifCLASSOPTIONcompsoc
\IEEEraisesectionheading{\section{Introduction}\label{sec:introduction}}
\else
\section{Introduction}
\label{sec:introduction}
\fi

%
%
%
\IEEEPARstart{W}{ith} burgeoning population and rapid urbanization, crowd gatherings have become more prominent in the recent years. Consequently, computer vision-based crowd analytics and surveillance \cite{li2015crowded,zhan2008crowd,idrees2013multi,zhang2015cross,zhang2016single,sindagi2017generating,sam2017switching,chan2008privacy,rodriguez2011density,zhu2014crowd,li2014anomaly,mahadevan2010anomaly, marsden2018people,sindagi2019dafe,sam2019almost,chan2009bayesian,chen2013cumulative}  have received increased interest. Furthermore, algorithms developed for the purpose of crowd analytics have found applications in other fields such as    agriculture monitoring  \cite{lu2017tasselnet}, microscopic biology \cite{lempitsky2010learning}, urban planning and environmental survey \cite{french2015convolutional,zhan2008crowd}. Current state-of-the-art counting networks achieve impressive error rates on a variety of datasets that contain numerous challenges. Their success can be broadly attributed to two major  factors: (i) development and publication of challenging datasets \cite{idrees2013multi,zhang2015cross,zhang2016single,idrees2018composition}, and  (ii) design of novel convolutional neural network (CNN) architectures specifically for improving count performance \cite{zhang2015cross, walach2016learning, onoro2016towards, sam2017switching, sindagi2017cnnbased, ranjan2018iterative, cao2018scale,sam2018top}. In this paper, we consider both of the above factors  in an attempt to further improve the crowd counting performance.

First, we identify the next set of challenges that require attention from the crowd counting research community and collect a large-scale dataset collected under a variety of conditions.  Existing efforts like \UCF \cite{idrees2013multi}, World Expo '10 \cite{zhang2015cross} and ShanghaiTech \cite{zhang2016data} have progressively increased the complexity of the datasets in terms of average count per image, image diversity \etc. While these datasets have enabled rapid progress in the counting task, they suffer from  shortcomings such as limited number of training samples, limited diversity in terms of environmental conditions, dataset bias in terms of positive samples, and limited set of annotations. Idrees \etal \cite{idrees2018composition} proposed a new dataset called UCF-QNRF that alleviates some of these challenges. Most recently, Wang \etal \cite{wang2020nwpu} released a large-scale crowd counting dataset consisting of 5,109 images with ~2.13 million annotations. Specifically, the images are collected under a variety of illumination conditions. Nevertheless, they do not specifically consider some of the challenges such as adverse environmental conditions, dataset bias and limited annotation data\footnote{Existing datasets provide only point-wise annotations.}.  

\begin{figure*}[htp!]
	\begin{center}
		\includegraphics[width=0.19\linewidth, height=0.13\linewidth]{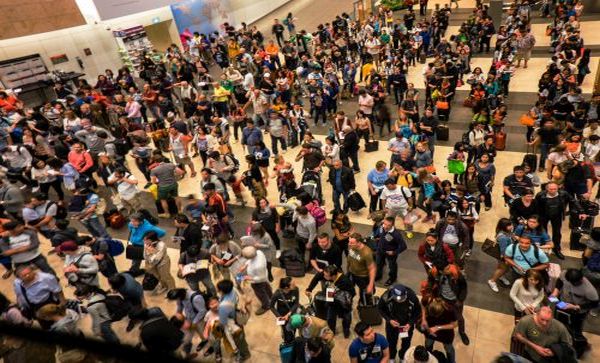}
		\includegraphics[width=0.19\linewidth, height=0.13\linewidth]{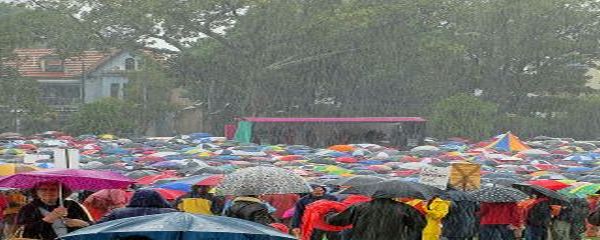}
		\includegraphics[width=0.19\linewidth, height=0.13\linewidth]{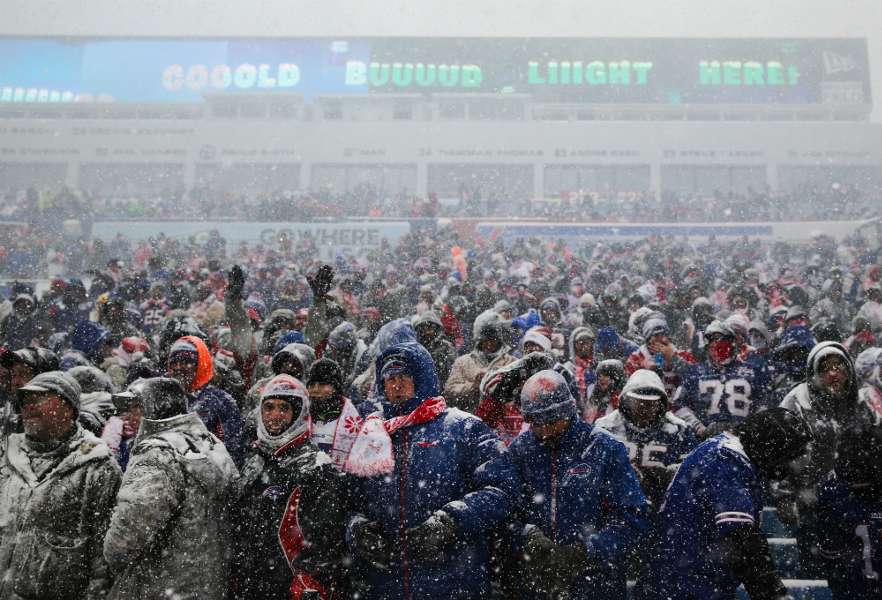}
		\includegraphics[width=0.19\linewidth, height=0.13\linewidth]{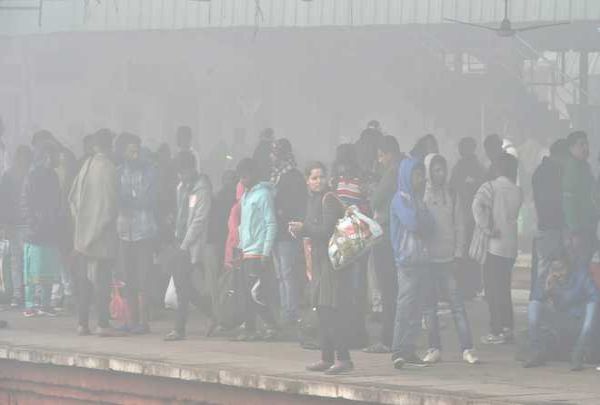}
		\includegraphics[width=0.19\linewidth, height=0.13\linewidth]{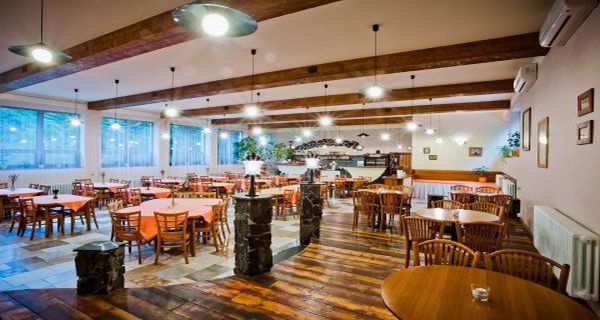}
		
		\includegraphics[width=0.19\linewidth, height=0.13\linewidth]{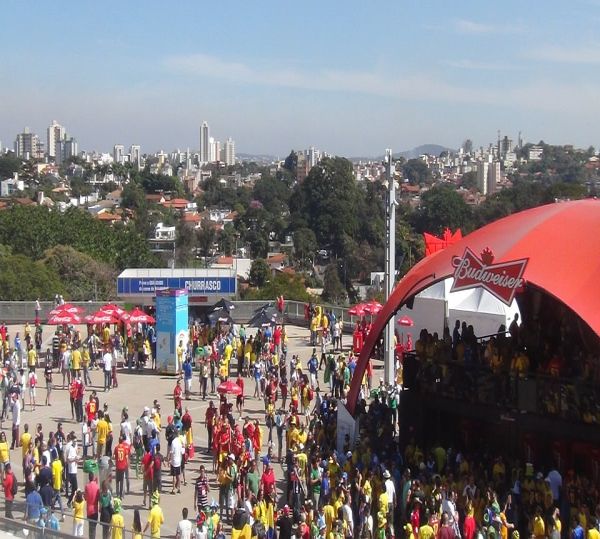}
		\includegraphics[width=0.19\linewidth, height=0.13\linewidth]{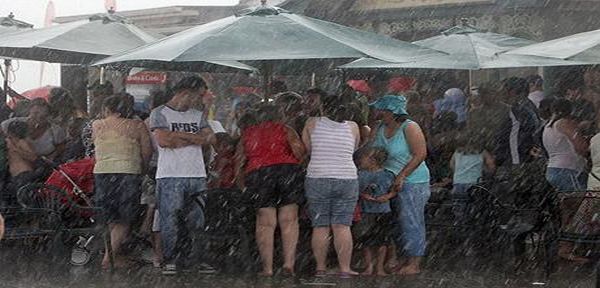}
		\includegraphics[width=0.19\linewidth, height=0.13\linewidth]{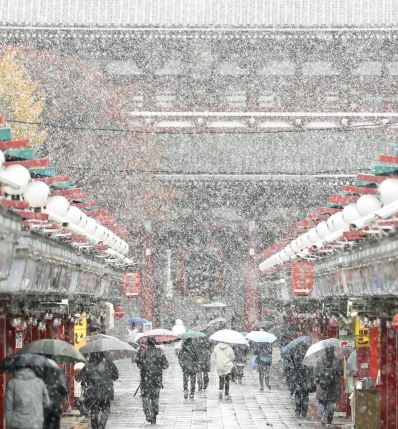}
		\includegraphics[width=0.19\linewidth, height=0.13\linewidth]{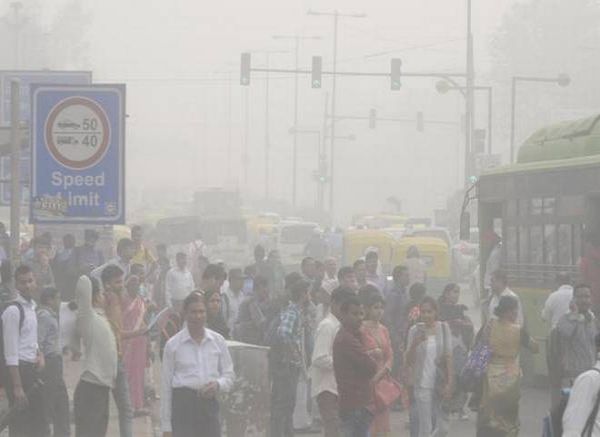}
		\includegraphics[width=0.19\linewidth, height=0.13\linewidth]{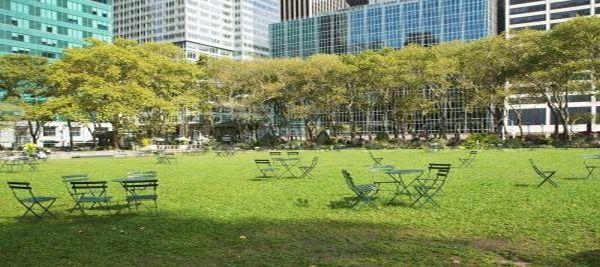}\\
		\includegraphics[width=0.19\linewidth, height=0.13\linewidth]{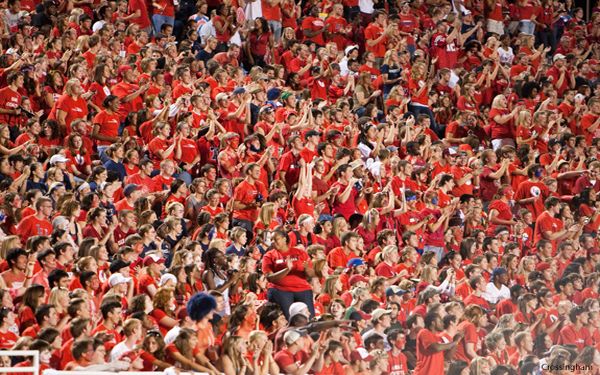}
		\includegraphics[width=0.19\linewidth, height=0.13\linewidth]{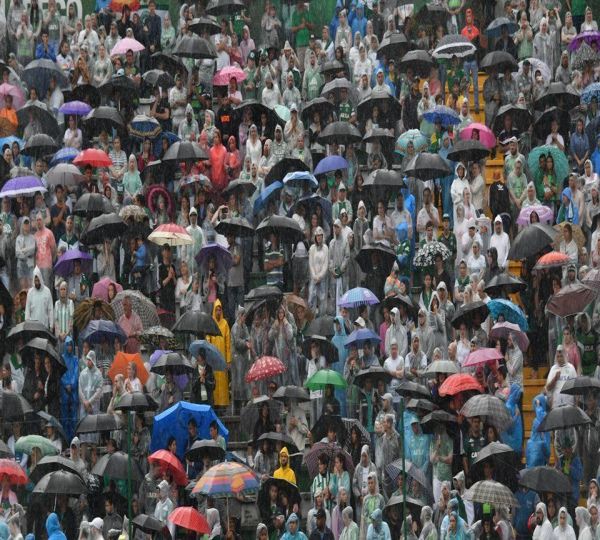}
		\includegraphics[width=0.19\linewidth, height=0.13\linewidth]{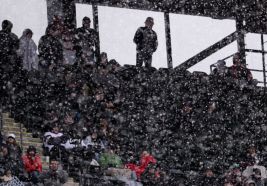}
		\includegraphics[width=0.19\linewidth, height=0.13\linewidth]{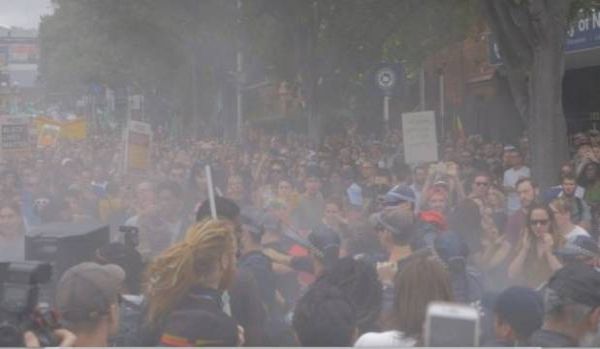}
		\includegraphics[width=0.19\linewidth, height=0.13\linewidth]{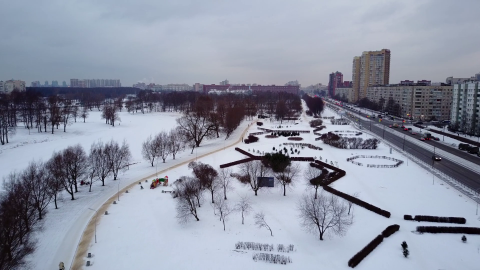}\\
		\includegraphics[width=0.19\linewidth, height=0.13\linewidth]{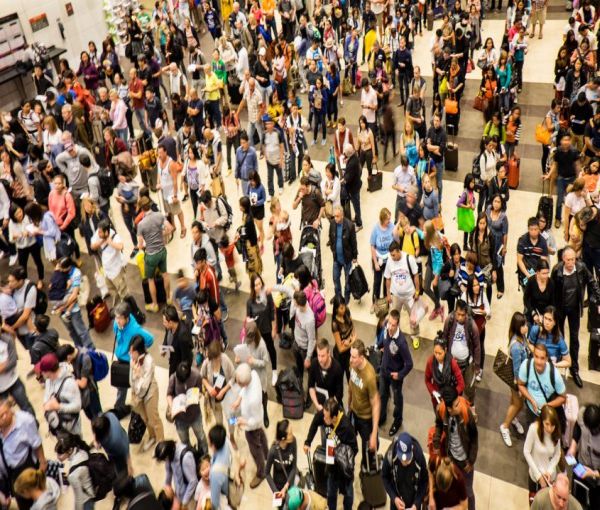}
		\includegraphics[width=0.19\linewidth, height=0.13\linewidth]{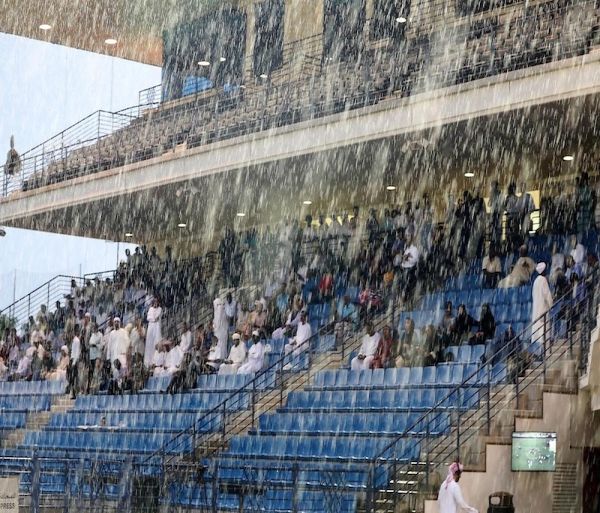}
		\includegraphics[width=0.19\linewidth, height=0.13\linewidth]{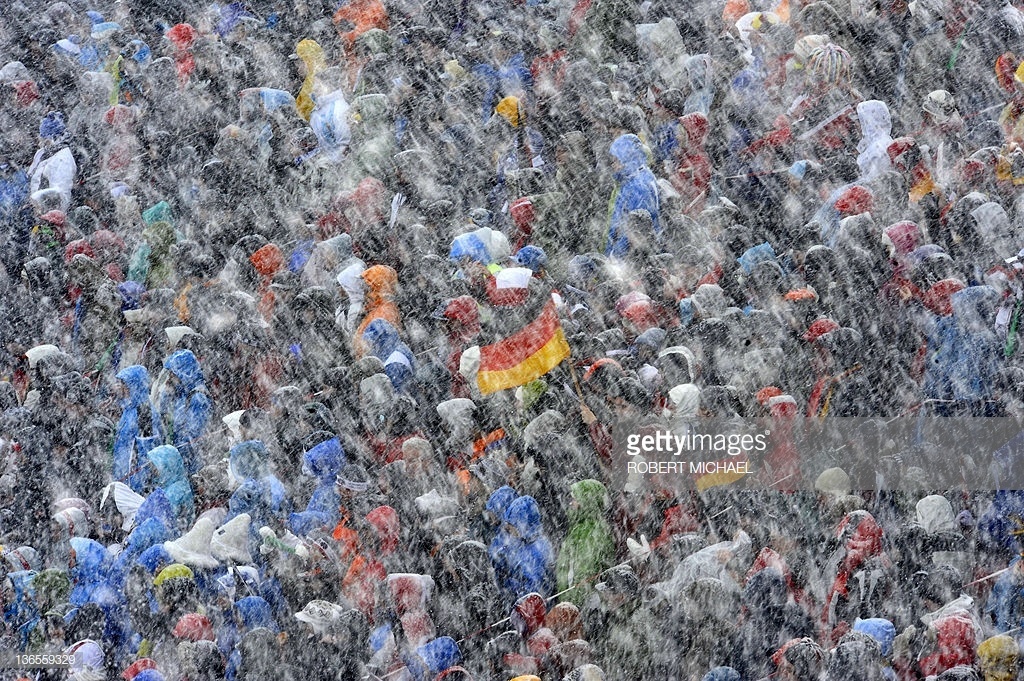}
		\includegraphics[width=0.19\linewidth, height=0.13\linewidth]{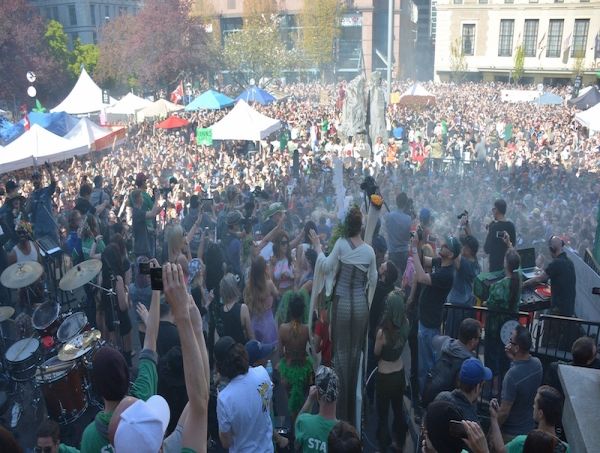}
		\includegraphics[width=0.19\linewidth, height=0.13\linewidth]{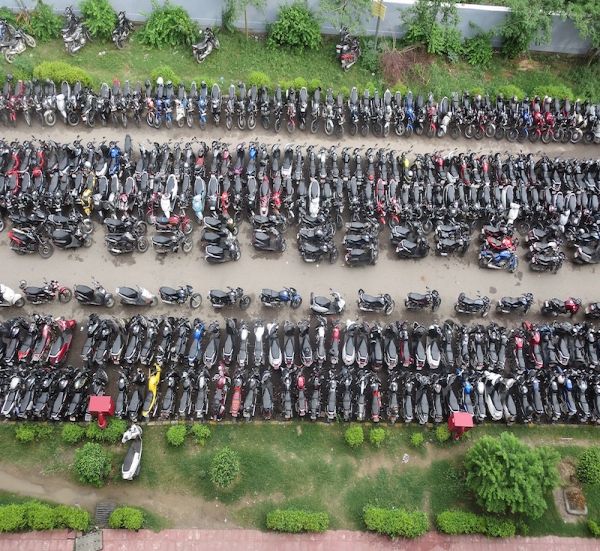}\\	
		\includegraphics[width=0.19\linewidth, height=0.13\linewidth]{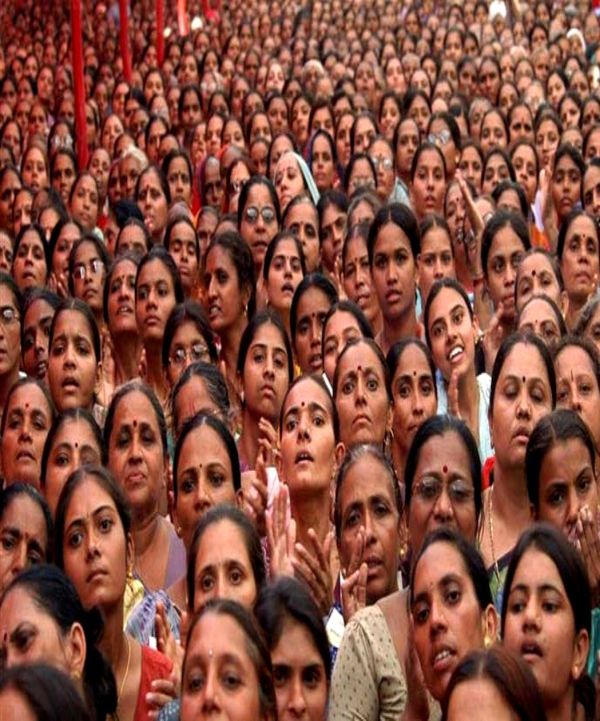}
		\includegraphics[width=0.19\linewidth, height=0.13\linewidth]{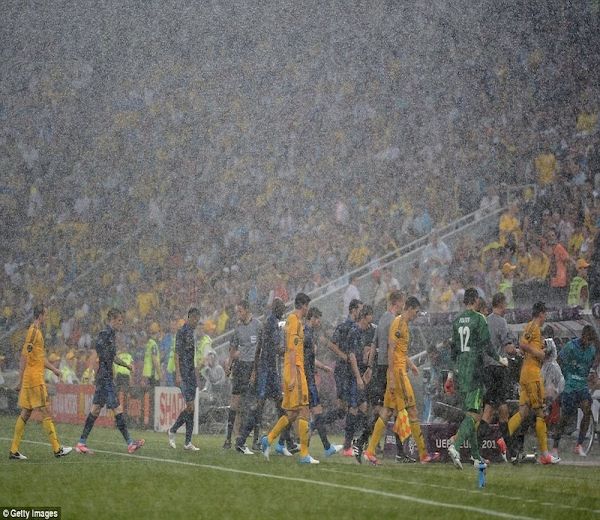}
		\includegraphics[width=0.19\linewidth, height=0.13\linewidth]{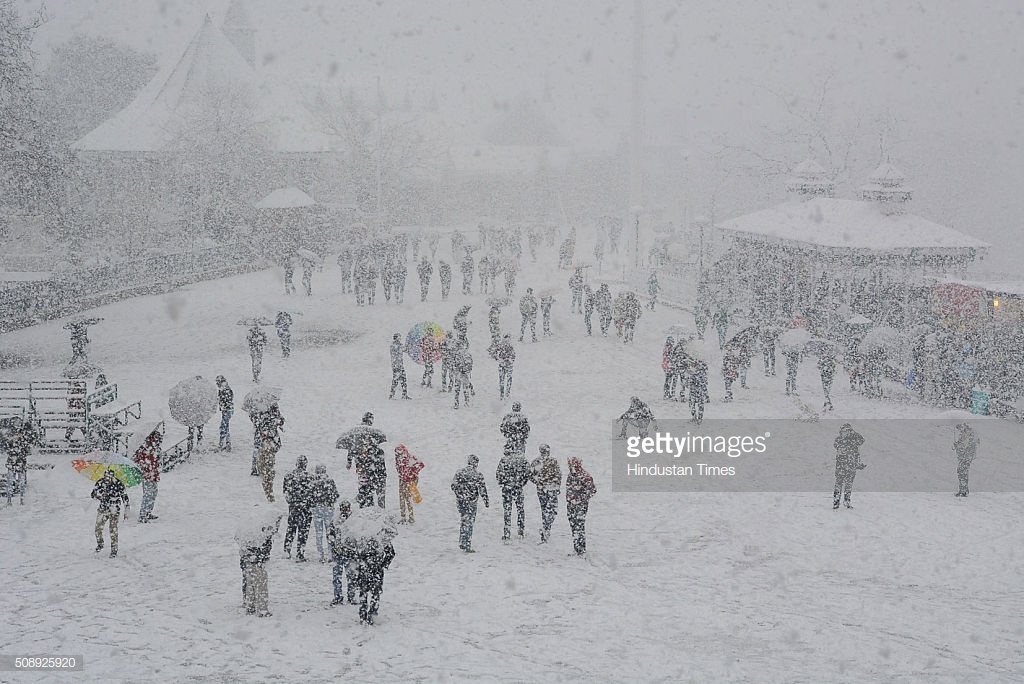}
		\includegraphics[width=0.19\linewidth, height=0.13\linewidth]{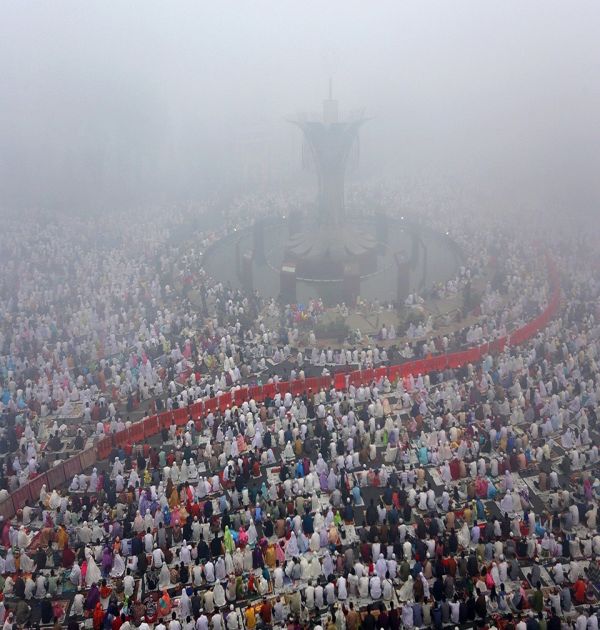}
		\includegraphics[width=0.19\linewidth, height=0.13\linewidth]{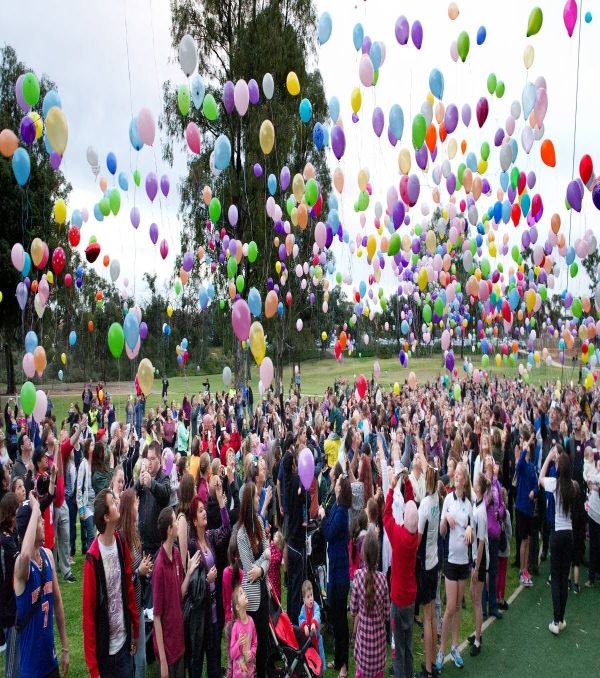}\\	
				\includegraphics[width=0.19\linewidth, height=0.13\linewidth]{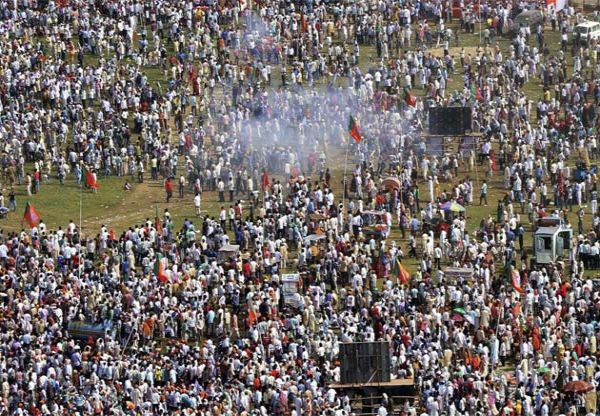}
		\includegraphics[width=0.19\linewidth, height=0.13\linewidth]{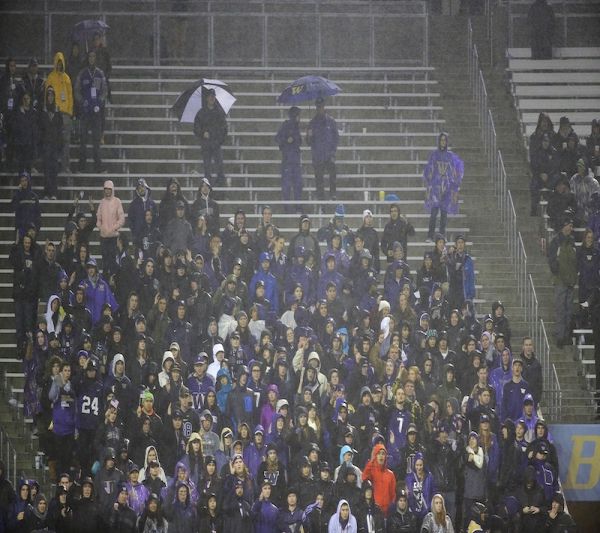}
		\includegraphics[width=0.19\linewidth, height=0.13\linewidth]{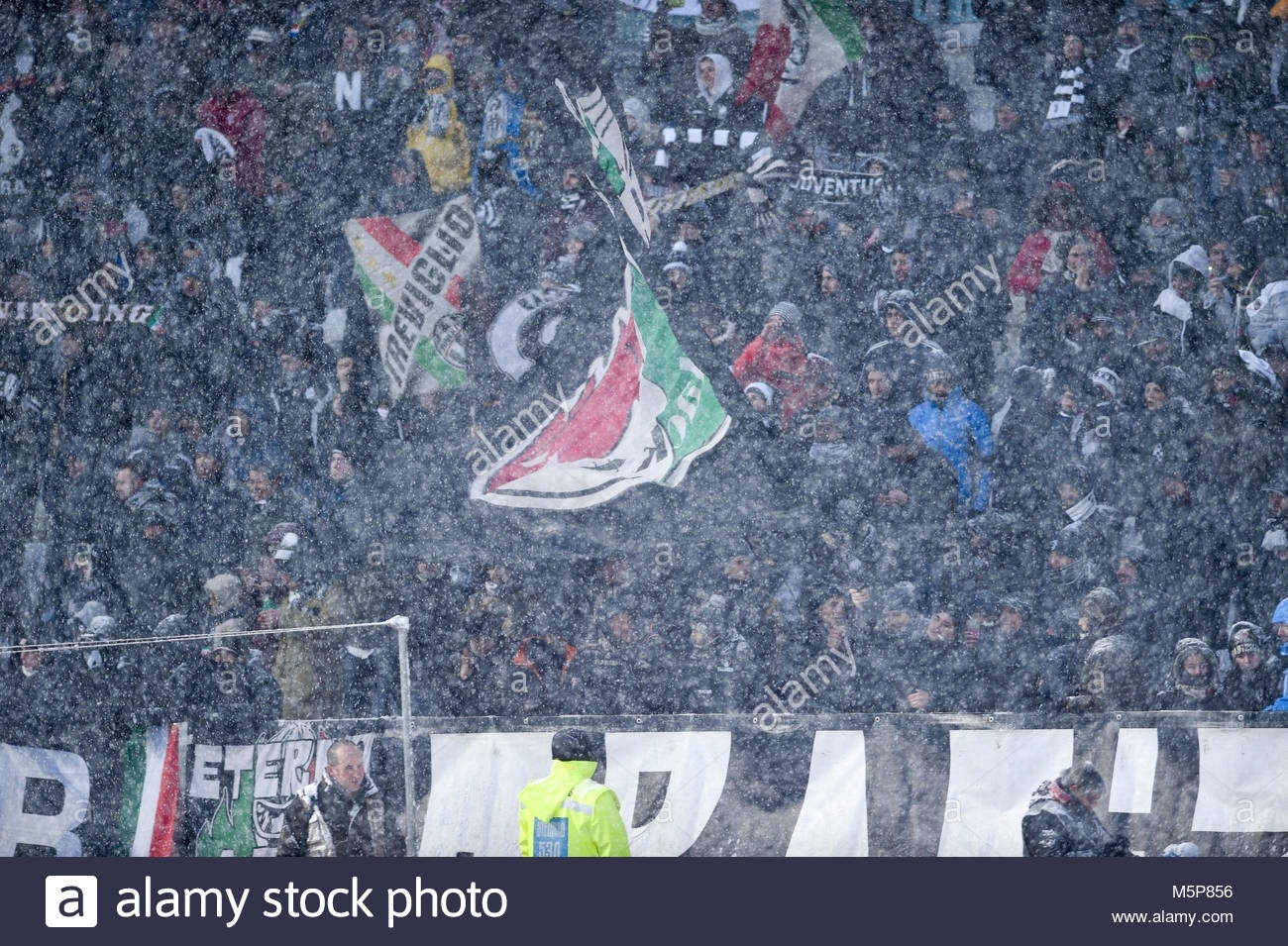}
		\includegraphics[width=0.19\linewidth, height=0.13\linewidth]{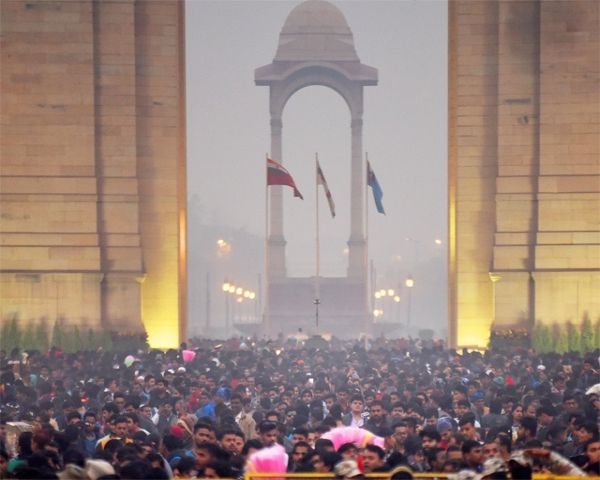}
		\includegraphics[width=0.19\linewidth, height=0.13\linewidth]{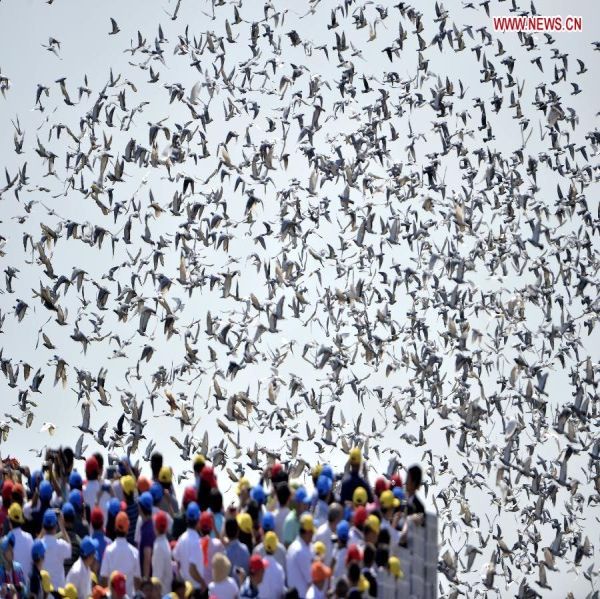}\\
				\includegraphics[width=0.19\linewidth, height=0.13\linewidth]{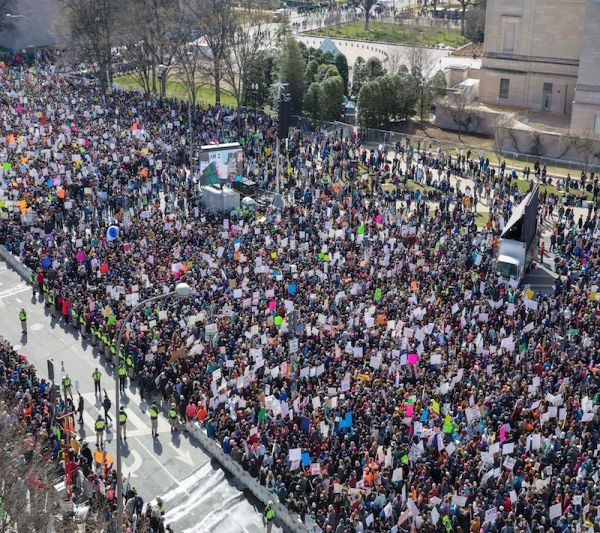}
		\includegraphics[width=0.19\linewidth, height=0.13\linewidth]{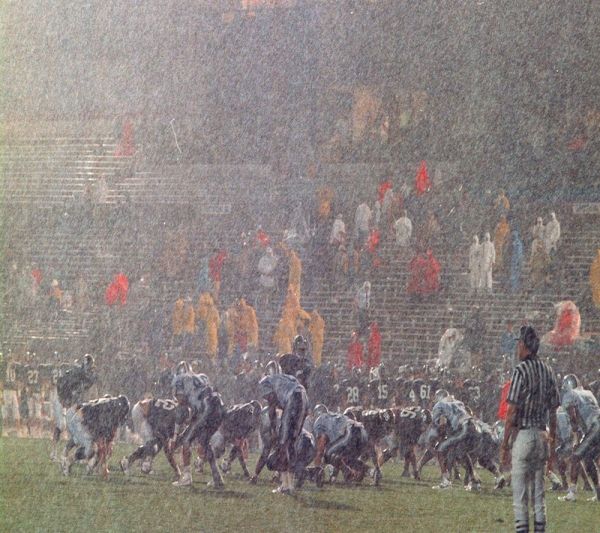}
		\includegraphics[width=0.19\linewidth, height=0.13\linewidth]{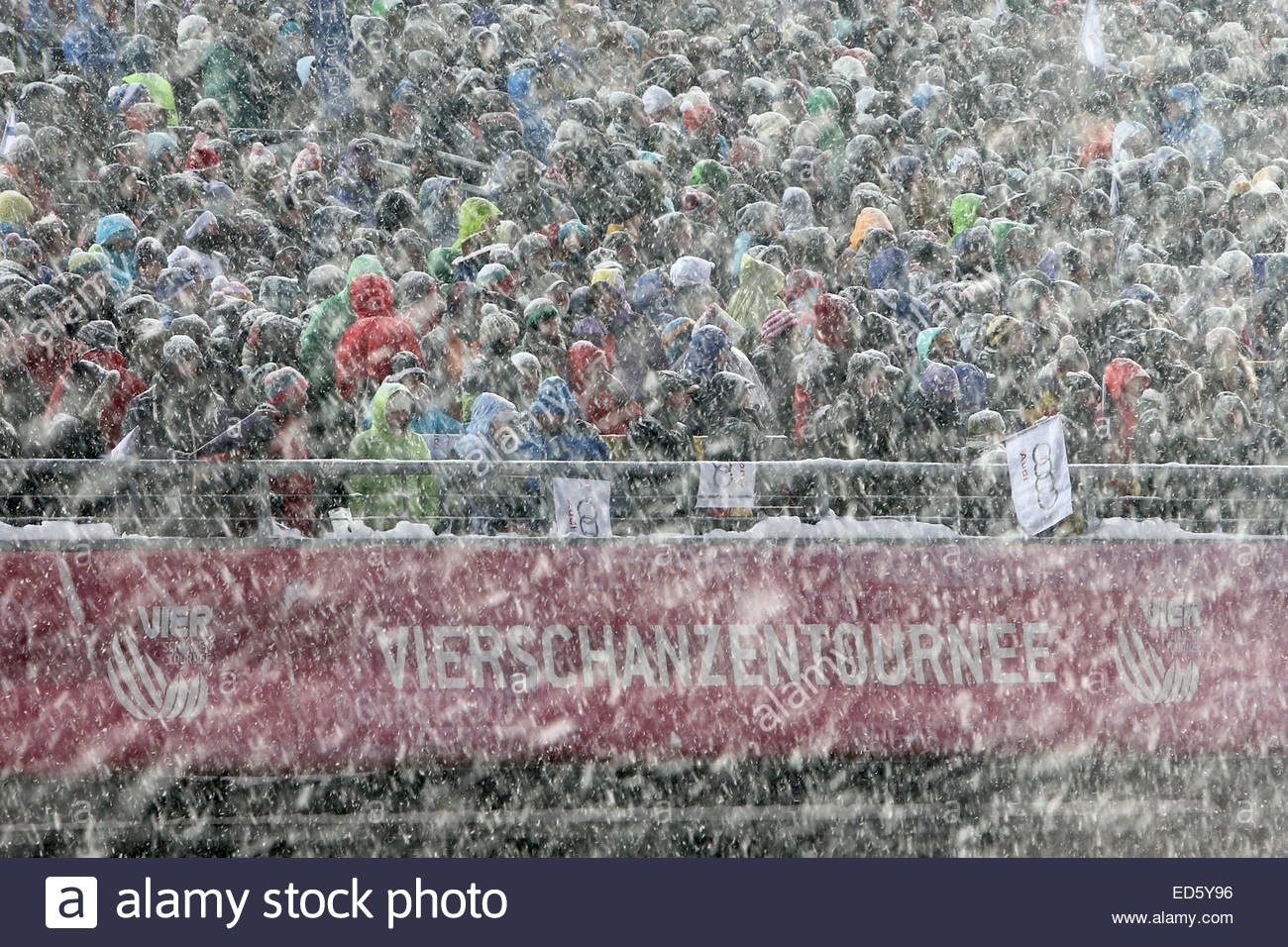}
		\includegraphics[width=0.19\linewidth, height=0.13\linewidth]{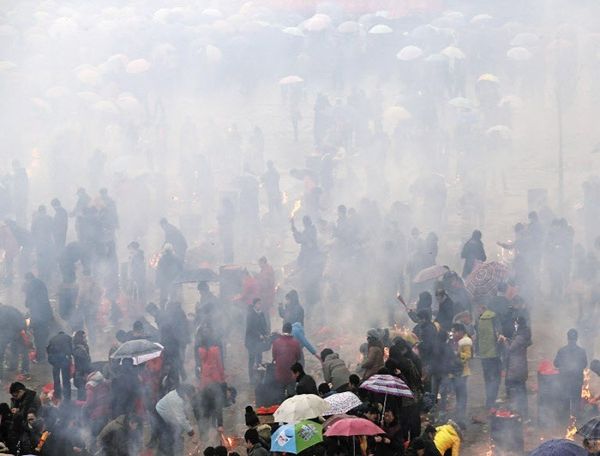}
		\includegraphics[width=0.19\linewidth, height=0.13\linewidth]{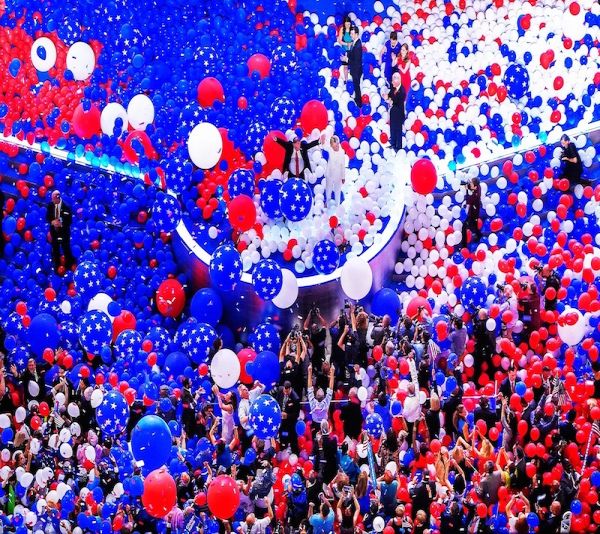}\\
				\includegraphics[width=0.19\linewidth, height=0.13\linewidth]{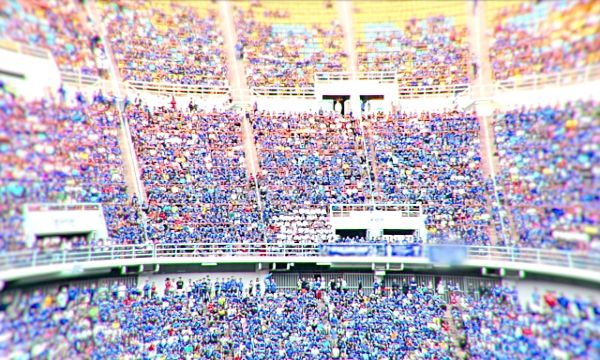}
		\includegraphics[width=0.19\linewidth, height=0.13\linewidth]{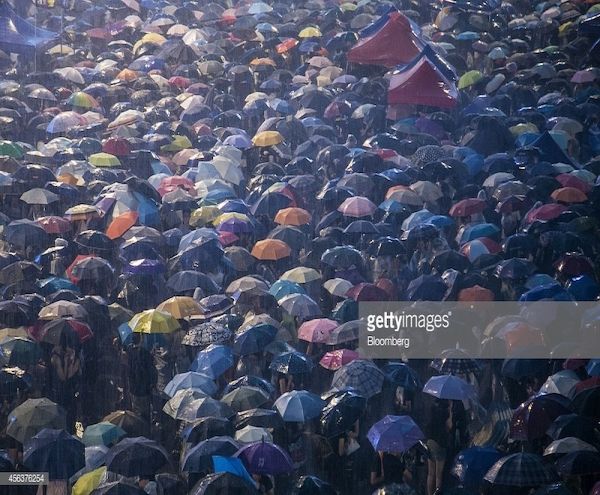}
		\includegraphics[width=0.19\linewidth, height=0.13\linewidth]{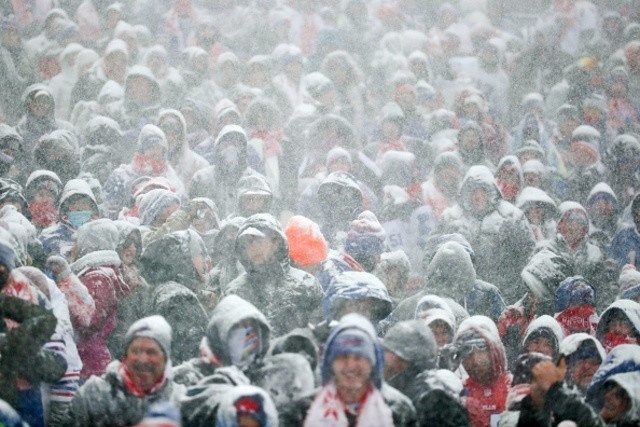}
		\includegraphics[width=0.19\linewidth, height=0.13\linewidth]{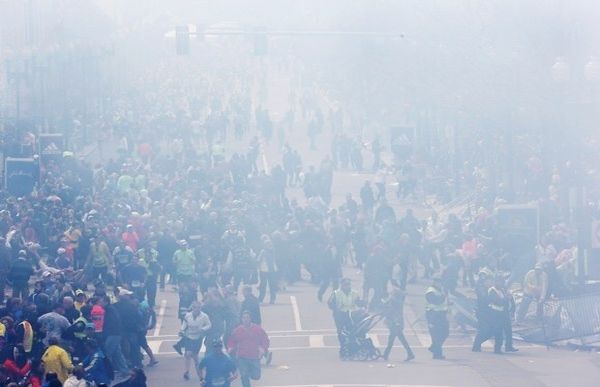}
		\includegraphics[width=0.19\linewidth, height=0.13\linewidth]{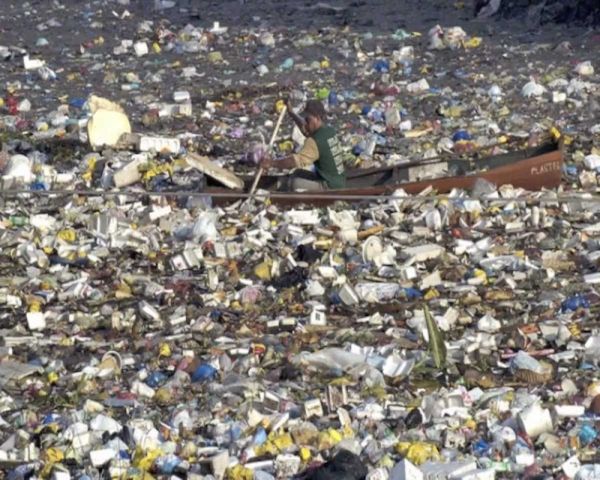}\\ 
		(a)\hskip 85pt(b)\hskip 85pt(c)\hskip 85pt(d)\hskip 85pt(e)  
	\end{center}
	\vskip -14pt \caption{Representative samples of the images in the JHU-CROWD++ dataset. (a) Overall (b) Rain (c) Snow (d) Haze (e) Distractors.}
	\label{fig:dataset_samples}
\end{figure*}

To address these issues, we propose a new large-scale  unconstrained dataset (JHU-CROWD++) with a total of 4,372 images (containing 1,515,005 head annotations) that are collected under a variety of conditions.  Specific care is taken to include images captured under various weather-based degradations. Additionally, we include a set of distractor images that are similar to the crowd images that contain complex backgrounds which may be confused for crowd.  Fig. \ref{fig:dataset_samples} illustrates representative samples of the images in the JHU-CROWD++ dataset under various categories. Furthermore, the  dataset also provides a much richer set of annotations at both image-level and head-level. These annotations include point-wise annotations, approximate sizes, blur-level, occlusion-level, weather-labels, \etc. We also benchmark several representative counting networks, providing an overview of the state-of-the-art performance.

 Next, we consider the design of network architecture for the task of counting. Design of novel networks specifically for the task of counting has improved the counting error by large margins. Architectures have evolved from the simple ones like \cite{zhang2015cross} which consisted of a set of convolutional and fully connected layers, to the most recent complex architectures like SA-Net \cite{cao2018scale} which consists of a set of scale aggregation modules. Typically, most existing works (\cite{zhang2015cross,zhang2016single,  walach2016learning, onoro2016towards, sam2017switching, sindagi2017cnnbased, ranjan2018iterative, cao2018scale,babu2018divide, sindagi2017generating, cao2018scale,sindagi2019ha}) have designed their networks by laying  a strong emphasis on addressing large variations of scale in crowd images. While this strategy of developing robustness towards scale changes has resulted in significant performance gains, it is nevertheless important to exploit other properties like in \cite{ranjan2018iterative, shen2018adversarial,shi2018crowd_negative} to further the improvements.
 
 In a similar attempt, we exploit residual learning mechanism for the purpose of improving crowd counting. Specifically, we present a novel  design based on the VGG16 network \cite{simonyan2014very}, and it employs residual learning to  progressively  generate better quality crowd density maps.  This use of residual learning is inspired by its success in several other tasks like  super-resolution \cite{tai2017image,kim2016accurate,lim2017enhanced,ke2017srn,kim2016accurate}. Although this technique results in improvements in performance, it is important to ensure that only highly confident residuals are used in order to ensure  the effectiveness of residual learning.  To address this issue, we draw inspiration from  the success of uncertainty-based learning mechanism \cite{kendall2017uncertainties,zhu2017deep,devries2018learning}. We propose an uncertainty-based confidence weighting module  that captures high-confidence regions in the feature maps to focus on during the residual learning.  The confidence weights ensure that only highly confident residuals get propagated to the output, thereby increasing the effectiveness of the residual learning mechanism.  Furthermore, we  exploit the additional image-level labels in the proposed dataset to extend the uncertainty-based confidence weighting module by  conditioning it  on the labels to improve the performance specifically in the adverse weather conditions. 
 
 To summarize, the following are our key contributions:
 \begin{itemize}[topsep=0pt,noitemsep,leftmargin=*]
 	\item  We propose a new large-scale unconstrained crowd counting dataset that contains 4,372 images and 1.51 million annotations. The dataset specifically includes a number of images collected under adverse weather conditions. Furthermore, this is the first counting dataset that provides a rich set of annotations such as occlusion, blur, scale, and image-level labels,  \etc. 
 	\item We propose a  crowd counting network that progressively incorporates residual mechanism to estimate high quality density maps.  Furthermore, a set of uncertainty-based confidence weighting modules are introduced in the network  to improve the efficacy of  residual learning. \\
 \end{itemize}

 Note that this work  is an extension of our ICCV 2019 work \cite{sindagi2019pushing}. Compared to our earlier work, we attempt to  improve  both dataset and  proposed method. These improvements are summarized below:
 \begin{itemize}[topsep=0pt,noitemsep,leftmargin=*]
 	\item \textbf{Dataset:} Specifically, we provide 3 key improvements as compared to the JHU-CROWD dataset \cite{sindagi2019pushing}: \\
 	(i) \textit{More number of images}: We increase the number of images in the dataset from 4,250 to 4,372.\\ 
 	(ii) \textit{More number of annotations}: The new dataset contains 31\% more annotations.  (1.51 million v/s 1.15 million).\\
 	(iii) \textit{Better scale annotations}: The earlier version of the dataset contains size indicators for each head in the image. In the new dataset, we provide better scale annotations which consist of approximate width and height of each head. 
 	\item \textbf{Method:} We provide the following improvements in the proposed method:\\
 	(i) \textit{Class conditioning}: We extend the CGDRN method proposed in \cite{sindagi2019pushing} to improve the counting performance in the adverse weather conditions. Specifically, we condition the uncertainty-guided residual estimation  on the image level labels to incorporate weather-based information into the learning process.\\
 	(ii) \textit{New backbone:}We demonstrate that the proposed uncertainty-based residual learning mechanism generalizes to other backbone networks like Res101 \cite{he2016deep}. This results in further improvements on all the datasets. 
 	\item \textbf{Experiments:} We conduct the following new experiments: \\
 	(i) Ablation study: We conduct additional ablation studies where we evaluate the efficacy of number of branches and different network architecture.\\
 	(ii)Benchmarking: We benchmark recently published methods on the newly proposed JHU-CROWD++ dataset. 
 \end{itemize}

 \section{Related work}
 \label{sec:related}

 \noindent\textbf{Crowd Datasets.}  Crowd counting datasets have evolved over time with respect to a number of factors such as size, crowd densities, image resolution, and diversity.  UCSD \cite{chan2008privacy} is among one of the early datasets proposed for counting and it contains 2000 video frames of low resolution with 49,885 annotations. The video frames are collected from a single frame and typically contain low density crowds. Zhang \etal \cite{zhang2015cross} addressed the limitations of UCSD dataset by introducing  the WorldExpo dataset that contains 108 videos with a total of 3,980 frames belonging to 5 different scenes. While the UCSD and WorldExpo datasets contain only low/low-medium densities, Idrees \etal \cite{idrees2013multi} proposed the \UCF dataset specifically for very high density crowd scenarios. However, the dataset consists of only 50 images rendering it impractical for training deep networks. Zhang \etal \cite{zhang2016single} introduced the ShanghaiTech dataset which has better diversity in terms of scenes and density levels as compared to earlier datasets.  The dataset is split into two parts: Part A (containing high density crowd images) and Part B (containing low density crowd images). The entire dataset contains 1,198 images with 330,165 annotations. Recently, Idrees \etal \cite{idrees2018composition} proposed  a new large-scale crowd dataset containing 1,535 high density images images with a total of  1.25 million annotations.  Wang \etal \cite{wang2019learning} introduced a synthetic crowd counting dataset that is based on GTA V electronic game. The dataset consists of 15,212 crowd images under a diverse set of scenes. In addition, they  proposed  a SSIM based CycleGAN \cite{zhu2017unpaired} for adapting the network trained on   synthetic images to real world images.  Most recently, Wang \etal \cite{wang2020nwpu} released a large-scale crowd counting dataset (NWPUCrowd) consisting of 5,109 images with ~2.13 million annotations.\\

 \noindent\textbf{Crowd Counting.}  Traditional approaches for crowd counting from single images are based on hand-crafted representations and different regression techniques. Loy \etal \cite{loy2013crowd} categorized these methods into (1) detection-based methods \cite{li2008estimating} (2) regression-based methods \cite{ryan2009crowd,chen2012feature,idrees2013multi} and (3) density estimation-based methods \cite{lempitsky2010learning,pham2015count,xu2016crowd}. Interested readers are referred to \cite{chen2012feature,li2015crowded} for a  more comprehensive study of different crowd counting methods.

 Recent advances in CNNs have been exploited for the task of crowd counting and these methods \cite{wang2015deep,zhang2015cross,sam2017switching,arteta2016counting,walach2016learning,onoro2016towards,zhang2016single,sam2017switching,sindagi2017generating,boominathan2016crowdnet,wang2018defense,onoro2018learning} have demonstrated significant improvements over the traditional methods. A recent survey \cite{sindagi2017survey} categorizes these approaches based on the network property and the inference process. Walach \etal \cite{walach2016learning} used CNNs with layered boosting approach to learn a non-linear function between an image patch and count. Recent work \cite{zhang2016single,onoro2016towards} addressed the scale issue using different architectures. Sam \etal \cite{sam2017switching} proposed a VGG16-based  switching classifier that first identifies appropriate regressor based on the content of the input image patch. More recently, Sindagi \etal \cite{sindagi2017generating} proposed to incorporate global and local context from the input image into the density estimation network. In another approach, Cao \etal \cite{cao2018scale} proposed a encoder-decoder network with scale aggregation modules.

 In contrast to these  methods that emphasize on specifically addressing large-scale variations in head sizes, the most recent methods (\cite{babu2018divide} ,\cite{shen2018adversarial}, \cite{shi2018crowd_negative}, \cite{liu2018leveraging}, \cite{ranjan2018iterative}) have focused on other properties of the problem. For instance, Babu \etal \cite{babu2018divide} proposed a mechanism  to incrementally increase the network capacity  conditioned on the dataset. Shen \etal \cite{shen2018adversarial} overcame the issue of blurred density maps by utilizing  adversarial loss.   In a more recent approach, Ranjan \etal \cite{ranjan2018iterative} proposed a two-branch network to  estimate density map in a cascaded manner.  Shi \etal \cite{shi2018crowd_negative} employed deep negative correlation based learning for more generalizable features. Liu \etal \cite{liu2018leveraging} used  unlabeled data for counting by proposing a new framework that involves learning to rank. 
 
 Recent approaches like \cite{liu2018adcrowdnet,wan2019residual,zhao2019leveraging,sindagi2019inverse,sindagi2019ha, sindagi2019multi} have aimed at incorporating various forms of related information like attention \cite{liu2018adcrowdnet}, semantic priors \cite{wan2019residual}, segmentation \cite{zhao2019leveraging}, inverse attention \cite{sindagi2019inverse}, and hierarchical attention \cite{sindagi2019ha} respectively into the network.   Other techniques such as \cite{jiang2019crowd,shi2019revisiting,liu2019context,zhang2019wide,wan2019adaptive} leverage features from different layers of the network using different techniques like   trellis style encoder decoder \cite{jiang2019crowd},  explicitly considering perspective \cite{shi2019revisiting},  context information  \cite{liu2019context},  adaptive density map generation \cite{wan2019adaptive} and multiple views \cite{zhang2019wide}. More recently, Sam \etal \cite{sam2019locate} introduced a detection framework for densely crowded scenarios where the network is trained using estimated bounding-boxes. Ma \etal \cite{ma2019bayesian} proposed a novel Bayesian loss function for training counting networks, which involves supervision on the count expectation at each annotated point. While most of the existing approaches are focused on counting in 2D plane, Zhang \etal \cite{zhang20203d} propose to solve the multi-view crowd counting task through 3D feature fusion with 3D scene-level density maps.  For a comprehensive study on various crowd counting techniques, the reader is referred to detailed surveys like  \cite{sindagi2017survey,gao2020cnn}.

 \begin{figure*}[htp!]
 	\begin{center}
 		\includegraphics[width=0.32\linewidth, height=0.17\linewidth]{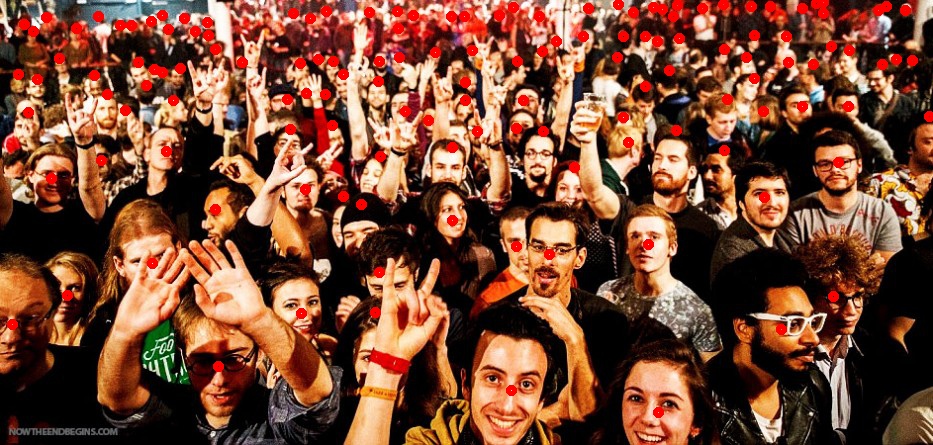} 
 		\includegraphics[width=0.32\linewidth, height=0.17\linewidth]{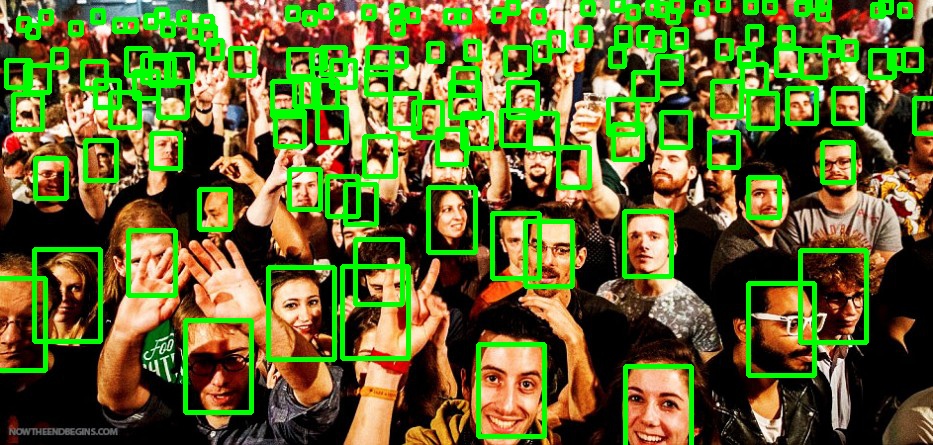}
 		\includegraphics[width=0.32\linewidth, height=0.17\linewidth]{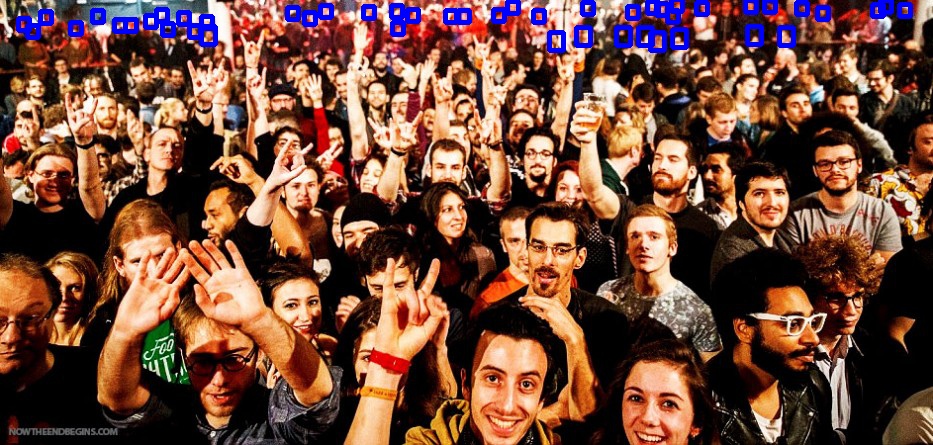} \\\vspace{1pt}
 		
 		\includegraphics[width=0.32\linewidth, height=0.2\linewidth]{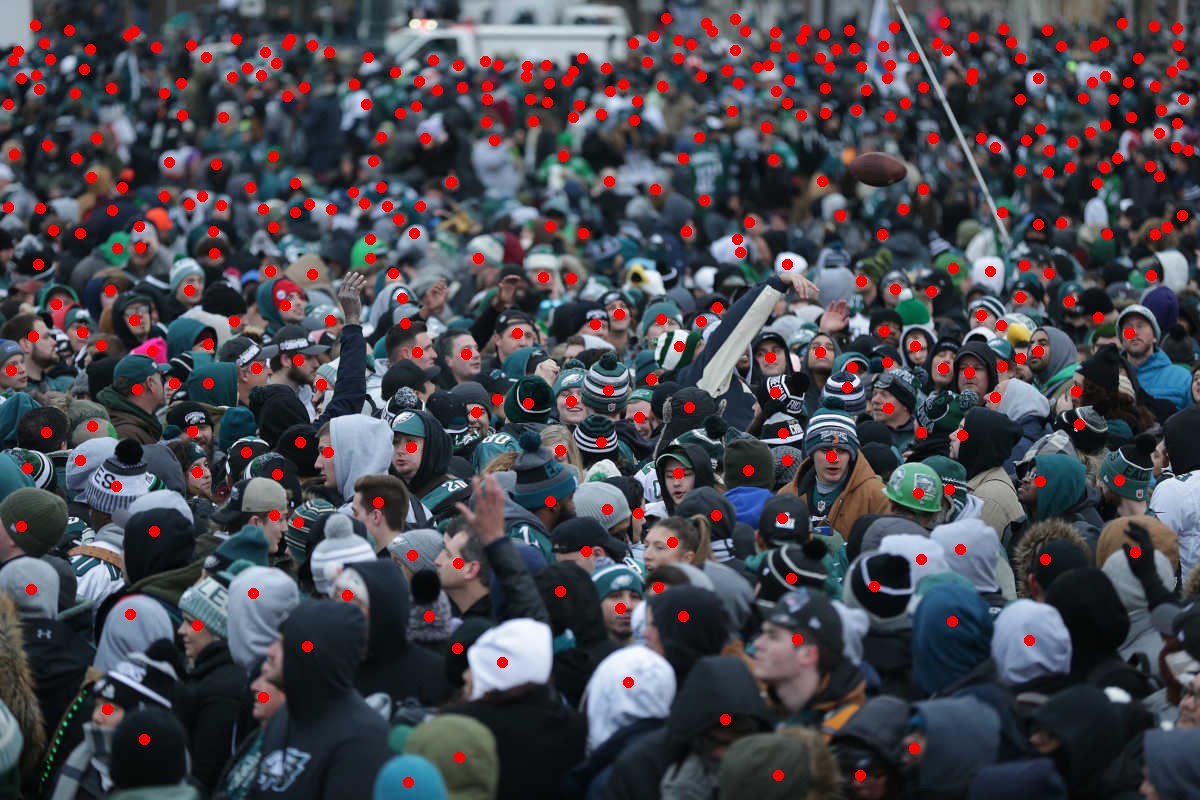}
 		\includegraphics[width=0.32\linewidth, height=0.2\linewidth]{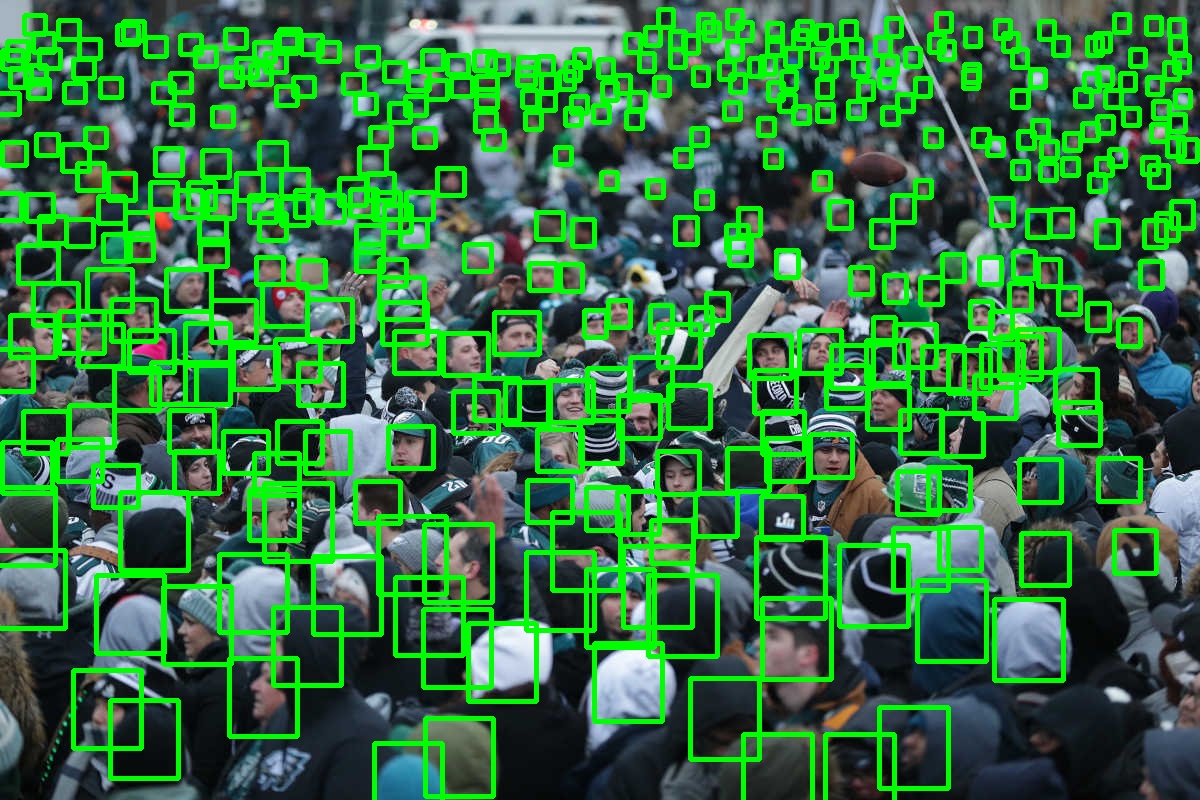}
 		\includegraphics[width=0.32\linewidth, height=0.2\linewidth]{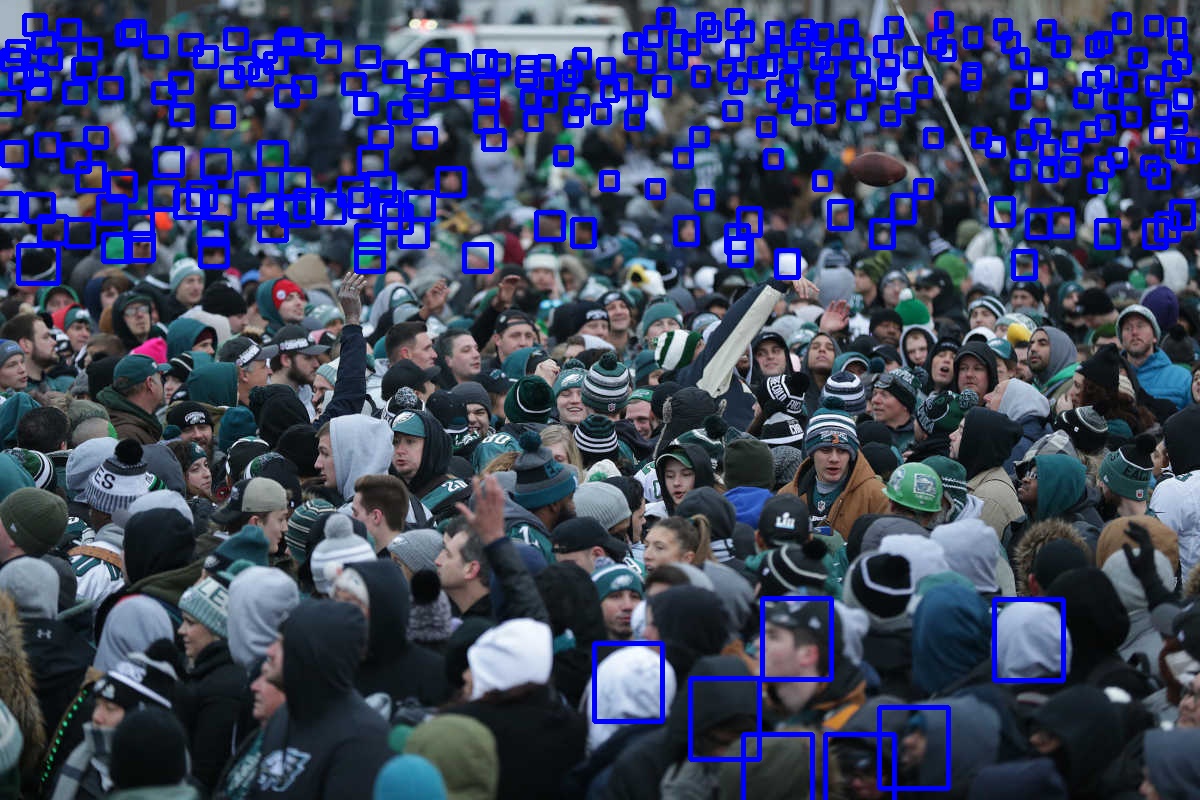} \\\vspace{1pt}
 		
 		\includegraphics[width=0.32\linewidth, height=0.2\linewidth]{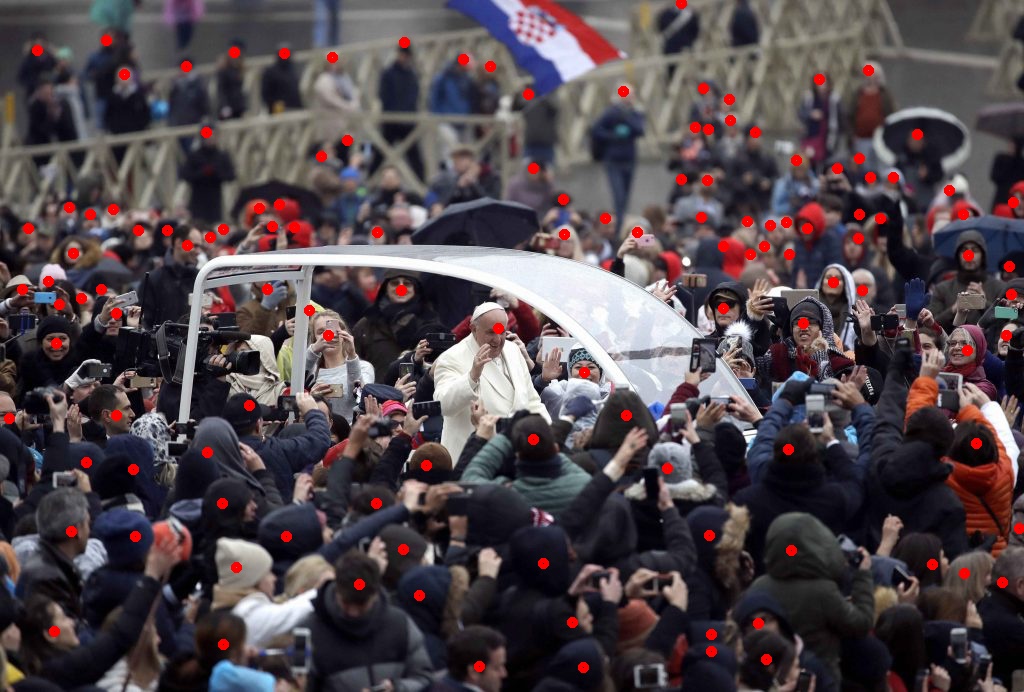}
 		\includegraphics[width=0.32\linewidth, height=0.2\linewidth]{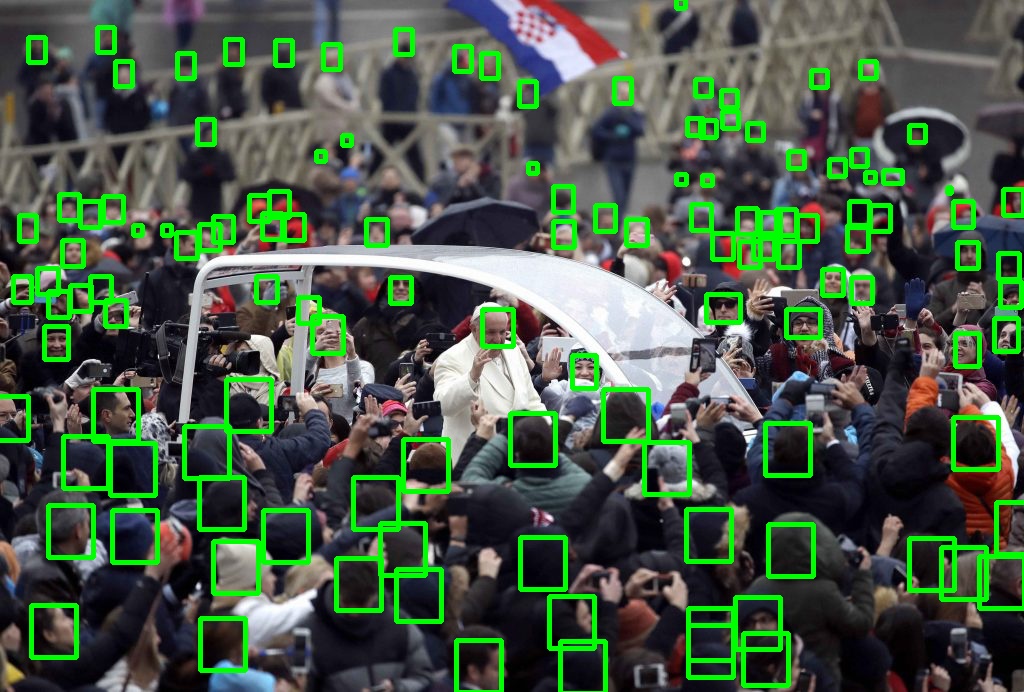}
 		\includegraphics[width=0.32\linewidth, height=0.2\linewidth]{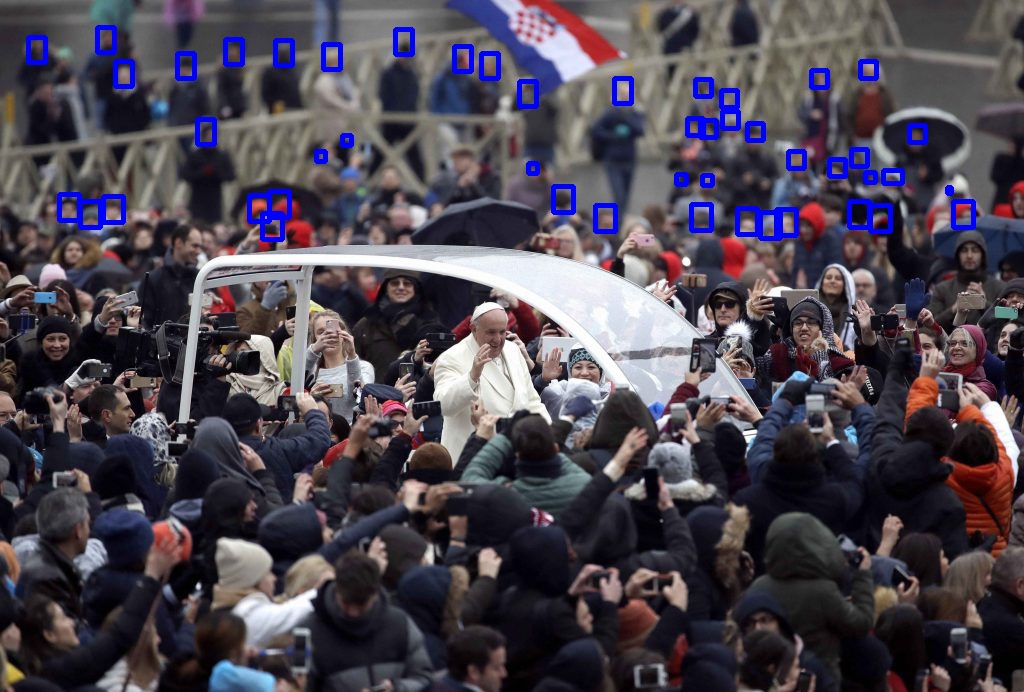} \\	\vspace{1pt}
 		
 		\includegraphics[width=0.32\linewidth, height=0.2\linewidth]{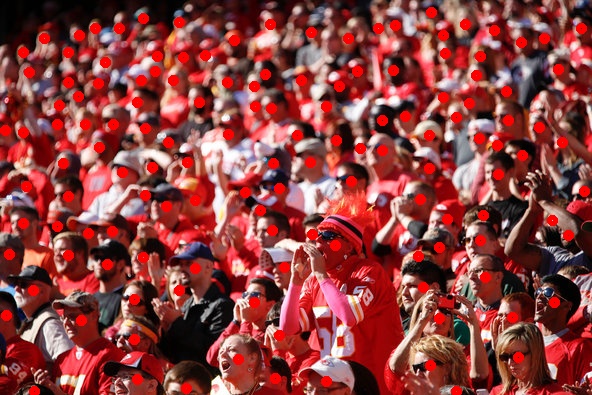}
 		\includegraphics[width=0.32\linewidth, height=0.2\linewidth]{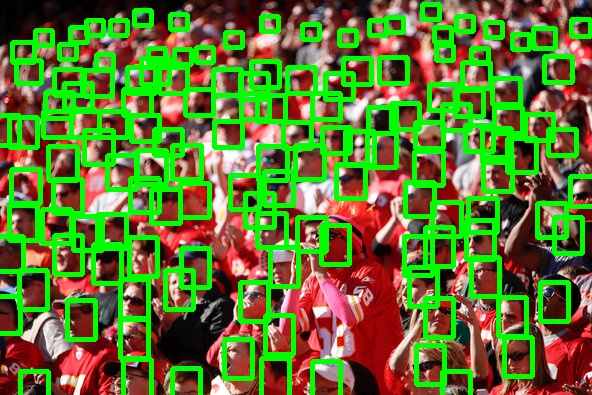}
 		\includegraphics[width=0.32\linewidth, height=0.2\linewidth]{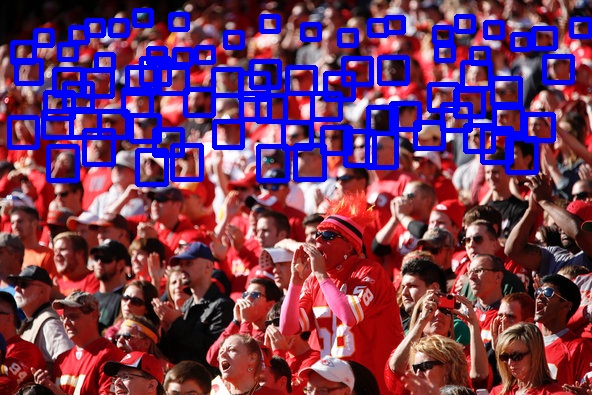} \\		\vspace{1pt}	
 		
 		(a)\hskip 160pt(b)\hskip 160pt(c)
 	\end{center}
 	\vskip -15pt \caption{Examples of head-level annotations: (a) Dots (b) Approximate sizes  (c) Blur-level.}
 	\label{fig:annotation_samples}
 \end{figure*}

 \begin{table*}[ht!]
 	\caption{Comparison of different datasets. P: Point-wise annotations for head locations, O: Occlusion level per head, B: Blur level per head, S: Size indicator per head, S$^\dagger$: Approximate size (w$\x$h), I: Image level labels.}
 	\label{tab:dataset_summary}
 	\vskip -8pt
 	\resizebox{1\columnwidth}{!}{%
 		\begin{tabular}{lcccccccc}
 			\hline
 			{Dataset}      & {\begin{tabular}[c]{@{}c@{}}Num of\\ Images\end{tabular}} & {\begin{tabular}[c]{@{}c@{}}Num of\\ Annotations\end{tabular}} & {\begin{tabular}[c]{@{}c@{}}Avg\\  Count\end{tabular}} & {\begin{tabular}[c]{@{}c@{}}Max \\ Count\end{tabular}} & {\begin{tabular}[c]{@{}c@{}}Avg \\ H$\x$W\end{tabular}} & {\begin{tabular}[c]{@{}c@{}}Weather\\ degradations\end{tabular}} & {Distractors} & {\begin{tabular}[c]{@{}c@{}}Type of \\ annotations\end{tabular}} \\ \hline\hline
 			UCSD \cite{chan2008privacy}          & 2,000 & 49,885  & 25 & 46  & 158$\times$238     & $\xmark$   & $\xmark$             & P     \\  
 			Mall \cite{chen2012feature}          & 2,000 & 62,325    & -      & 53     & 320$\times$240  & $\xmark$  & $\xmark$ & P    \\  
 			\UCF \cite{idrees2013multi}          & 50     & 63,974      & 1,279        & 4,543       & 2101$\times$2888        & $\xmark$    & $\xmark$ & P  \\  
 			WorldExpo '10 \cite{zhang2015cross} & 3,980  & 199,923      & 50         & 253       & 576$\times$720 & $\xmark$   & $\xmark$ & P \\  
 			ShanghaiTech \cite{zhang2016single}         & 1,198    & 330,165  & 275 & 3,139 & 598$\times$868 & $\xmark$ & $\xmark$ & P\\  
 			UCF-QNRF \cite{idrees2018composition} & 1,535 & 1,251,642 & 815   & 12,865  & 2,013$\times$2,902    & $\xmark$ & $\xmark$  &P \\  
 			NWPU-CROWD \cite{wang2020nwpu} & 5,109 & 2,133,238 & 418   & 20,033  & 2,311$\times$ 3,383    & $\xmark$ & $\cmark$  &P \\  \hline\hline
 			JHU-CROWD (ours)      & 4,250  & 1,114,785 & 262 & 7,286 & 900$\times$1,450 & $\cmark$ & $\cmark$ & P, O, B, S, I \\ 
 			{JHU-CROWD++ (ours)}       & 4,372  & 1,515,005 & 346 & 25,791 & 910$\times$1,430 & $\cmark$ & $\cmark$ & P, O, B, S$^\dagger$, I \\ \hline
 		\end{tabular}
 	}
 	\vskip-8pt
 \end{table*}

 \section{JHU-CROWD++: Large-scale crowd counting dataset}
 In this section, we first motivate the need for a new crowd counting dataset, followed by a detailed description of the various factors and conditions while collecting the dataset. 
 
 \subsection{Motivation and dataset details}
 \label{ssec:motivation_dataset}
 As discussed earlier, existing datasets (such as \UCF \cite{idrees2013multi}, World Expo '10 \cite{zhang2015cross} and ShanghaiTech \cite{zhang2016data})  have enabled researchers to develop novel counting networks that are robust to several factors such as variations in scale, pose, view \etc. Several recent methods have specifically addressed the large variations in scale by proposing different approaches such as multi-column networks \cite{zhang2016single}, incorporating global and local context \cite{sindagi2017generating}, scale aggregation network \cite{cao2018scale}, \etc. These methods are largely successful in addressing issues in the existing datasets, and there is pressing need to identify newer set of challenges that require attention from the crowd counting community.

 In what follows, we describe the shortcomings of existing datasets and discuss the ways in which we overcome them: 
 
 \noindent(i) \textit{Limited number of training samples}:  Typically, crowd counting datasets have limited number of images available for training and testing. For example, ShanghaiTech dataset \cite{zhang2016single} has only 1,198 images and this low number of images results in lower diversity of the training samples. Due to this issue, networks trained on this dataset will have reduced generalization capabilities. Although datasets like Mall \cite{chen2012feature}, WorldExpo '10 \cite{zhang2015cross} have higher number of images, it is   important to note that these images are from a set of video sequences from surveillance cameras and hence, they have limited diversity in terms of background scenes and number of people. Most recently, Idrees \etal \cite{idrees2018composition} addressed this issue by introducing a high-quality dataset (UCF-QNRF) that has images collected from various geographical locations under a variety of conditions and scenarios. Although it has a large set of diverse scenarios, the number of samples is still limited from the perspective of training deep neural networks. 
 
 \begin{figure}[t!]
 	\begin{center}
 		\includegraphics[width=1\linewidth]{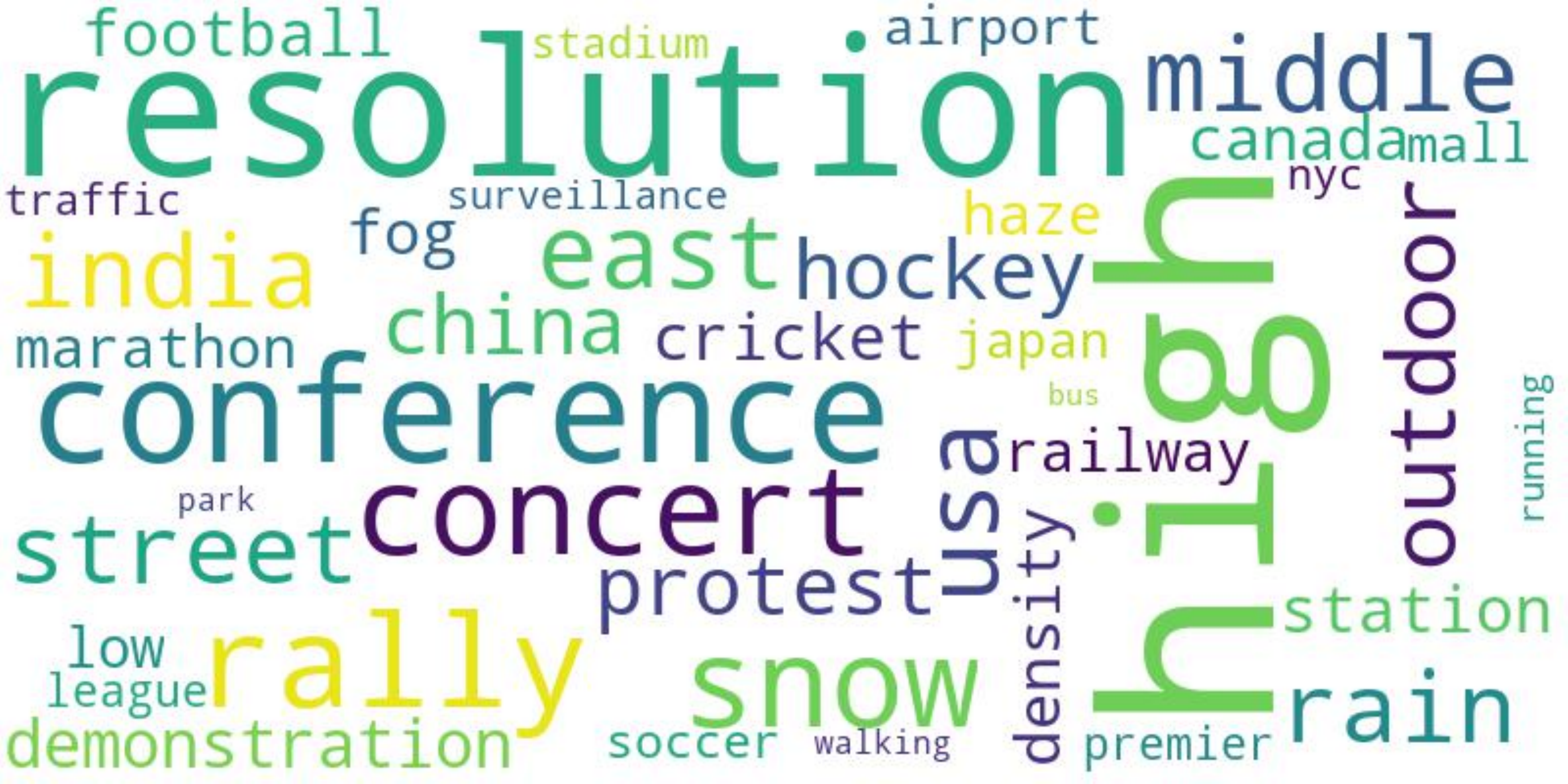}
 	\end{center}
 	\vskip -10pt \caption{Summary of keywords used to scrape the internet for images.}
 	\label{fig:wc}
 \end{figure}

 To address this issue, we collect a new large scale  unconstrained dataset with a total of 4,372 images that are collected under a variety of conditions. Such a large number of images results in increased diversity in terms of count, background regions, scenarios, \etc. as compared to existing datasets.  The images are collected from several sources on the internet using different keywords such as crowd, crowd+marathon, crowd+walking, crowd+India, \etc. A summary of the keywords used for the search purpose is illustrated in Fig. \ref{fig:wc}. \\

 \noindent(ii) \textit{Absence of adverse conditions}: Typical application of crowd counting is  video surveillance in outdoor scenarios which involve regular weather-based degradations such as haze, snow, rain \etc. It is crucial that networks, deployed under such conditions, are robust. 
 
 To overcome this issue, specific care is taken during our dataset collection efforts to include images captured under various weather-based degradations such as rain, haze, snow, \etc. (as as shown in Fig. \ref{fig:dataset_samples}(b-d)).   Table \ref{tab:weather} summarizes images collected under adverse conditions. \\

 \begin{table}[h!]
 	\centering
 	\caption{Summary of images collected under adverse conditions.}
 	\label{tab:weather}
 	\vskip -8pt 
 	\resizebox{1\textwidth}{!}{%
 		\begin{tabular}{l|ccc|c}
 			\hline
 			{Degradation type} & {Rain} & {Snow} & {Fog/Haze} & {Total} \\ \hline\hline
 			No. of images            & 145           & 201           & 168               & 514            \\   
 			No. of annotations       & 40,328        & 47,347        & 48,821            & 136,496         \\ \hline
 		\end{tabular}%
 		
 	}
 \end{table}
 
 \noindent(iii) \textit{Dataset bias}: Existing datasets focus on collecting only images with crowd, due to which a deep network trained on such a dataset may end up learning bias in the dataset. Due to this error, the network will erroneously predict crowd even in scenes that do not contain crowd.
 
 In order to address this, we include a set of distractor images that are similar to crowd images but contain very few people. These images can enable the network to avoid learning bias in the dataset. The total number of distractor images in the dataset is 106.  Fig. \ref{fig:dataset_samples}(e) shows sample distractor images. \\
 
 \noindent(iv) \textit{Limited annotations}: Typically, crowd counting datasets provide point-wise annotations for every head/person in the image, \ie each image is provided with a list of $x,y$ locations of the head centers. While these annotations enable the networks to learn the counting task,  absence of more information such as occlusion level, head sizes, blur level \etc. limits the learning ability of the networks. For instance, due to the presence of large variations in perspective, size of the head is crucial to determine the precise count. One of the reasons for these missing annotations is that crowd images typically contain several people and it is highly labor intensive to obtain detailed annotations such as size. 
 
\begin{figure}[t!]
	\begin{center}
		\includegraphics[width=1\linewidth]{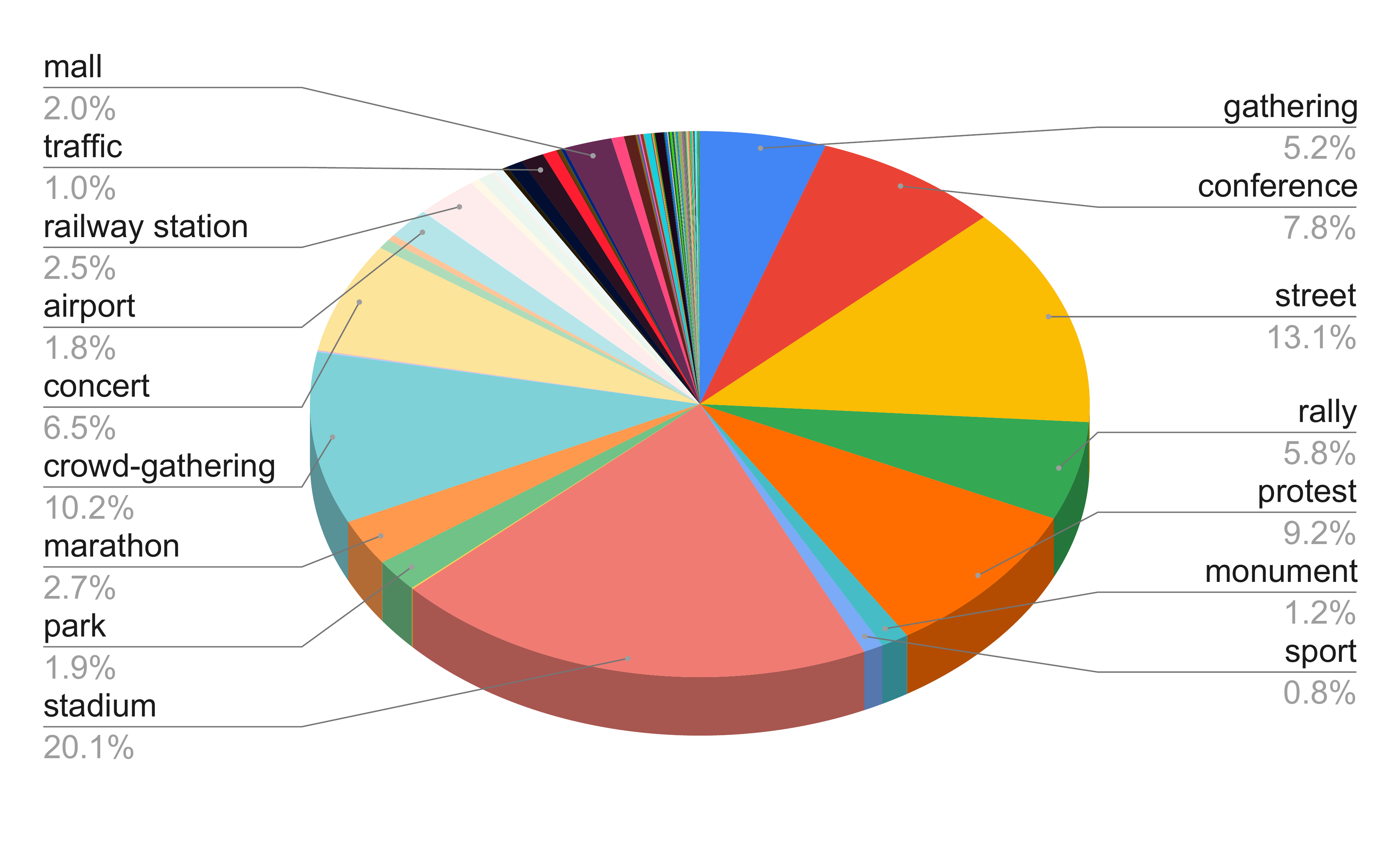}
	\end{center}
	\vskip -20pt \caption{Distribution of image-level labels.}
	\label{fig:image-level-labels}
\end{figure}

 To enable more effective learning, we collect a much  richer set of annotations at both  head-level/point-level and image-level. These are described below:
 \begin{itemize}[topsep=0pt,noitemsep,leftmargin=*]
 	\item  Head-level/point-level annotations include $x,y$ locations of heads and corresponding occlusion level, blur level and size level. The total number of point-level annotations in the dataset are 1,515,005. Occlusion label has three levels: \{\textit{un-occluded, partially occluded, fully occluded}\}. Blur level has two labels: \{\textit{blur, no-blur}\}. In JHU-CROWD \cite{sindagi2019pushing},  each head is labeled with a size indicator.  We improve over these size annotations  by providing ``approximate" size (width and height) for each head annotation. To obtain these, annotators were instructed to annotate bounding boxes for a set of neighbouring heads which have similar sizes. \textit{Note that these bounding boxes are only ``approximate'' and are not as accurate as the ones found in detection datasets}. Fig. \ref{fig:annotation_samples} illustrates sample annotations provided in our dataset. 
 	\item  Image level annotations include scene-labels (such as \textit{marathon, mall, railway station, stadium, \etc.}) and the weather-labels (rain, snow and fog). Fig. \ref{fig:image-level-labels} illustrates the distribution of scene-labels in the proposed dataset.  \\
 \end{itemize}

 \begin{table}[t!]
 	\centering
 	\caption{Distribution of images under different densities.}
 	\label{tab:low-med-high}
 	\vskip -10pt 
 	\resizebox{1\textwidth}{!}{%
 		\begin{tabular}{l|c|c|c|c}
 			\hline
 			{Density} & {Low (0-50)} & {Med (51-500)} & {High (500+)} & {Total} \\ \hline
 			No. of images            & 1,228           & 2,512           & 632               & 4,372            \\   
 			\hline
 		\end{tabular}%
 	}
 	
 \end{table}
 
 \subsection{Data collection process}
 
 We used different sources like google images, bing images, flickr, \etc for collecting the images. The keywords used for searching were carefully selected to ensure diversity in terms of density, weather, geographical locations, scene-type, resolution, events, physical structures \etc. The keywords are summarized in Fig. \ref{fig:wc} and Table \ref{tab:keywords}. After the collection process, duplicate images were detected  and eliminated from the dataset.   The remaining images were annotated with the help of Amazon Mechanical Turk workers. The workers were explicitly instructed to annotate the center of head. Additionally, they were also instructed to label the occlusion type and blur-level. The annotated images were then evaluated manually verified  to filter out incorrectly labeled images. Such images were sent again to the annotation  and verification processes.   
 
 \begin{table}[h!]
 	\caption{List of keywords used for searching.}
 	\label{tab:keywords}
 	\vskip-10pt
 	\resizebox{1\textwidth}{!}{%
 	\begin{tabular}{l|l}
 		\hline
 		Factor                & Keyword                                                                                                                                                       \\ \hline\hline
 		Crowd density         & low crowd, small crowd, large crowd, high density crowd                                                                                                       \\ \hline
 		Geographical location & US, China, India, Iran, Iraq, Canada                                                                                                                          \\ \hline
 		Structures            & airport, mall, railway station, street, park, stadium, traffic                                                                                                \\ \hline
 		Events                & \begin{tabular}[c]{@{}l@{}}concert, protest, rally, festival, sports, cricket, football, \\ soccer, hockey, premier league, conference, marathon\end{tabular} \\ \hline
 		Weather               & rain, snow, fog, haze,  low-light                                                                                                                             \\ \hline
 	\end{tabular}
}
\vskip -15pt
 \end{table}

\begin{figure}[t!]
	\begin{center}
		\includegraphics[width=1\linewidth]{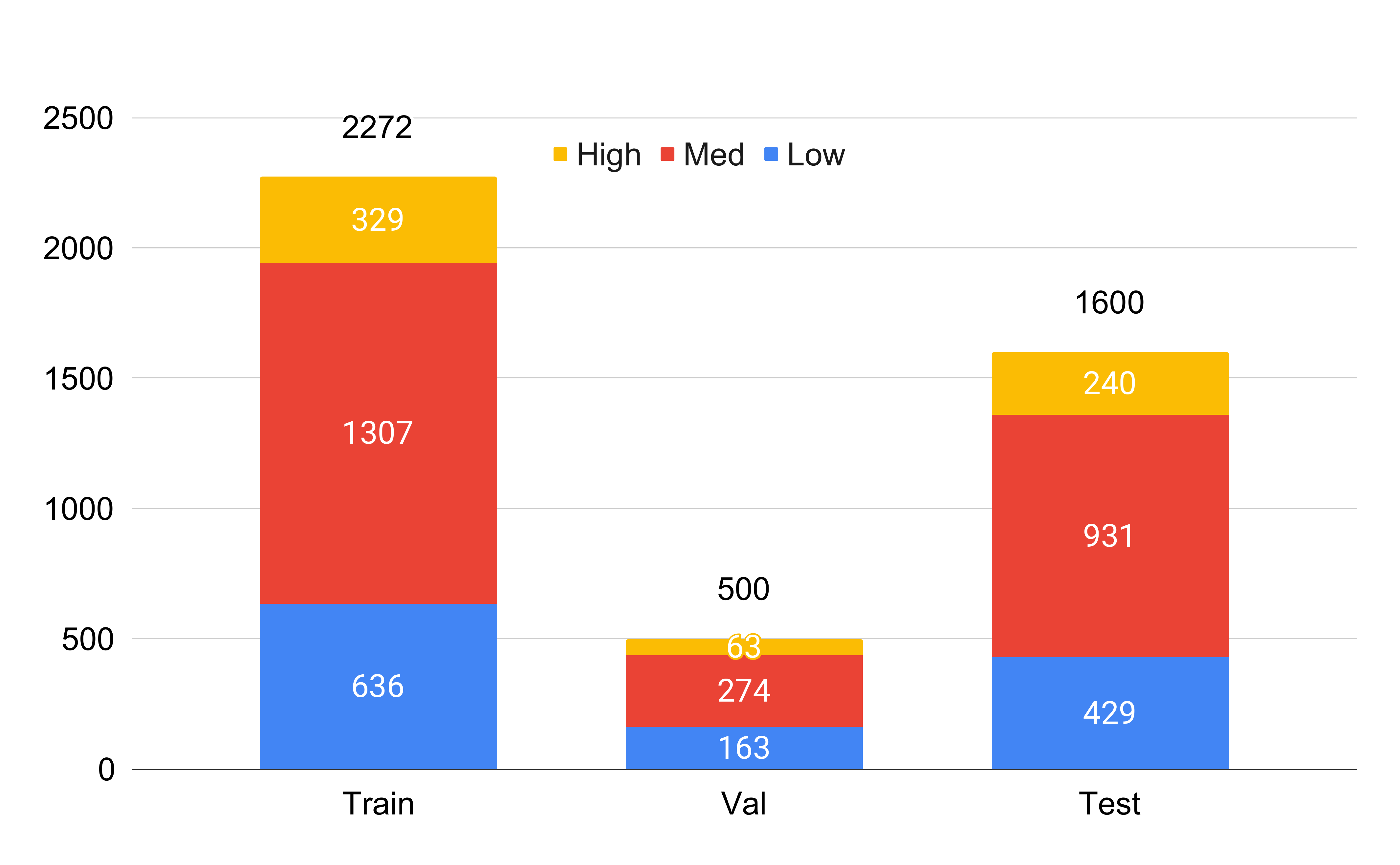}
	\end{center}
	\vskip -20pt \caption{Distribution of images of different density levels in train, val and test sets.}
	\label{fig:split-density}
\end{figure}

\begin{figure}[t!]
	\begin{center}
		\includegraphics[width=1\linewidth]{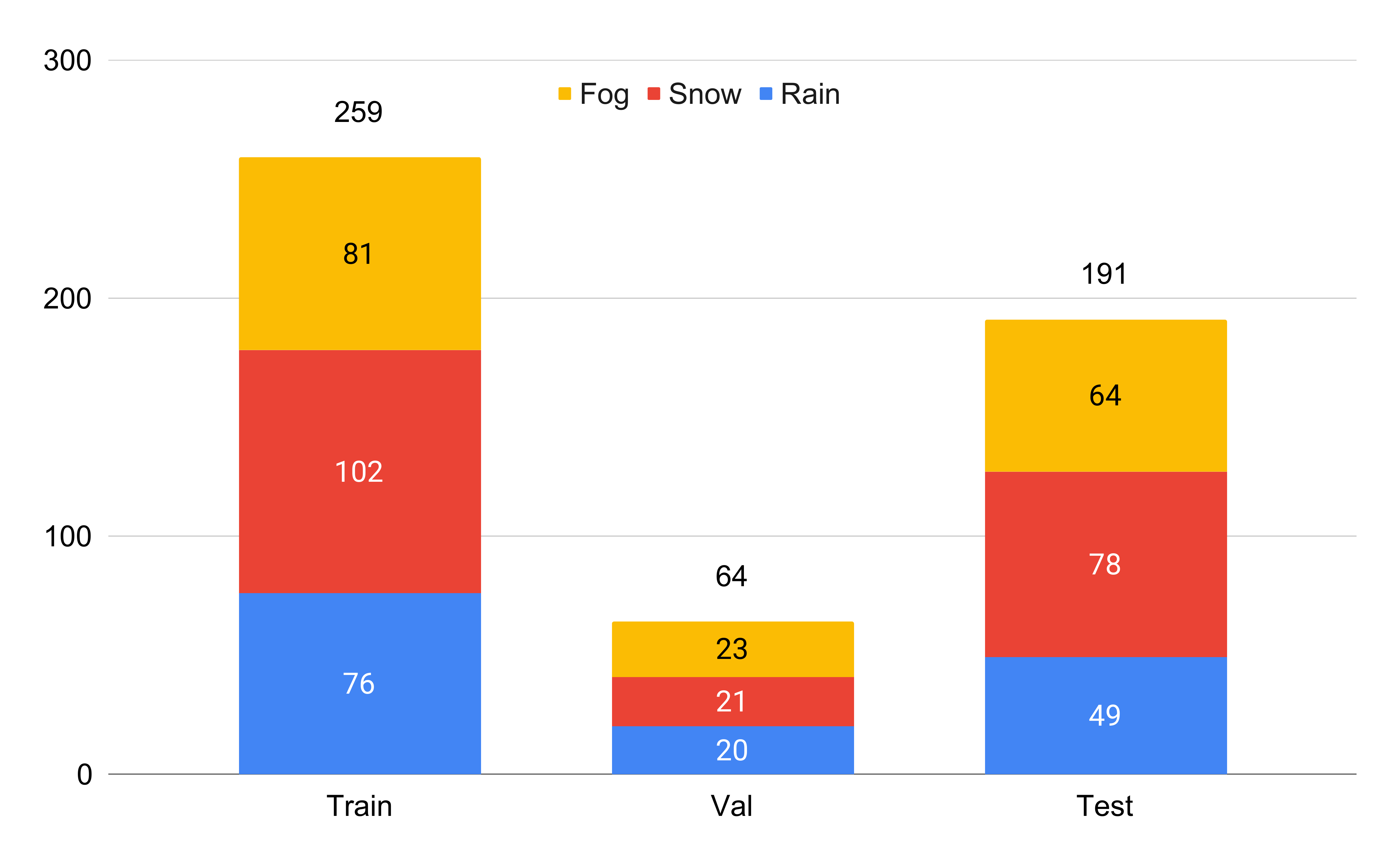}
	\end{center}
	\vskip -20pt \caption{Distribution of images of weather conditions  in train, val and test sets.}
	\label{fig:split-weather}
\end{figure}

 \begin{figure*}[ht!]
	\begin{center}
		\includegraphics[width=1\linewidth]{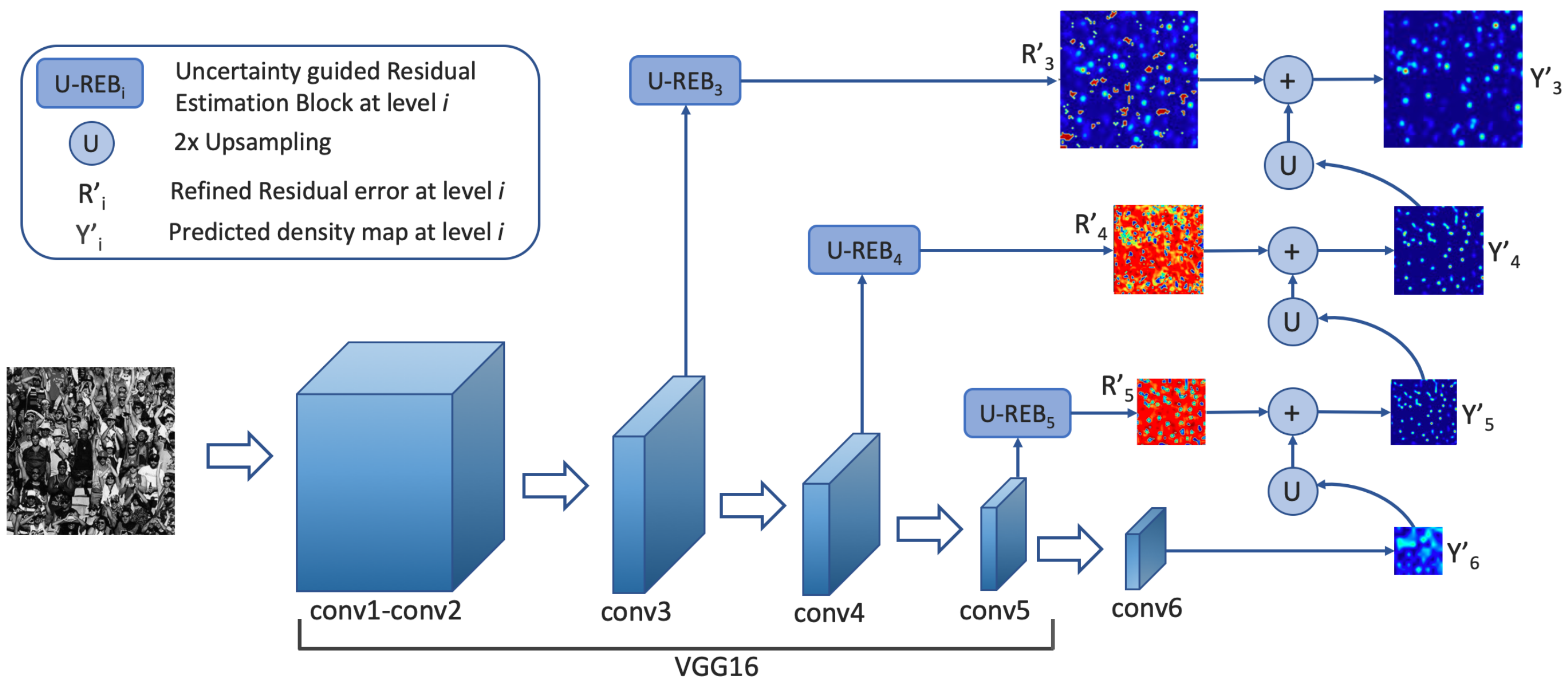}
	\end{center}
	\vskip -15pt \caption{Overview of the proposed method. Coarse density map from the deepest layer of the base network is refined using the residual map estimated by the shallower layer. The residual estimation is performed by  $U$-$REB_i$. In the residual maps, red indicates negative values and cyan indicates positive value.}
	\label{fig:arch}
\end{figure*}

 \subsection{Summary and evaluation protocol}
 Fig. \ref{fig:dataset_samples} illustrates representative samples of the images in the JHU-CROWD++ dataset under various categories.  Table \ref{tab:dataset_summary} summarizes the proposed dataset in comparison with  the existing ones. It can be observed that the proposed dataset enjoys a host of  properties such as a richer set of  annotations, weather-based degradations and distractor images. With these properties, the proposed dataset will serve as a good complementary to other datasets such as UCF-QNRF and NWPU-CROWD.  The dataset is randomly split into train, val and test sets, which contain 2722, 500 and 1600 images respectively. 
 
 Following the existing works, we perform evaluation using the  standard  MAE and MSE metrics.  Furthermore, these metrics are calculated for the following sub-categories of images: \\
 (i) Low density: images containing count between 0 and 50,  \\
 (ii) Med density: images with count between 51 and 500,  \\
 (iii) High density: images with count more than 500 people,  \\
 (iv) Weather degraded images, and\\
 (v) Overall.\\
 The metrics under these sub-categories will provide  a holistic understanding of the network performance. 
 
 Fig. \ref{fig:split-density} and Fig. \ref{fig:split-weather} illustrate the distribution the number of images among the density and weather sub-categories respectively. Table \ref{tab:low-med-high} shows the distribution of images for different density-levels.

 \section{Proposed method}
 
 In this section, we present the details  of  the proposed Confidence Guided Deep Residual Crowd Counting (CG-DRCN)  along with the training and inference specifics. Fig. \ref{fig:arch} shows the architecture of the proposed network. 
 
 \subsection{Base network}
 \label{ssec:base_network}
 Following recent approaches \cite{sindagi2017generating,sam2017switching,cao2018scale}, we perform counting based on the density estimation framework. In this framework, the network is trained to estimate the density map ($\hat{Y}$) from an input crowd image ($X$). The target density map  ($Y$) for training the network is generated by imposing normalized 2D Gaussian at head locations provided by the dataset annotations:  
 \begin{equation}
 \label{eq:densitymap}
 Y(x) = \sum_{{x_g \in S}}\mathcal{N}(x-x_g,\sigma),
 \end{equation}
 where, $S$ is the set of all head locations ($x_g$) in the input image and $\sigma$ is scale parameter of 2D Gaussian kernel. Due to this formulation, the density map contains per-pixel density information of the scene, which when integrated results in the  count of people in the image.

 The proposed network consists of conv1$\sim$conv5  layers ($C_1-C_5$) of the VGG16 architecture as a part of the backbone, followed by a conv block ($CB_6$) and a max-pooling layer with stride 2.  First, the input image (of size $W\x H$) is passed through $C_1-C_5$, $CB_6$ and the max pooling layer to produce the corresponding density map ($\Yh _6$) of size $\frac{W}{32}\x \frac{H}{32}$. $CB_6$ is defined by  \textit\{conv\sub{512,32,1}-relu-conv\sub{32,32,3}-relu-conv\sub{32,1,3}\}\footnote{ \label{fn:conv}\textit{conv\sub{N\sub{i},N\sub{o},k}} denotes conv layer (with \textit{N\sub{i}} input channels, \textit{N\sub{o}} output channels, \textit{k}$\times$\textit{k} filter size), \textit{relu} denotes ReLU activation}). Due to its low resolution, ($\Yh _6$)  can be considered as a coarse estimation, and learning this will implicitly incorporate global context in the image due the large receptive field at the deepest layer in the network.

 \subsection{Residual learning}
 \label{ssec:residual_learning}
 Although $\Yh _6$ provides a good estimate of the number of people in the image, the density map lacks several local details as shown in Fig. \ref{fig:coarse_to_fine} (a).  This is because deeper layers learn to capture abstract concepts and tend to lose low level details in the image. On the other hand, the shallower layers have relatively more detailed local information as compared to their deeper counterparts \cite{ranjan2016hyperface}. Based on this observation, we  propose to refine the coarser density maps by employing shallower layers in a residual learning framework. This refinement mechanism is inspired in part by several leading work on super-resolution \cite{tai2017image,kim2016accurate,lim2017enhanced} that incorporate residual learning to learn finer details required to generate a high quality  super-resolved image. Specifically, features from $C_5$ are forwarded through a uncertainty guided residual estimation block (\UREB$_5$ to generate a residual map $\hat{R}_5$, which is then added to an appropriately up-sampled version of $\Yh_6$ to produce the density map $\Yh_5$ of size $\frac{W}{16}\x \frac{H}{16}$, \ie
 \begin{equation}
 \Yh_5 = \hat{R}_5 + up(\Yh_6).
 \end{equation}
 Here, $up()$ denotes up-sampling by a factor of 2$\x$ via bilinear interpolation. By enforcing \UREB$_5$ to learn a residual map, the network  focuses on the local errors emanating from  the deeper layer, resulting in better learning of the offsets required to refined the coarser density map. \UREB\; is described in Section \ref{ssec:ceb}.
 
 The above refinement is further repeated to recursively generate  finer density maps $\Yh_4$ and $\Yh_3$ using the feature maps from the shallower layers $C_4$ and $C_3$, respectively. Specifically, the output of  $C_4$ and $C_3$ are forwarded through \UREB$_4$, \UREB$_3$  to learn residual maps $\hat{R}_4$ and $\hat{R}_3$, which are then added to the appropriately up-sampled versions of the coarser maps $\Yh_5$ and $\Yh_4$ to produce $\Yh_4$ and $\Yh_3$ respectively in that order. 
 Specifically, $\Yh_4$ and $\Yh_3$ are obtained as follows: 
 
 \begin{equation}
 \Yh_4 = \hat{R}_4 + up(\Yh_5), \quad\quad
 \Yh_3 = \hat{R}_3 + up(\Yh_4)
 \end{equation}


 \subsection{Uncertainty guided residual learning (\UREB)}
 \label{ssec:ceb}
 
 In this section, we provide a detailed description of the uncertainty guided residual estimation block (\UREB\;)  that is used to refine the residual estimation process. Specifically, 
 features ($F_i$)  from the main branch are forwarded through a conv block ($CB_i$) which estimates the residual map $R_i$.  In order to improve the efficacy of the residual learning mechanism, we propose an uncertainty guided confidence estimation  block ($CEB$) that guides the refinement process. The task of conv blocks $CB_i$ is to capture residual errors that can be incorporated into the coarser density maps to produce high quality density maps in the end. For this purpose, these conv blocks employ feature maps from shallower conv layers $C_i$ from the main branch.
 
 \begin{figure}[htp!]
 	\begin{center}
 		\includegraphics[width=1\linewidth]{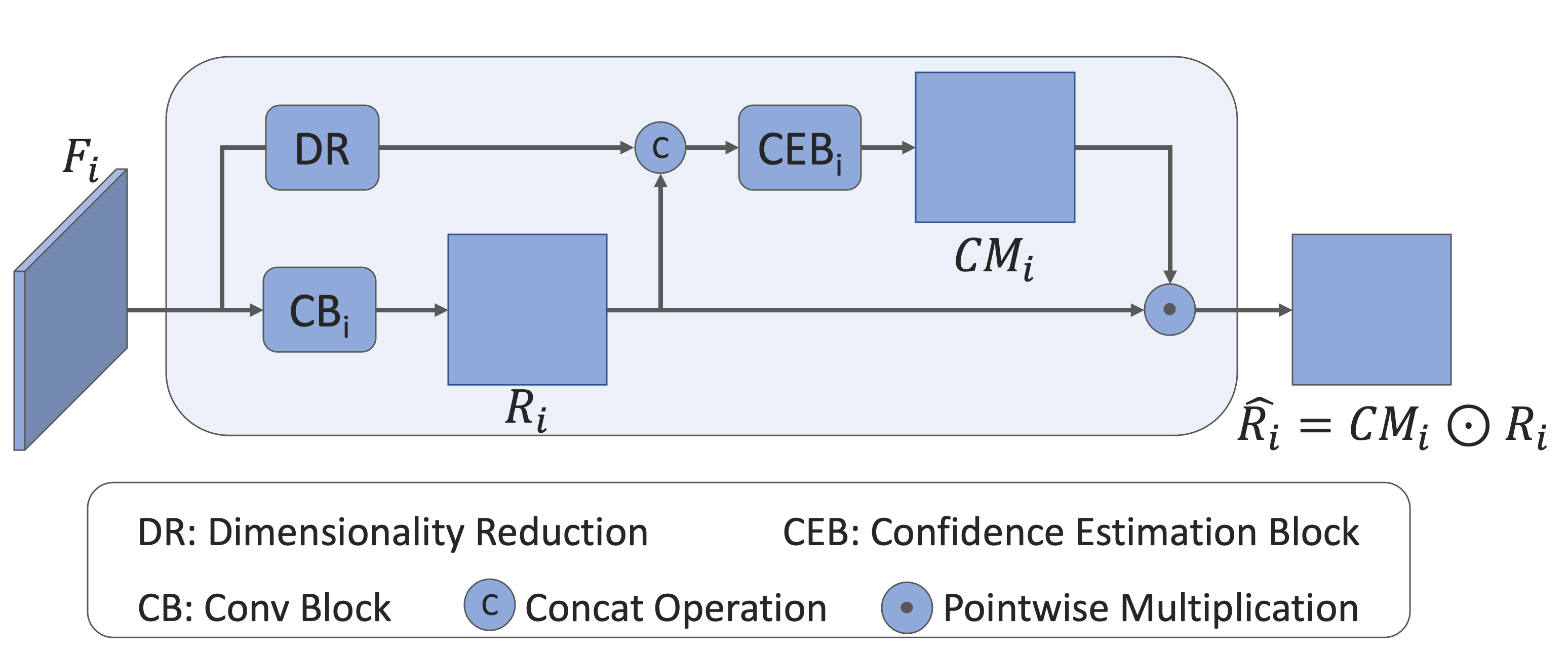}
 	\end{center}
 	\vskip -10pt \caption{Uncertainty-guided residual estimation block (\UREB).}
 	\label{fig:ceb}
 \end{figure}
 
 \begin{figure*}[htp!]
 	\begin{center}
 		\includegraphics[width=0.195\linewidth]{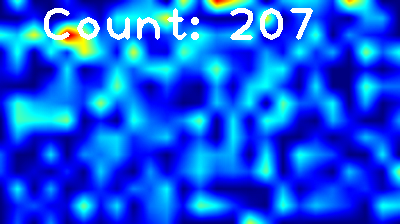}
 		\includegraphics[width=0.195\linewidth]{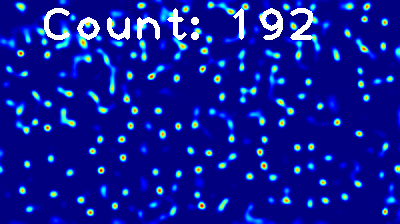}
 		\includegraphics[width=0.195\linewidth]{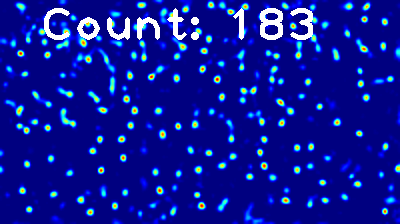}
 		\includegraphics[width=0.195\linewidth]{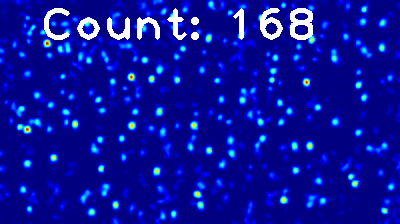}
 		\includegraphics[width=0.195\linewidth]{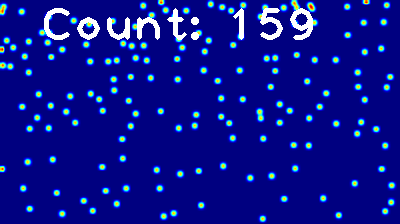}\\
 		(a)\hskip 90pt(b)\hskip 90pt(c)\hskip 90pt(d)\hskip 90pt(e)  
 	\end{center}
 	\vskip -15pt \caption{Density maps estimated by different layers of the proposed network. (a) $\Yh_6$ (b) $\Yh_5$  (c) $\Yh_4$  (d) $\Yh_3$   (e) $Y $(ground-truth). It can be observed that the output of the deepest layer ($\Yh_6$) looks very coarse, and it is refined in a progressive manner using the residual learned by  \UREB$_5$, \UREB$_4$, \UREB$_3$ to obtain the $\Yh_5,\Yh_4, \Yh_3$ respectively. Note that fine details and the total count in the density maps improve as we  move from $\Yh_6$  to $\Yh_3$.}
 	\label{fig:coarse_to_fine}
 \end{figure*}
 
 Since the conv layers  in the  main branch are  primarily trained for estimating the coarsest density map, their features have high responses in regions where crowd is present, and hence, they may  not necessarily produce effective residuals. In order to overcome this issue, we propose to gate the residuals that are not effective. This is achieved by  using uncertainty estimation. Inspired by uncertainty estimation in CNNs \cite{kendall2017uncertainties,zhu2017deep,devries2018learning,Yasarla_2019_CVPR}, we aim to model pixel-wise aleatoric uncertainty of the residuals  estimated by $CB_i$.  That is we, predict the pixel-wise confidence (inverse of the uncertainties) of the residuals which are then     used to gate the residuals before being passed on to the subsequent outputs. This ensures that only highly confident residuals get propagated to the output. Note that $CB_5, CB_4, CB_3$ are defined as follows:\\

 \noindent$CB_5$:  \textit\{conv\sub{512,32,1}-relu-conv\sub{32,32,3}-relu-conv\sub{32,1,3}\}\textsuperscript{\ref{fn:conv}}. \\
 $CB_4$:  \textit\{conv\sub{512,32,1}-relu-conv\sub{32,32,3}-relu-conv\sub{32,1,3}\}\textsuperscript{\ref{fn:conv}}. \\
 $CB_3$:  \textit\{conv\sub{256,32,1}-relu-conv\sub{32,32,3}-relu-conv\sub{32,1,3}\}\textsuperscript{\ref{fn:conv}}.\\

 In terms of the overall architecture, we introduce a set of \UREB s as shown in Fig. \ref{fig:arch}. Each residual branch consists of one such block.  Fig. \ref{fig:ceb} illustrates the mechanism of the proposed \UREB.  UREB$_i$ takes the feature map $F_i$ from the main branch and forwards them through a conv block $CB_i$ to produce residual map ($R_i$). This residual map is then concatenated with  dimensionality reduced features\footnote{We use a dimensionality reduction (DR) block which consists of $1\times1$ conv layer to reduce the number of channels to 32.} from the main branch and forwarded through confidence estimation block ($CEB_i$). This block is  defined by  \textit\{conv\sub{33,32,1}-relu-conv\sub{32,16,3}-relu-conv\sub{16,16,3}-relu-conv\sub{16,1,1}\} and it produces a confidence map $CM_i$ which is then multiplied element-wise with the input to form the refined residual map: 
 
 \begin{equation}
 \hat{R}_i = R_i\odot CM_i,
 \end{equation}
 where  $\odot$ denotes element-wise multiplication. 
 
 \begin{figure}[b!]
 	\begin{center}
 		
 		\includegraphics[width=0.32\linewidth]{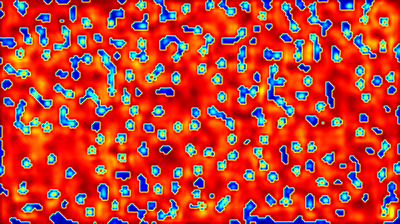}
 		\includegraphics[width=0.32\linewidth]{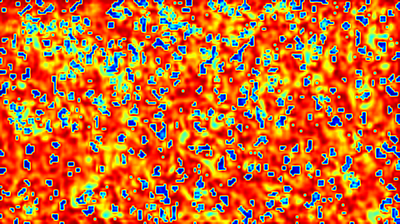}
 		\includegraphics[width=0.32\linewidth]{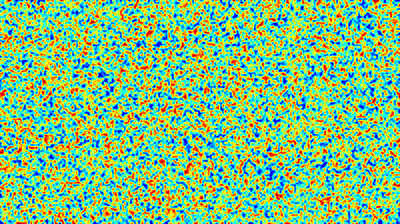}\\
 		\vspace{2pt}
 		\includegraphics[width=0.32\linewidth]{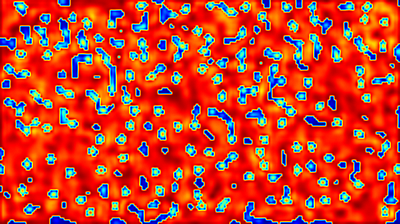}
 		\includegraphics[width=0.32\linewidth]{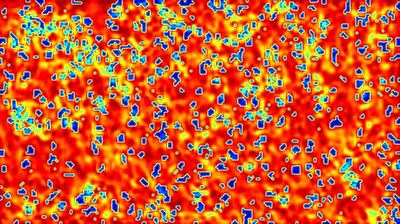}
 		\includegraphics[width=0.32\linewidth]{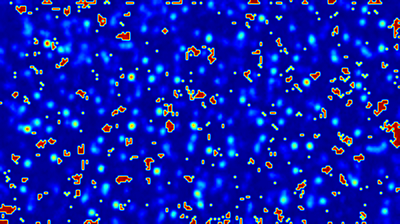}\\
 		(a) \hskip 60pt (b) \hskip 60pt (c)
 	\end{center}
 	\vskip -10pt \caption{Residual maps. \textit{Top row}: Without confidence gating. \textit{Bottom row}: With confidence gating. (a) $R_5$ (b) $R_4$ (c) $R_3$. Red indicates negative values and cyan indicates positive values. The use of confidence gating  improves the residual maps significantly, especially for the shallower layers. }
 	\label{fig:residuals}
 \end{figure}

 In order to learn these confidence maps, the loss function $L_f$ used to train the network is defined as follows,
 \begin{equation}
 \label{eq:lossnew}
 L_f = L_d - \lambda_c L_c,
 \end{equation} 
 where, $\lambda_{c}$ is a regularization constant, $L_d$ is the pixel-wise regression loss to minimize the density map prediction error and is defined as:
 \begin{equation}
 \label{eq:lossdensity}
 L_d = \sum_{i\in\{3,4,5,6\}}\| (CM_i \odot Y_i) - (CM_i \odot \Yh_i)\|_2,
 \end{equation}
 where,  $\Yh_i$ is the predicted density map, $i$ indicates the index of the conv layer from which the predicted density map is taken,  $Y_i$ is the corresponding target. 
 
 $L_c$ is the confidence guiding loss, defined as,
 \begin{equation}
  \label{eq:lossconf}
 L_c = \sum_{i\in\{3,4,5,6\}} \sum_{j=1}^{H} \sum_{k=1}^{W}  \log(CM_i^{j,k}), 
 \end{equation} 
 where, $W\times H$ is the dimension of the confidence map $CM_i$. As it can be seen from  Eq. \eqref{eq:lossnew}, the loss $L_f$ has two parts $L_d$ and $L_c$. The first term minimizes the Euclidean distance between the prediction and target features, whereas $L_c$ maximizes the confidence scores $CM_i$ by making them closer to 1.

 Note that optimizing $L_c$ (Equation \ref{eq:lossconf}) forces the values of confidence map close to 1. This is true when the loss function is Equation \ref{eq:lossconf} alone. However, our final loss function (Equation \ref{eq:lossnew}) contains both $L_d$ and $L_c$. When we minimize Equation \ref{eq:lossnew}, if $CM_i$ is driven to ``one" due to loss $L_c$, the first loss in Equation \ref{eq:lossdensity} ($L_d$) increases. Due to this, the network tries to reduce $L_d$ by forcing the network to  predict lower values of $CM_i$ in addition to learning to predict accurate density maps. Hence, due to the presence of $CM_i$ in both the parts of the loss function $L_f$ (Equation \ref{eq:lossnew}), the network ``does not'' learn to trivially predict ones for $CM_i$. This is analogous to the use of aleatoric uncertainty prediction for improving the network performance described in \cite{kendall2017uncertainties}. 
 
 Fig. \ref{fig:coarse_to_fine} illustrates the output density maps ($\Yh_6,\Yh_5,\Yh_4,\Yh_3$) generated by the proposed network for a sample crowd image. It can be observed that the density maps progressively improve in terms of fine details and the count value. 
 
 Fig. \ref{fig:residuals} illustrates the residual maps generated with and without the confidence gating.  It can be clearly observed that the use of confidence scores aids in better feature learning.

 \subsection{Class-conditioned Uncertainty guided residual learning (\UREBC)}
 \label{ssec:cceb}
 
\begin{figure}[h!]
 	\begin{center}
 		\includegraphics[width=1\linewidth]{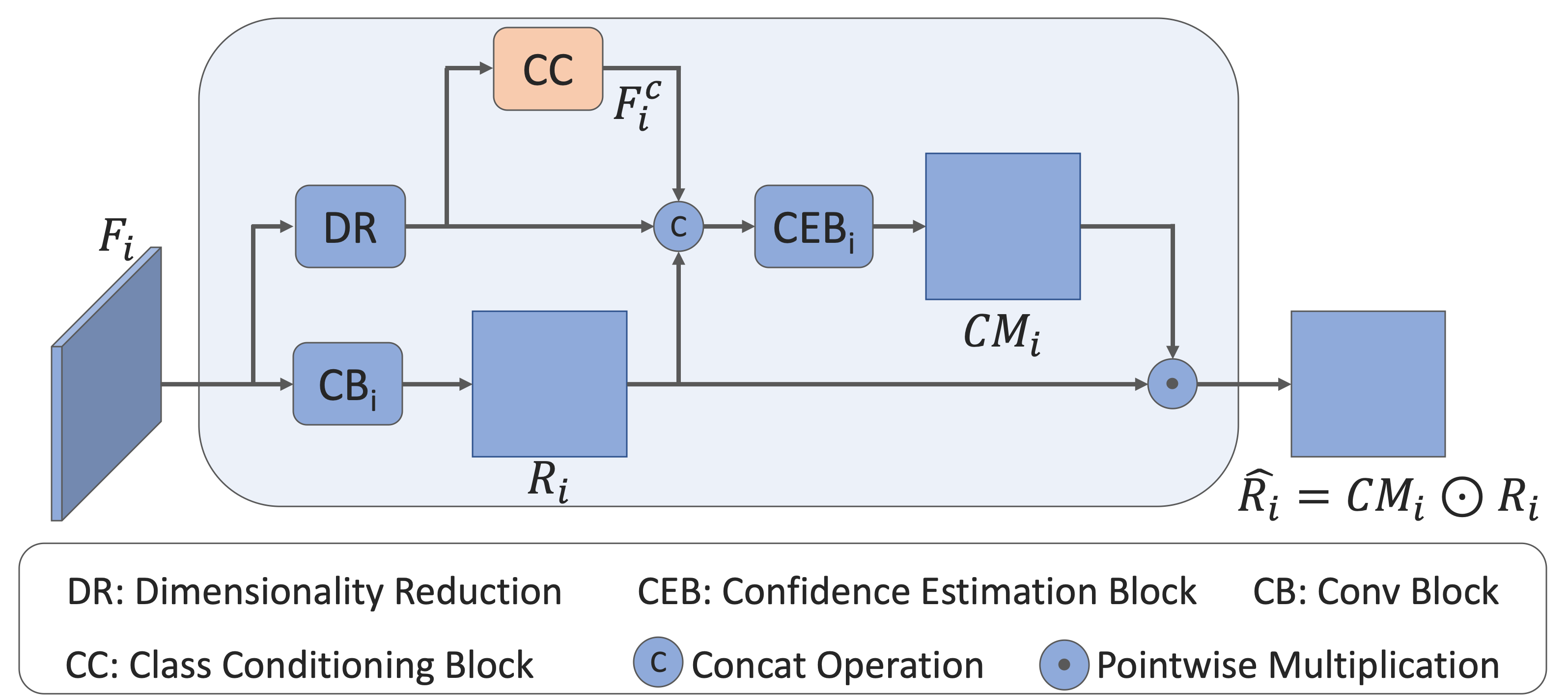}
 	\end{center}
 	\vskip -10pt \caption{Class-conditioned uncertainty-guided residual estimation block (\UREBC).}
 	\label{fig:cceb}
 \end{figure}

 In order to leverage additional information provided in the proposed JHU-CROWD++ dataset, we propose to condition the residual estimation based on the image-level labels (specifically, weather labels). That is, we augment the \UREB\; module with additional class conditioning ($CC$)  block as shown in Fig. \ref{fig:cceb}. This block consists of a set of 2 conv relu-layers ( \textit\{conv\sub{32,32,3}-relu-conv\sub{32,4,3}\}\textsuperscript{\ref{fn:conv}}) followed by an average-pool layer and a soft-max layer. Note that the output of this block is 4 classes corresponding to \textit{rain}, \textit{fog}, \textit{haze} and \textit{normal}. The $CC$ block is trained via cross-entropy error using labels available in the dataset. 
 To condition the uncertainty estimation on the classes, the feature maps ($F_i^c$)  prior to the average-pool layer in $CC$ are concatenated with the residual map $R_i$ and the dimensionality reduced features from the main branch. These concatenated feature maps are then forwarded through the confidence estimation block $CEB_i$ to predict the confidences as described earlier in Section \ref{ssec:ceb}.
 
 For training the network, we modify the loss function in Eq. \ref{eq:lossnew} as follows:
 \begin{equation}
 \label{eq:lossnew_weather}
 L_f = L_d - \lambda_c L_c + \lambda_w L_w,
 \end{equation} 
 
 where, $L_w$ is the cross-entropy loss for the weather classification and $\lambda_w$ is a weighting factor and we set it to 0.01. Note that the distribution of weather images is imbalanced. Hence, we weight the each class proportionately  based on the number of samples in each category.

 \subsection{Training and inference details}
 \label{ssec:training_details}
 \noindent The training dataset is obtained by cropping patches from multiple random locations in each training image. The cropped patch-size is 256$\times$256. For JHU-CROWD++, we use the validation set for model selection and hyper-parameter tuning. For other datasets, we use 10\% of the training images as validation set. We use the Adam optimizer to train the network. We use a learning rate of 0.00001 and a momentum of 0.9 with a batch-size of 24.  Before cropping, we resize all the images such that the minimum dimension is 512 and maximum dimension is 2048 while maintaining the aspect ratio.


 For inference, the density map $\Yh_3$ is considered as the final output. 
 The count performance is measured using the standard error metrics: mean absolute error ($MAE$) and mean squared error ($MSE$). These metrics are defined   as follows: 
 \begin{equation}
 \begin{aligned}
 MAE = \frac{1}{N}\sum_{i=1}^{N}|C(Y^i)-C(\hat{Y}_3^i)|,   \\ 
 MSE = \sqrt{\frac{1}{N}\sum_{i=1}^{N}|C(Y^i)-C(\hat{Y}_3^i)|^2},
 \end{aligned}
 \end{equation} 
 where  $N$ is the number of test samples, $Y^i$ is the ground-truth count and $\hat{Y}_3^i$ is the estimated count corresponding to the $i^{th}$ sample, and $C(Y)$ denotes the sum of all the values in $Y$.

 \section{Ablation Study}
 
 In this section, we  discuss  the results of different ablation studies conducted to analyze (i)  the effect of different components in the proposed network, (ii) generalizability to other network architectures,  (iii) the effect of different branches in the proposed architecture for residual estimation, and (iv) effect of reduced data during training.   Due to the presence of various complexities  such as high density crowds, large variations in scales,   occlusion, \etc. we choose to perform the ablation study on JHU-CROWD++ val set. 
 

 \subsection{Residual Learning, Uncertainty and Class-conditioning}
 

 The ablation study consisted of evaluating the following configurations of the  proposed method: \\
 (i) \textit{Base network}: VGG16 network with an additional conv block ($CB_6$) at the end.\\
 (i) \textit{Base network + R}: the base network with residual learning.\\
 (iii) \textit{Base network + R + \UREB\; ($\lambda_c=0$)}: the base network with residual learning guided by the confidence estimation blocks as discussed in Section \ref{ssec:ceb}. In this configuration, we aim to measure the performance due to the addition of the confidence estimation blocks without the uncertainty estimation mechanism by setting  $\lambda_c$ is set to $0$.\\
 (iv) \textit{Base network + R + \UREB\; ($\lambda_c=1$)}: the base network with residual learning guided by the confidence estimation blocks as discussed in Section \ref{ssec:ceb}.\\ 
 (v) \textit{Base network + R + \UREBC\; ($\lambda_w=0$)}: the base network with residual learning guided by the class-conditioned confidence estimation blocks as discussed in Section \ref{ssec:cceb}. In this configuration, we aim to measure the performance due to the addition of  conv block in the class conditioning $CC$  module without image-level  training by setting  $\lambda_w$ is set to $0$. \\
 (vi) \textit{Base network + R + \UREBC\; ($\lambda_w=0.01$)}: the base network with residual learning guided by the class-conditioned  confidence estimation blocks as discussed in Section \ref{ssec:cceb}. \\
 
 The results of these experiments are shown in Table \ref{tab:ablation}. It can be seen that  there are considerable improvements in the performance due to the inclusion of residual learning into the network. The use of confidence-based weighting of the residuals  results in further improvements, thus highlighting its significance in improving the efficacy of uncertainty-based residual learning.

 \begin{table}[h!]
 	\centering
 	\caption{Results of ablation study using \textbf{``VGG16''} base network on the JHU-CROWD++  dataset (val-set).}
 	\label{tab:ablation}
 	\vskip -5pt 
 	\resizebox{0.9\linewidth}{!}{
 		\begin{tabular}{l|cc}
 			\hline 
 			Method & MAE & MSE \\			\hline\hline
 			Base network & 81.1 & 300.5 \\
 			Base network + R  & 77.5 & 290.6 \\
 			Base network + R + UREB ($\lambda_c=0$)    & 77.1 & 290.5 \\
 			Base network + R + UREB ($\lambda_c=1$)   & 74.1 & 275.5 \\
 			Base network + R + UREB-C ($\lambda_w=0$)    & 74.6 & 274.1 \\
 			Base network + R + UREB-C ($\lambda_w=0.01$)   & 67.9 & 262.1 \\
 			\hline
 			
 		\end{tabular}
 	}
 	\vskip-10pt
 \end{table} 

  Further, we perform additional ablation studies to demonstrate the effectiveness of  conditioning the estimation based on the class labels. Specifically, we conduct the following three experiments (see Table \ref{tab:ablation_weather}): \\
(i) W/o conditioning: This is the baseline experiment that does not involve any use of the weather labels, \\
(ii) With multi-task learning: Here, we use the weather labels to train the $CC$ block. However, the feature maps from the $CC$ block ``are not'' concatenated with the residual map $R_i$. This experiment demonstrates the benefit of using 
the class labels in a naive manner. \\
(iii) With multi-task learning \& conditioning: This corresponds to the proposed method of conditioning the residual estimation on the class labels (as described in \ref{ssec:cceb}).  It can be observed that conditioning  results in better improvements as compared to the naive use class labels. This leads to  significant  improvements in the overall error. 

\begin{figure*}[htp!]
 	\begin{center}
	\includegraphics[width=1\linewidth, height=0.8\linewidth]{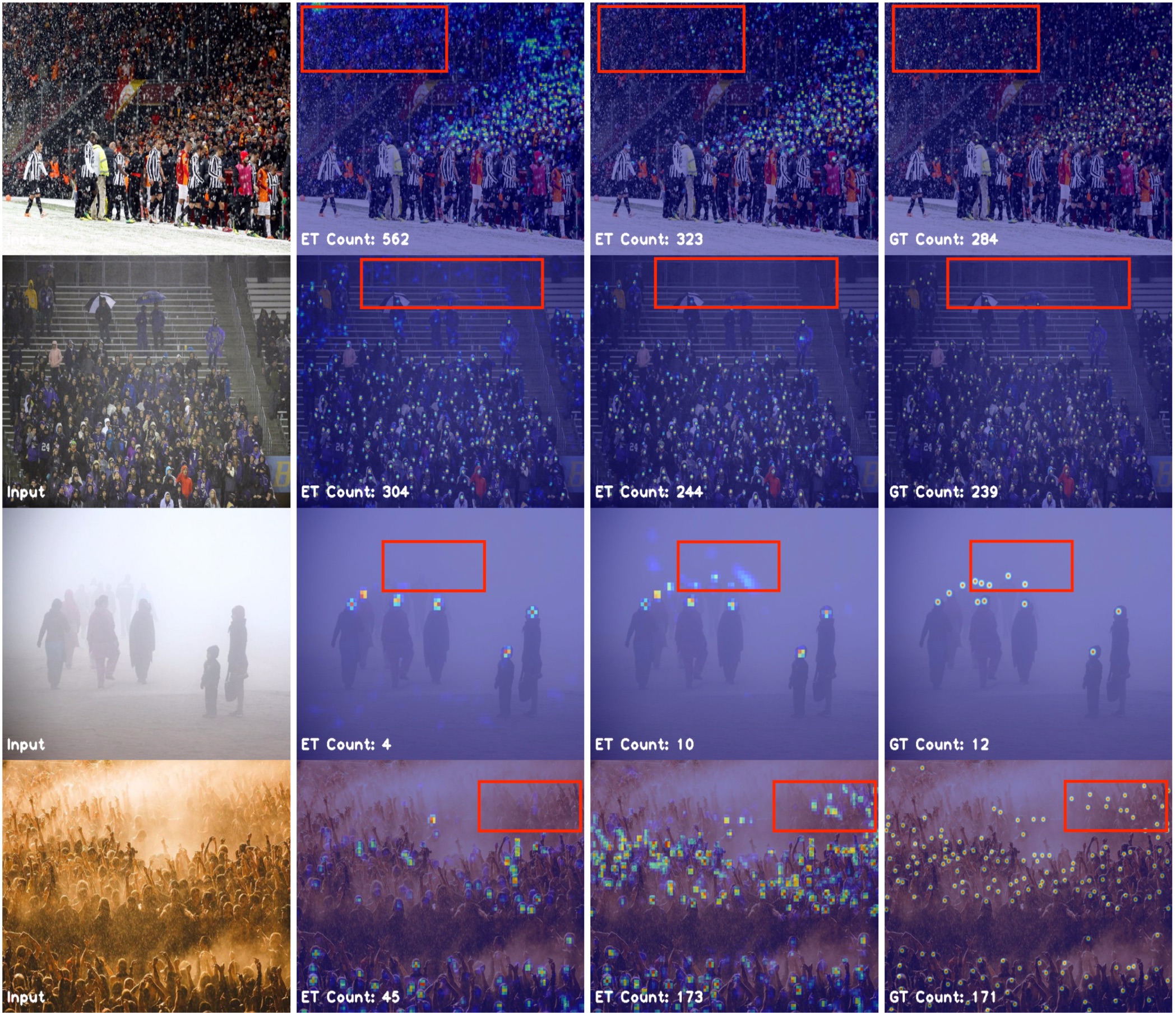}
 	\end{center}
 	(a) \hskip 120pt (b) \hskip 120pt (c) \hskip 110pt (d)  \\
 	 \caption{Ablation results: \textbf{``Class-conditioning''} for weather-conditions study on the JHU-CROWD++ weather dataset (val-set). (a) Input (b) Without class-conditioning (c) With class-conditioning (d) Ground truth. Class-conditioning improves the quality of density maps and prediction error. Boxes marked in red highlight the regions where class-conditioning improves the quality. }
 	\label{fig:ablation_weather}
 \end{figure*}

Fig. \ref{fig:ablation_weather}  visualizes the density map estimation results from the above experiments. It can be observed that  class-conditioning improves the quality of density maps and prediction error.	
 
 \begin{table}[h!]
 	\centering
 	\caption{Ablation results: \textbf{``Class-conditioning''} for weather-conditions study on the JHU-CROWD++ weather dataset (val-set).}
 	\label{tab:ablation_weather}
 	\vskip -5pt 
 	\resizebox{0.9\linewidth}{!}{
 		\begin{tabular}{l|cc}
 			\hline 
 			Method & MAE & MSE \\			\hline\hline
 			W/o conditioning & 78.4 & 170.5 \\
 			With multi-task learning & 74.4 & 140.1\\
 			With multi-task learning \& conditioning   & 63.6 & 116.6 \\
 			
 			\hline
 			
 		\end{tabular}
 	
 	}
  	\vskip -10pt
 \end{table}

 \subsection{Res101 backbone network}
 
 In order to  demonstrate that the proposed uncertainty-guided residual learning mechanism is not network-dependent, we evaluate the method using a different base network: Res101 \cite{he2016deep}. To employ the Res101 architecture as the base network: we  (i)  add the uncertainty-based residual estimation blocks $U$-$REB_3$, $U$-$REB_4$ and $U$-$REB_5$ after layers $2$, $3$ and $4$  in Res101 respectively, (ii) add conv6 layer after layer $5$ with the input number of channels changed appropriately to match the number of output channels of layer 4 in Res101, and (iii) change the number of input channels in the conv blocks in $U$-$REB_i$'s to match the number of output channels of the respective blocks in the main branch of Res101. Furthermore, since the $U$-$REB_3$ is added to a shallower layer, we weight the loss function corresponding to $\Yh_6$. That is, we modify Eq. \ref{eq:lossdensity} as follows:
 
 \begin{equation}
 \label{eq:lossdensity1}
 L_d = \sum_{i\in\{3,4,5,6\}}\lambda_i\| (CM_i \odot Y_i) - (CM_i \odot \Yh_i)\|_2.
 \end{equation}
 
 \noindent In the above equation, we set $\lambda_3$ = $0.1$ and $\lambda_4$ = $\lambda_5$ = $\lambda_6$ = $1$.
 
 Table \ref{tab:ablation_res101} shows the results of the proposed network using Res101 backbone network. We make similar observations as in the case of VGG16 base network. That is, the use of residual learning results in better performance compared to the base network. Further, incorporating uncertainty-guided residual estimation and class conditioning results in further improvements. From this experiment, we can observe that the proposed method can generalize to other types of network architectures.

  \begin{table}[ht!]
 	\centering
 	\caption{Results of ablation study using \textbf{``Res101''} base network on the JHU-CROWD++  dataset (val-set).}
 	\label{tab:ablation_res101}
 	\vskip -5pt 
 	\resizebox{0.9\linewidth}{!}{
 		\begin{tabular}{l|cc}
 			\hline 
 			Method & MAE & MSE \\			\hline\hline
 			Base network & 72.1 & 280.5 \\
 			Base network + R  & 68.5 & 270.9 \\
 			Base network + R + UREB ($\lambda_c=0$)    & 68.2 & 271.2 \\
 			Base network + R + UREB ($\lambda_c=1$)   & 62.5 & 258.1 \\
 			Base network + R + UREB-C ($\lambda_w=0$)    & 63.1 & 259.9 \\
 			Base network + R + UREB-C ($\lambda_w=0.01$)   & 57.6 & 244.4 \\
 			\hline
 			
 		\end{tabular}
 	}
 	\vskip-5pt
 \end{table}

 \subsection{Number of branches}
 
 Since the proposed method involves residual learning at multiple scales of the base network, we conduct a set of experiments to understand the effectiveness of using multiple scales. We evaluate for two backbone architectures: VGG16 and Res101. Specifically,  we conduct  experiments where we sequentially add the residual estimation blocks at conv5, conv4 and conv3 for VGG16 and at layer4, layer 3  and layer 2 for Res101. Table \ref{tab:ablation_branches} shows the  results of these experiments. It can be observed for both architectures that as we  add more residual estimation blocks at different layers, the errors drops by considerable margins.

 \begin{table}[t!]
 	\centering
 	\caption{Results of ablation on the \textbf{``branches''} used for density estimation on the JHU-CROWD++  dataset (val-set).}
 	\label{tab:ablation_branches}
 	\vskip -5pt 
 	\resizebox{0.85\linewidth}{!}{
 		\begin{tabular}{l|cc|cc}
 			\hline 
 			Base network                & \multicolumn{2}{c|}{VGG16}                                                               & \multicolumn{2}{c}{Res101}   \\\hline
 			Branch & MAE & MSE & MAE & MSE  \\			\hline\hline
 			$\Yh_6$ 									       & 81.1 & 300.5 & 72.1 & 280.5 \\
 			$\Yh_6 + \Yh_5$  				  		      & 72.1 & 280.1 & 60.6 & 251.4 \\
 			$\Yh_6 + \Yh_5 + \Yh_4$  				 & 70.7 & 270.5 & 58.8 & 249.4 \\
 			$\Yh_6 +  \Yh_5 + \Yh_4 + \Yh_3$   & 67.9 & 262.1 & 57.6 & 244.4 \\
 			\hline
 			
 		\end{tabular}
 	}
 	\vskip-10pt
 \end{table}

 \subsection{Effect of reduced training data}
 In order to demonstrate the need for a large dataset with high diversity, we conducted a set of experiments by varying the number of data samples. More specifically, we trained CG-DRCN-CC-Res101 with different percentages (5\%, 25\%, 50\%, 75\% and 100\%) of the JHU-CROWD++ train dataset. Note that the evaluation is performed on the val set. The results of this experiment are provided in Table \ref{tab:ablation_percentage}. As it can be observed, the error reduces as we increase the number of training data samples.
 
 \begin{table}[ht!]
 	
 	\caption{Results with varying training dataset size. MAE reduces with increase in the number of samples used for training.}
 	\label{tab:ablation_percentage}
 	\resizebox{1\linewidth}{!}{
 		\begin{tabular}{l|c|c|c|c|c}
 			\hline
 			\% of training data & 5\%   & 25\%  & 50\% & 75\% & 100\% \\ \hline
 			MAE                     & 124.3 & 105.4 & 93.1 & 78.9 & 57.6  \\ \hline
 		\end{tabular}
 	}
 \end{table}

 \begin{table*}[t!]
 	\centering
 	\caption{Results on JHU-CROWD++ dataset (\textbf{``Val Set''}). {\color[HTML]{CB0000} {\ul \textbf{RED}}} indicates best error and {\color[HTML]{3531FF} \textbf{BLUE}} indicates second-best error.}
 	\label{tab:uccd_val}
 	\vskip -5pt 
 	\resizebox{0.99\textwidth}{!}{%
 		\begin{tabular}{l|c|cc|cc|cc|cc|cc}
 			\hline
 			\multicolumn{2}{l|}{Category}          & \multicolumn{2}{c|}{Low}                                                               & \multicolumn{2}{c|}{Medium}                                                             & \multicolumn{2}{c|}{High}                                                                 & \multicolumn{2}{c|}{Weather}                                                            & \multicolumn{2}{c}{Overall}                                                             \\ \hline
 			Method      & Model            & MAE                                       & MSE                                        & MAE                                        & MSE                                        & MAE                                         & MSE                                         & MAE                                        & MSE                                        & MAE                                        & MSE                                         \\ \hline\hline
 			MCNN \cite{zhang2016single} (CVPR 16)     & Custom     & 90.6                                      & 202.9                                      & 125.3                                      & 259.5                                      & 494.9                                       & 856.0                                       & 241.1                                      & 532.2                                      & 160.6                                      & 377.7                                       \\ 
 			CMTL \cite{sindagi2017cnnbased} (AVSS 17)  & Custom        & 50.2                                      & 129.2                                      & 88.1                                       & 170.7                                      & 583.1                                       & 986.5                                       & 165.0                                      & 312.9                                      & 138.1                                      & 379.5                                       \\ \hline\hline
 			CSR-Net \cite{li2018csrnet}(CVPR 18)  & VGG16      & 22.2                                      & 40.0                                       & 49.0                                       & 99.5                                       & {\color[HTML]{3531FF} \textbf{302.5}}                                       & {\color[HTML]{3531FF} \textbf{669.5}}                                       & 83.0                                       & 168.7                                      & {\color[HTML]{3531FF} \textbf{72.2}}                                       & {\color[HTML]{3531FF} \textbf{249.9}}                                       \\ 
 			SA-Net \cite{cao2018scale}(ECCV 18)  & VGG16        & 13.6                                      & 26.8                                       & 50.4                                       & 78.0                                       & 397.8                                       & 749.2                                       & {\color[HTML]{3531FF} \textbf{72.2}}                                       & {\color[HTML]{3531FF} \textbf{126.7}}                                      & 82.1                                       & 272.6                                       \\ 
 			CACC \cite{liu2019context} (CVPR 19)    & VGG16       & 34.2                                      & 69.5                                       & 65.6                                       & 115.3                                      & 336.4                                       & {\color[HTML]{CB0000} {\ul \textbf{619.7}}} & 101.8                                      & 179.3                                      & 89.5                                       & {\color[HTML]{CB0000} {\ul \textbf{239.3}}} \\ 
 			
 			DSSI-Net \cite{liu2019crowd} (ICCV 19)  & VGG16      & 50.3                                      & 85.9                                       & 82.4                                       & 164.5                                      & 436.6                                       & 814.0                                       & 155.7                                      & 314.8                                      & 116.6                                      & 317.4                                       \\ 
 			MBTTBF \cite{sindagi2019multi} (ICCV 19)  & VGG16       & 23.3                                      & 48.5                                       & 53.2                                       & 119.9                                      & {\color[HTML]{CB0000} {\ul \textbf{294.5}}}                                      & 674.5                                       & 88.2                                       & 200.8                                      & 73.8                                       & 256.8                                       \\

 			LSC-CNN \cite{sam2019locate} (PAMI 20)    & VGG16      & {\color[HTML]{CB0000} {\ul \textbf{6.8}}} & {\color[HTML]{CB0000} {\ul \textbf{10.1}}} & {\color[HTML]{CB0000} {\ul \textbf{39.2}}}                                        & {\color[HTML]{CB0000} {\ul \textbf{64.1}}}                                       & 504.7                                       & 860.0                                       & 77.6                                       & 187.2                                      & 87.3                                       & 309.0                                       \\ 
 			CG-DRCN-CC-VGG16 (ours)  & VGG16 & {\color[HTML]{3531FF} \textbf{17.1}}                                      & {\color[HTML]{3531FF} \textbf{44.7}}                                       & {\color[HTML]{3531FF} \textbf{40.8}}                                       & {\color[HTML]{3531FF} \textbf{71.2}}      & 317.4                                       & 719.8                                       & {\color[HTML]{CB0000} {\ul \textbf{63.5}}}                                      & {\color[HTML]{CB0000} {\ul \textbf{116.6}}}                                     & {\color[HTML]{CB0000} {\ul \textbf{67.9}}}                                       & 262.1                                       \\  \hline\hline
 			SFCN \cite{wang2019learning} (CVPR 19)  & ResNet-101        & 11.8                                      & 19.8                                       & {\color[HTML]{3531FF} \textbf{39.3} }      & {\color[HTML]{3531FF} \textbf{73.4}}                                       & 297.3                                       & 679.4                                       & {\color[HTML]{CB0000} {\ul \textbf{52.3}}} & {\color[HTML]{CB0000} {\ul \textbf{93.6}}} & 62.9                                       & 247.5                                       \\ 
 			BCC \cite{ma2019bayesian}(ICCV 19)    & VGG19       & {\color[HTML]{CB0000} {\ul \textbf{6.9}}}       & {\color[HTML]{CB0000} {\ul \textbf{10.3}}}       & 39.7                                       & 85.2                                       & {\color[HTML]{3531FF} \textbf{279.8}}       & {\color[HTML]{CB0000} {\ul \textbf{620.4}}}                                      & 58.9                                       & 124.7                                      & {\color[HTML]{3531FF} \textbf{59.3}}                                       & {\color[HTML]{CB0000} \ul{\textbf{229.2}}}                                        \\ 
 			CG-DRCN-CC-Res101 (ours) & ResNet-101  & {\color[HTML]{3531FF} \textbf{11.7}}                                     & {\color[HTML]{3531FF} \textbf{24.8}}                                       & {\color[HTML]{CB0000} {\ul \textbf{35.2}}} & {\color[HTML]{CB0000} {\ul \textbf{57.5}}} & {\color[HTML]{CB0000} {\ul \textbf{273.9}}} & {\color[HTML]{3531FF} \textbf{676.8}}       & {\color[HTML]{3531FF} \textbf{54.0}}       & {\color[HTML]{3531FF} \textbf{106.8}}      & {\color[HTML]{CB0000} {\ul \textbf{57.6}}} & {\color[HTML]{3531FF} \textbf{244.4}}       \\ \hline
 		\end{tabular}
 	}
 \end{table*}

 \begin{table*}[t!]
 	\centering
 	\caption{Results on JHU-CROWD++ dataset (\textbf{``Test Set''}). {\color[HTML]{CB0000} {\ul \textbf{RED}}} indicates best error and {\color[HTML]{3531FF} \textbf{BLUE}} indicates second-best error.}
 	\label{tab:uccd_test}
 	\vskip -5pt 
 	\resizebox{0.99\textwidth}{!}{%
 		\begin{tabular}{l|c|cc|cc|cc|cc|cc}
 			\hline
 			\multicolumn{2}{l|}{Category}          & \multicolumn{2}{c|}{Low}                                                                & \multicolumn{2}{c|}{Medium}                                                             & \multicolumn{2}{c|}{High}                                                                 & \multicolumn{2}{c|}{Weather}                                                              & \multicolumn{2}{c}{Overall}                                                             \\ \hline
 			Method          &Model      & MAE                                        & MSE                                        & MAE                                        & MSE                                        & MAE                                         & MSE                                         & MAE                                         & MSE                                         & MAE                                        & MSE                                         \\ \hline\hline
 			MCNN \cite{zhang2016single} (CVPR 16)  & Custom       & 97.1                                       & 192.3                                      & 121.4                                      & 191.3                                      & 618.6                                       & 1,166.7                                     & 330.6                                       & 852.1                                       & 188.9                                      & 483.4                                       \\  
 			CMTL \cite{sindagi2017cnnbased} (AVSS 17)   & Custom      & 58.5                                       & 136.4                                      & 81.7                                       & 144.7                                      & 635.3                                       & 1,225.3                                     & 261.6                                       & 816.0                                       & 157.8                                      & 490.4                                       \\  \hline\hline
 			CSR-Net \cite{li2018csrnet} (CVPR 18)  & VGG16      & 27.1                                       & 64.9                                       & 43.9                                       & 71.2                                       & {\color[HTML]{3531FF} \textbf{356.2}}                                      & {\color[HTML]{3531FF} \textbf{784.4}}                                      & 141.4                                       & 640.1                                       & 85.9                                       & {\color[HTML]{3531FF} \textbf{309.2}}                                       \\  
 			SA-Net \cite{cao2018scale} (ECCV 18)  & VGG16       & {\color[HTML]{3531FF} \textbf{17.3}}                                       & {\color[HTML]{3531FF} \textbf{37.9}}                                      & 46.8                                       & 69.1                                       & 397.9                                       & 817.7                                       & 154.2                                       & 685.7                                       & 91.1                                       & 320.4                                       \\  
 			CACC \cite{liu2019context} (CVPR 19)  & VGG16        & 37.6                                       & 78.8                                       & 56.4                                       & 86.2                                       & 384.2                                       & 789.0                                       & 155.4                                       & {\color[HTML]{CB0000} {\ul \textbf{617.0}}}                                       & 100.1                                      & 314.0                                       \\  
 			
 			DSSI-Net \cite{liu2019crowd} (ICCV 19)  & VGG16     & 53.6                                       & 112.8                                      & 70.3                                       & 108.6                                      & 525.5                                       & 1,047.4                                     & 229.1                                       & 760.3                                       & 133.5                                      & 416.5                                       \\  
 			MBTTBF \cite{sindagi2019multi} (ICCV 19)   & VGG16     & 19.2                                       & 58.8                                       & 41.6                                       & 66.0                                       & {\color[HTML]{CB0000} {\ul \textbf{352.2}}}                                       & {\color[HTML]{CB0000} {\ul \textbf{760.4}}}                                      & {\color[HTML]{3531FF} \textbf{138.7}}                                       & {\color[HTML]{3531FF} \textbf{631.6}}                                       & {\color[HTML]{CB0000} {\ul \textbf{81.8}}}                                       & {\color[HTML]{CB0000} {\ul \textbf{299.1}}}       \\  
 			LSCCNN \cite{sam2019locate}  (PAMI 20)   & VGG16    & {\color[HTML]{CB0000} {\ul \textbf{10.6}} }      & {\color[HTML]{CB0000} {\ul \textbf{31.8}}} & {\color[HTML]{CB0000} {\ul \textbf{34.9}}}       & {\color[HTML]{CB0000} {\ul \textbf{55.6}}}                                      & 601.9                                       & 1,172.2                                     & 178.0                                       & 744.3                                       & 112.7                                      & 454.4                                       \\
 			CG-DRCN-CC-VGG16 (ours) & VGG16  & 19.5                                       & 58.7                                       & {\color[HTML]{3531FF} \textbf{38.4}}                                       & {\color[HTML]{3531FF} \textbf{62.7}}                                      & 367.3                                       & 837.5                                       & {\color[HTML]{CB0000} {\ul \textbf{138.6}}}                                      & 654.0                                       & {\color[HTML]{3531FF} \textbf{82.3}}                                       & 328.0                                       \\  \hline\hline
 			SFCN \cite{wang2019learning} (CVPR 19)   & ResNet-101       & 16.5                                       & 55.7                                       & 38.1                                       & 59.8                                       & {\color[HTML]{3531FF} \textbf{341.8}}       & {\color[HTML]{3531FF} \textbf{758.8}}       & {\color[HTML]{3531FF} \textbf{122.8}}       & {\color[HTML]{3531FF} \textbf{606.3}}       & 77.5                                       & {\color[HTML]{3531FF} \textbf{297.6 }}                                     \\  
 			BCC \cite{ma2019bayesian} (ICCV 19)    & VGG19     & {\color[HTML]{CB0000} {\ul \textbf{10.1}}} &{\color[HTML]{CB0000} {\ul \textbf{32.7}}}       & {\color[HTML]{CB0000} {\ul \textbf{34.2}}} & {\color[HTML]{3531FF} \textbf{54.5}}       & 352.0                                       & 768.7                                       & 140.1                                       & 675.7                                       & {\color[HTML]{3531FF} \textbf{75.0}}       & 299.9                                       \\  
 			CG-DRCN-CC-Res101 (ours) & ResNet-101   & {\color[HTML]{3531FF} \textbf{14.0}}                                       & {\color[HTML]{3531FF} \textbf{42.8}}                                      & {\color[HTML]{3531FF} \textbf{35.0}}                                      & {\color[HTML]{CB0000} {\ul \textbf{53.7}}} & {\color[HTML]{CB0000} {\ul \textbf{314.7}}} & {\color[HTML]{CB0000} {\ul \textbf{712.3}}} & {\color[HTML]{CB0000} {\ul \textbf{120.0}}} & {\color[HTML]{CB0000} {\ul \textbf{580.8}}} & {\color[HTML]{CB0000} {\ul \textbf{71.0}}} & {\color[HTML]{CB0000} {\ul \textbf{278.6}}} \\ \hline 
 		\end{tabular}
 	}
 \end{table*}

 \begin{figure*}[ht!]
 	\begin{center}
 		\includegraphics[width=1\linewidth]{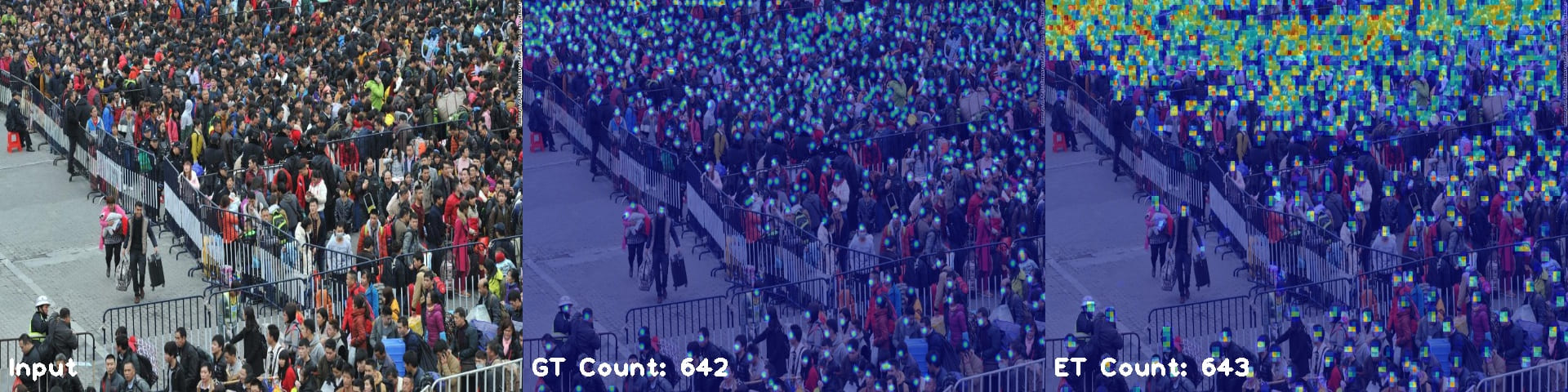}\\\vspace{2pt}
 		\includegraphics[width=1\linewidth]{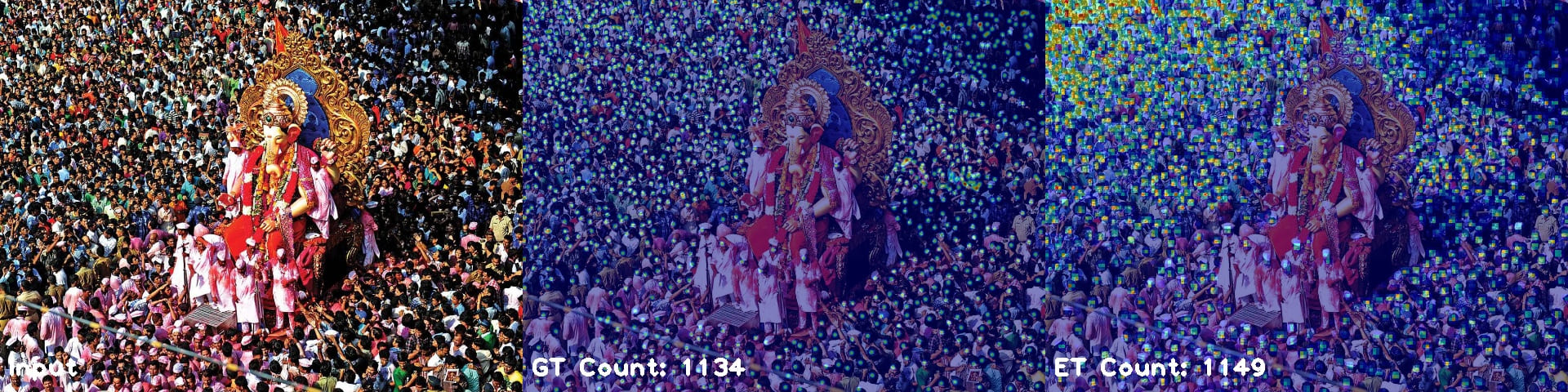}\\\vspace{2pt}
 		\includegraphics[width=1\linewidth]{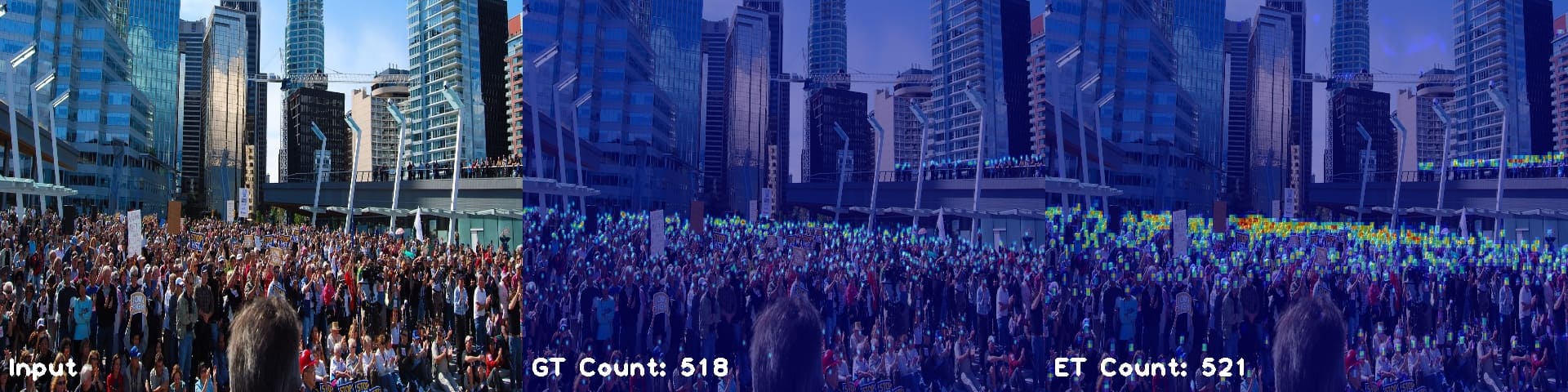}\\\vspace{2pt}
 		\includegraphics[width=1\linewidth]{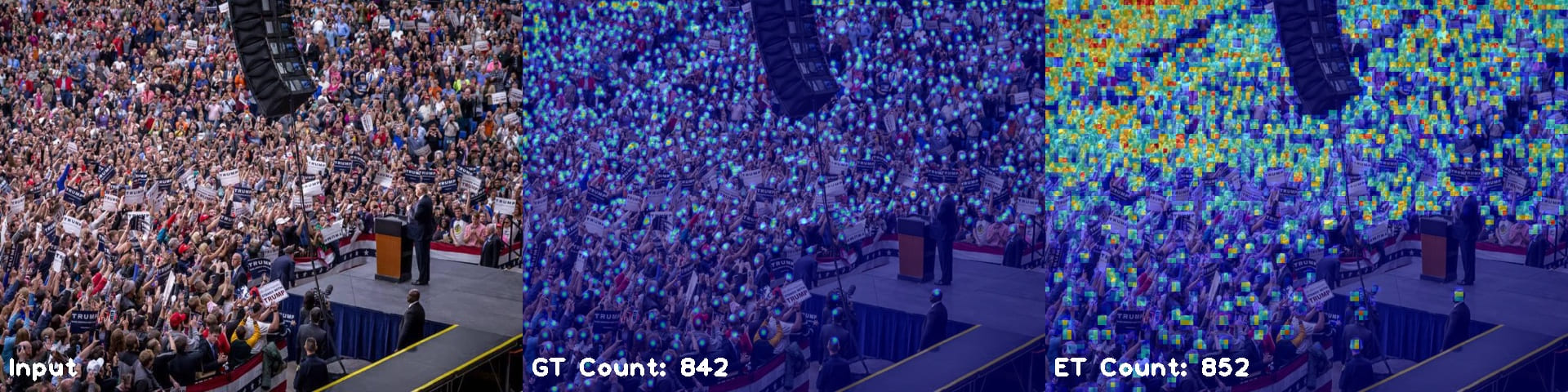}\\\vspace{2pt}
 		(a) \hskip 160pt (b) \hskip 160pt (c)   
 	\end{center}
 	\vskip -5pt \caption{Results of the proposed dataset on sample images from the JHU-CROWD++ dataset. (a) Input image (b) Ground-truth density map (c) Estimated density map.}
 	\label{fig:results_uccd}
 \end{figure*}

 \section{Benchmarking on JHU-CROWD++ dataset}

 In this section, we present results of  benchmarking of several recent algorithms including the proposed method on the JHU-CROWD++ dataset. Specifically, we evaluate the following recent works: mulit-column network (MCNN) \cite{zhang2016single}, cascaded multi-task learning for crowd counting (CMTL) \cite{sindagi2017cnnbased}, CSR-Net \cite{li2018csrnet},  SA-Net \cite{cao2018scale}, context-aware crowd counting (CACC) \cite{liu2019crowd}, spatial fully convolutional network (SFCN) \cite{wang2019learning}, deep structured scale integration network (DSSI-Net) \cite{liu2019crowd}, multi-level bottom-top and top-bottom feature fusion \cite{sindagi2019multi}, Bayesian loss for counting (BCC) \cite{ma2019bayesian} and locate-size-count-CNN (LSC-CNN) \cite{sam2019locate}. In addition,  we also evaluate the proposed class-conditioned uncertainty-guided residual estimation method (CG-DRCN-CC) and demonstrate  its effectiveness over the other methods. 
 
 All the networks are trained using the  training set. We use  the validation set  for model selection.   Table \ref{tab:uccd_val} and \ref{tab:uccd_test}  show  the results of the above experiments for various sub-categories of images. Based on these results we make the following observations:\\
 (i) The proposed method (CG-DRCN-CC) with Res101 base network achieves lowest overall MAE while obtaining comparable performance for validation set.\\
 (ii) The proposed method (CG-DRCN-CC) with Res101 base network achieves lowest overall MAE/MSE as compared to all the other methods on the test set. In addition,  it achieves best errors for the ``high-density" and ``weather'' categories while obtaining comparable performance for the rest of the categories. \\
 (iii) The proposed method (CG-DRCN-CC) with VGG16 base network achieves comparable performance in all categories with respect to the other methods.\\
 (iv) BCC \cite{ma2019bayesian} and LSC-CNN \cite{sam2019locate} achieve lowest errors in the ``low-density" categories. These methods do not follow the traditional density-estimation based approach for supervising the networks. Instead they incorporate size information during the training through  strategies like Bayesian-loss and bounding box-based supervision.\\
 (v) Res101-based methods tend to perform better compared to VGG16-based approaches in terms of overall error.

 \section{Evaluation on other datasets}
 
 In this section, we evaluate the proposed method on other datasets like ShanghaiTech \cite{zhang2016single} and UCF-QNRF \cite{idrees2018composition}. In addition, we compare the proposed method with several recent methods and demonstrate that our method is able to achieve comparable performance with respect to the state-of-the-art methods.
 
 \subsection{ShanghaiTech Dataset \cite{zhang2016single}}
 This dataset  contains 1,198 annotated images with a total of 330,165 people. This dataset consists of two parts: Part A with 482 images and Part B with 716 images. Both parts are further divided into training and test datasets with training set of Part A containing 300 images and that of Part B containing 400 images. Rest of the images are used as test set.

 The proposed network is trained on the train splits using the same strategy as discussed in Section \ref{ssec:training_details}.  Table \ref{tab:shtech} shows the results of the proposed method on ShanghaiTech  as compared with several recent approaches:  CP-CNN\cite{sindagi2017generating}, IG-CNN \cite{babu2018divide}, D-ConvNet \cite{shi2018crowd_negative}, Liu \etal \cite{liu2018leveraging}, CSR-Net \cite{li2018csrnet},  ic-CNN \cite{ranjan2018iterative}, SA-Net\cite{cao2018scale}, ACSCP \cite{shen2018adversarial} and Jian \etal \cite{jiang2019crowd}, CA-Net \cite{liu2019crowd}, BCC \cite{ma2019bayesian}, DSSI-Net \cite{liu2019crowd}, MBTTBF \cite{sindagi2019multi} and LSC-CNN \cite{sam2019locate}.  It can be observed that the proposed method outperforms all existing methods on Part A of the dataset, while achieving comparable performance on Part B. 
 
 \begin{table}[htp!]
 	\centering
 	\caption{Results on \textbf{``ShanghaiTech''} dataset \cite{zhang2016single}.}
 	\vskip-5pt
 	\label{tab:shtech}
 	\resizebox{0.9\linewidth}{!}{
 		\begin{tabular}{l|cc|cc}
 			\hline
 			& \multicolumn{2}{c|}{Part-A} & \multicolumn{2}{c}{Part-B} \\ \hline
 			Method          & MAE          & MSE          & MAE          & MSE          \\ \hline\hline
 			CP-CNN \cite{sindagi2017generating}           & 73.6        & 106.4        & 20.1         & 30.1         \\ 
 			IG-CNN \cite{babu2018divide}           & 72.5        & 118.2        & 13.6         & 21.1          \\ 
 			Liu \etal \cite{liu2018leveraging}           & 73.6        & 112.0        & 13.7         & 21.4        \\
 			D-ConvNet \cite{shi2018crowd_negative}           & 73.5        & 112.3        & 18.7         & 26.0           \\ 
 			CSRNet \cite{li2018csrnet}           & 68.2        & 115.0        & 10.6         & 16.0    \\
 			ic-CNN \cite{ranjan2018iterative}           & 69.8        & 117.3        & 10.7         & 16.0       \\
 			SA-Net  \cite{cao2018scale}           & {67.0}        & {104.5}        &  {{8.4}}         &  {{13.6}}       \\
 			ACSCP  \cite{shen2018adversarial}           & 75.7        & 102.7        & 17.2         & 27.4       \\
 			Jian \etal  \cite{jiang2019crowd} & {{64.2}}        & {{109.1}}        & {{8.2}}         & {{12.8}}\\
 			CA-Net    \cite{liu2019context} & {{61.3}}        & {{100.0}}        & {{7.8}}         & {{12.2}}\\
 			BCC    \cite{ma2019bayesian} & {{62.8}}        & {{117.0}}        & {{8.1}}         & {{12.7}}\\
 			DSSI-Net    \cite{liu2019crowd} & {{60.6}}        & {{96.0}}        & {\color[HTML]{CB0000} {\ul \textbf{6.8}}}         & {\color[HTML]{CB0000} {\ul \textbf{10.3}}}\\
 			MBTTBF    \cite{sindagi2019multi} & {\color[HTML]{3531FF} \textbf{60.2}}        & {\color[HTML]{3531FF} \textbf{94.1}}        & {{8.0}}         & {{15.5}}\\
 			LSC-CNN    \cite{sam2019locate} & {{66.5}}        & {{101.8}}        & {{7.7}}         & {{12.7}}\\\hline\hline
 			CG-DRCN-VGG16 (ours) & {64.0}        & {98.4}        & {8.5}         & {14.4}         \\
 			CG-DRCN-Res101 (ours) & {\color[HTML]{CB0000} {\ul \textbf{60.2}}}        & {\color[HTML]{CB0000} {\ul \textbf{94.0}}}        & {\color[HTML]{3531FF} \textbf{7.5}}         & {\color[HTML]{3531FF} \textbf{12.1}}         \\ \hline
 		\end{tabular}
 		
 	}
 	\vskip -0pt
 	
 \end{table}
 
 \subsection{UCF-QNRF Dataset\cite{idrees2018composition}}
 UCF-QNRF  is a large crowd counting dataset with 1535 high-resolution images and 1.25 million head annotations. There are 1201 training images and 334 test images. It contains extremely congested scenes where the maximum count of an image can reach up to 12,865.
 
 Table \ref{tab:resultsucf} shows results on the  UCF-QNRF dataset.   The proposed method is compared  with the following recent methods: Idrees \etal \cite{idrees2013multi}, MCNN \cite{zhang2016single}, CMTL \cite{sindagi2017cnnbased}, Switching-CNN \cite{sam2017switching}, Idrees \etal        
 \cite{idrees2018composition}, Jian \etal \cite{jiang2019crowd}, CA-Net \cite{liu2019crowd}, BCC \cite{ma2019bayesian}, DSSI-Net \cite{liu2019crowd}, MBTTBF \cite{sindagi2019multi} and LSC-CNN \cite{sam2019locate}. It can be observed that the proposed method achieves comparable performance with respect to the recent state-of-the-art methods.

 \begin{table}[htp!]
 	\centering
 	\caption{Results on \textbf{``UCF-QNRF ''} dataset \cite{idrees2018composition}.}
 	\label{tab:resultsucf}
 	\vskip-5pt
 	\resizebox{0.65\linewidth}{!}{
 		\begin{tabular}{l|cc}
 			\hline
 			Method & MAE & MSE \\			\hline\hline
 			Idrees \etal \cite{idrees2013multi}& 315.0 & 508.0 \\
 			Zhang \etal \cite{zhang2015cross}& 277.0 & 426.0 \\
 			CMTL \etal \cite{sindagi2017cnnbased}& 252.0 & 514.0 \\		
 			Switching-CNN \cite{sam2017switching} & 228.0 & 445.0 \\
 			Idrees \etal \cite{idrees2018composition}& {132.0} & {{191.0}} \\
 			Jian \etal \cite{jiang2019crowd}  & {{113.0}} & {{188.0 }}\\
 			CA-Net  \cite{liu2019context}  & {{107.0}} & {{183.0 }}\\
 			DSSI-Net \cite{liu2019crowd}  & {{99.1}} & {{159.2 }}\\
 			MBTTBF \cite{sindagi2019multi}  & {{97.5}} & {{165.2 }}\\
 			BCC \cite{ma2019bayesian}  & {\color[HTML]{CB0000} {\ul \textbf{88.7}}} & {\color[HTML]{CB0000} {\ul \textbf{154.8}}}\\
 			LSC-CNN \cite{sam2019locate}  & {{120.5}} & {{218.2 }}\\\hline\hline
 			CG-DRCN-VGG16 (ours) & {{112.2}} & {{176.3 }}\\
 			CG-DRCN-Res101 (ours) &  {\color[HTML]{3531FF} \textbf{95.5}} &  {\color[HTML]{3531FF} \textbf{164.3}} \\
 			\hline
 		\end{tabular}
 	}
 	\vskip -0pt 
 \end{table}

 \section{Conclusions}
 In this work, we  introduce a new large scale unconstrained crowd counting dataset (JHU-CROWD++) consisting of 4,372 images with 1.51 million annotations. The new dataset is collected under a variety of conditions and includes images with weather-based degradations and other distractors. Additionally, the dataset provides a rich set of annotations such as head locations, blur-level, occlusion-level, approximate bounding boxes and other image-level labels. In addition, we benchmark several recent state-of-the-art crowd counting techniques on the new dataset. 
 
 Furthermore,  we present a novel crowd counting network that employs residual learning mechanism in a progressive fashion to estimate coarse to fine density maps. The efficacy of residual learning is further improved by introducing an uncertainty-based confidence weighting  mechanism that is designed to enable the network to propagate only high-confident residuals to the output. Additionally, we incorporate class-conditioning mechanism to leverage the image-level labels in the new dataset for improving the performance in adverse weather conditions. The proposed method is evaluated on  recent datasets and we demonstrate that it achieves comparable performance with respect to the state-of-the-art methods.

\ifCLASSOPTIONcompsoc
  \section*{Acknowledgments}
\else
  \section*{Acknowledgment}
\fi

\noindent This work was supported by the NSF grant 1910141.

We would like to express our deep gratitude to everyone who contributed to the creation of this dataset including the members of the JHU-VIU lab and the numerous Amazon Mturk workers. We would like to specially thank Kumar Siddhanth, Poojan Oza, A. N. Sindagi, Jayadev S, Supriya S,  Shruthi S and S. Sreevali for providing assistance in annotation and verification efforts.

\ifCLASSOPTIONcaptionsoff
  \newpage
\fi



%
	
\bibliographystyle{IEEEtran}
\bibliography{egbib}

\begin{thebibliography}{10}
\providecommand{\url}[1]{#1}
\csname url@samestyle\endcsname
\providecommand{\newblock}{\relax}
\providecommand{\bibinfo}[2]{#2}
\providecommand{\BIBentrySTDinterwordspacing}{\spaceskip=0pt\relax}
\providecommand{\BIBentryALTinterwordstretchfactor}{4}
\providecommand{\BIBentryALTinterwordspacing}{\spaceskip=\fontdimen2\font plus
\BIBentryALTinterwordstretchfactor\fontdimen3\font minus
  \fontdimen4\font\relax}
\providecommand{\BIBforeignlanguage}[2]{{%
\expandafter\ifx\csname l@#1\endcsname\relax
\typeout{** WARNING: IEEEtran.bst: No hyphenation pattern has been}%
\typeout{** loaded for the language `#1'. Using the pattern for}%
\typeout{** the default language instead.}%
\else
\language=\csname l@#1\endcsname
\fi
#2}}
\providecommand{\BIBdecl}{\relax}
\BIBdecl

\bibitem{li2015crowded}
T.~Li, H.~Chang, M.~Wang, B.~Ni, R.~Hong, and S.~Yan, ``Crowded scene analysis:
  A survey,'' \emph{IEEE Transactions on Circuits and Systems for Video
  Technology}, vol.~25, no.~3, pp. 367--386, 2015.

\bibitem{zhan2008crowd}
B.~Zhan, D.~N. Monekosso, P.~Remagnino, S.~A. Velastin, and L.-Q. Xu, ``Crowd
  analysis: a survey,'' \emph{Machine Vision and Applications}, vol.~19, no.
  5-6, pp. 345--357, 2008.

\bibitem{idrees2013multi}
H.~Idrees, I.~Saleemi, C.~Seibert, and M.~Shah, ``Multi-source multi-scale
  counting in extremely dense crowd images,'' in \emph{Proceedings of the IEEE
  Conference on Computer Vision and Pattern Recognition}, 2013, pp. 2547--2554.

\bibitem{zhang2015cross}
C.~Zhang, H.~Li, X.~Wang, and X.~Yang, ``Cross-scene crowd counting via deep
  convolutional neural networks,'' in \emph{Proceedings of the IEEE Conference
  on Computer Vision and Pattern Recognition}, 2015, pp. 833--841.

\bibitem{zhang2016single}
Y.~Zhang, D.~Zhou, S.~Chen, S.~Gao, and Y.~Ma, ``Single-image crowd counting
  via multi-column convolutional neural network,'' in \emph{Proceedings of the
  IEEE Conference on Computer Vision and Pattern Recognition}, 2016, pp.
  589--597.

\bibitem{sindagi2017generating}
V.~A. Sindagi and V.~M. Patel, ``Generating high-quality crowd density maps
  using contextual pyramid cnns,'' in \emph{The IEEE International Conference
  on Computer Vision (ICCV)}, Oct 2017.

\bibitem{sam2017switching}
D.~B. Sam, S.~Surya, and R.~V. Babu, ``Switching convolutional neural network
  for crowd counting,'' in \emph{Proceedings of the IEEE Conference on Computer
  Vision and Pattern Recognition}, 2017.

\bibitem{chan2008privacy}
A.~B. Chan, Z.-S.~J. Liang, and N.~Vasconcelos, ``Privacy preserving crowd
  monitoring: Counting people without people models or tracking,'' in
  \emph{Computer Vision and Pattern Recognition, 2008. CVPR 2008. IEEE
  Conference on}.\hskip 1em plus 0.5em minus 0.4em\relax IEEE, 2008, pp. 1--7.

\bibitem{rodriguez2011density}
M.~Rodriguez, I.~Laptev, J.~Sivic, and J.-Y. Audibert, ``Density-aware person
  detection and tracking in crowds,'' in \emph{2011 International Conference on
  Computer Vision}.\hskip 1em plus 0.5em minus 0.4em\relax IEEE, 2011, pp.
  2423--2430.

\bibitem{zhu2014crowd}
F.~Zhu, X.~Wang, and N.~Yu, ``Crowd tracking with dynamic evolution of group
  structures,'' in \emph{European Conference on Computer Vision}.\hskip 1em
  plus 0.5em minus 0.4em\relax Springer, 2014, pp. 139--154.

\bibitem{li2014anomaly}
W.~Li, V.~Mahadevan, and N.~Vasconcelos, ``Anomaly detection and localization
  in crowded scenes,'' \emph{IEEE transactions on pattern analysis and machine
  intelligence}, vol.~36, no.~1, pp. 18--32, 2014.

\bibitem{mahadevan2010anomaly}
V.~Mahadevan, W.~Li, V.~Bhalodia, and N.~Vasconcelos, ``Anomaly detection in
  crowded scenes.'' in \emph{CVPR}, vol. 249, 2010, p. 250.

\bibitem{marsden2018people}
M.~Marsden, K.~McGuinness, S.~Little, C.~E. Keogh, and N.~E. O'Connor,
  ``People, penguins and petri dishes: Adapting object counting models to new
  visual domains and object types without forgetting,'' in \emph{The IEEE
  Conference on Computer Vision and Pattern Recognition (CVPR)}, June 2018.

\bibitem{sindagi2019dafe}
V.~A. Sindagi and V.~M. Patel, ``Dafe-fd: Density aware feature enrichment for
  face detection,'' in \emph{2019 IEEE Winter Conference on Applications of
  Computer Vision (WACV)}.\hskip 1em plus 0.5em minus 0.4em\relax IEEE, 2019,
  pp. 2185--2195.

\bibitem{sam2019almost}
D.~B. Sam, N.~N. Sajjan, H.~Maurya, and R.~V. Babu, ``Almost unsupervised
  learning for dense crowd counting,'' in \emph{Thirty-Third AAAI Conference on
  Artificial Intelligence}, 2019.

\bibitem{chan2009bayesian}
A.~B. Chan and N.~Vasconcelos, ``Bayesian poisson regression for crowd
  counting,'' in \emph{2009 IEEE 12th International Conference on Computer
  Vision}.\hskip 1em plus 0.5em minus 0.4em\relax IEEE, 2009, pp. 545--551.

\bibitem{chen2013cumulative}
K.~Chen, S.~Gong, T.~Xiang, and C.~Change~Loy, ``Cumulative attribute space for
  age and crowd density estimation,'' in \emph{Proceedings of the IEEE
  conference on computer vision and pattern recognition}, 2013, pp. 2467--2474.

\bibitem{lu2017tasselnet}
H.~Lu, Z.~Cao, Y.~Xiao, B.~Zhuang, and C.~Shen, ``Tasselnet: Counting maize
  tassels in the wild via local counts regression network,'' \emph{Plant
  Methods}, vol.~13, no.~1, p.~79, 2017.

\bibitem{lempitsky2010learning}
V.~Lempitsky and A.~Zisserman, ``Learning to count objects in images,'' in
  \emph{Advances in Neural Information Processing Systems}, 2010, pp.
  1324--1332.

\bibitem{french2015convolutional}
G.~French, M.~Fisher, M.~Mackiewicz, and C.~Needle, ``Convolutional neural
  networks for counting fish in fisheries surveillance video,'' in
  \emph{British Machine Vision Conference Workshop}.\hskip 1em plus 0.5em minus
  0.4em\relax BMVA Press, 2015.

\bibitem{idrees2018composition}
H.~Idrees, M.~Tayyab, K.~Athrey, D.~Zhang, S.~Al-Maadeed, N.~Rajpoot, and
  M.~Shah, ``Composition loss for counting, density map estimation and
  localization in dense crowds,'' in \emph{European Conference on Computer
  Vision}.\hskip 1em plus 0.5em minus 0.4em\relax Springer, 2018, pp. 544--559.

\bibitem{walach2016learning}
E.~Walach and L.~Wolf, ``Learning to count with cnn boosting,'' in
  \emph{European Conference on Computer Vision}.\hskip 1em plus 0.5em minus
  0.4em\relax Springer, 2016, pp. 660--676.

\bibitem{onoro2016towards}
D.~Onoro-Rubio and R.~J. L{\'o}pez-Sastre, ``Towards perspective-free object
  counting with deep learning,'' in \emph{European Conference on Computer
  Vision}.\hskip 1em plus 0.5em minus 0.4em\relax Springer, 2016, pp. 615--629.

\bibitem{sindagi2017cnnbased}
V.~A. Sindagi and V.~M. Patel, ``Cnn-based cascaded multi-task learning of
  high-level prior and density estimation for crowd counting,'' in
  \emph{Advanced Video and Signal Based Surveillance (AVSS), 2017 IEEE
  International Conference on}.\hskip 1em plus 0.5em minus 0.4em\relax IEEE,
  2017.

\bibitem{ranjan2018iterative}
V.~Ranjan, H.~Le, and M.~Hoai, ``Iterative crowd counting,'' in \emph{European
  Conference on Computer Vision}.\hskip 1em plus 0.5em minus 0.4em\relax
  Springer, 2018, pp. 278--293.

\bibitem{cao2018scale}
X.~Cao, Z.~Wang, Y.~Zhao, and F.~Su, ``Scale aggregation network for accurate
  and efficient crowd counting,'' in \emph{European Conference on Computer
  Vision}.\hskip 1em plus 0.5em minus 0.4em\relax Springer, 2018, pp. 757--773.

\bibitem{sam2018top}
D.~B. Sam and R.~V. Babu, ``Top-down feedback for crowd counting convolutional
  neural network,'' in \emph{Thirty-Second AAAI Conference on Artificial
  Intelligence}, 2018.

\bibitem{zhang2016data}
C.~Zhang, K.~Kang, H.~Li, X.~Wang, R.~Xie, and X.~Yang, ``Data-driven crowd
  understanding: A baseline for a large-scale crowd dataset,'' \emph{IEEE
  Transactions on Multimedia}, vol.~18, no.~6, pp. 1048--1061, 2016.

\bibitem{wang2020nwpu}
Q.~Wang, J.~Gao, W.~Lin, and X.~Li, ``Nwpu-crowd: A large-scale benchmark for
  crowd counting,'' \emph{arXiv preprint arXiv:2001.03360}, 2020.

\bibitem{babu2018divide}
D.~Babu~Sam, N.~N. Sajjan, R.~Venkatesh~Babu, and M.~Srinivasan, ``Divide and
  grow: Capturing huge diversity in crowd images with incrementally growing
  cnn,'' in \emph{Proceedings of the IEEE Conference on Computer Vision and
  Pattern Recognition}, 2018, pp. 3618--3626.

\bibitem{sindagi2019ha}
V.~A. Sindagi and V.~M. Patel, ``Ha-ccn: Hierarchical attention-based crowd
  counting network,'' \emph{arXiv preprint arXiv:1907.10255}, 2019.

\bibitem{shen2018adversarial}
Z.~Shen, Y.~Xu, B.~Ni, M.~Wang, J.~Hu, and X.~Yang, ``Crowd counting via
  adversarial cross-scale consistency pursuit,'' in \emph{The IEEE Conference
  on Computer Vision and Pattern Recognition (CVPR)}, June 2018.

\bibitem{shi2018crowd_negative}
Z.~Shi, L.~Zhang, Y.~Liu, X.~Cao, Y.~Ye, M.-M. Cheng, and G.~Zheng, ``Crowd
  counting with deep negative correlation learning,'' in \emph{The IEEE
  Conference on Computer Vision and Pattern Recognition (CVPR)}, June 2018.

\bibitem{simonyan2014very}
K.~Simonyan and A.~Zisserman, ``Very deep convolutional networks for
  large-scale image recognition,'' in \emph{International Conference on
  Learning Representations}, 2015.

\bibitem{tai2017image}
Y.~Tai, J.~Yang, and X.~Liu, ``Image super-resolution via deep recursive
  residual network,'' in \emph{Proceedings of the IEEE Conference on Computer
  Vision and Pattern Recognition}, vol.~1, no.~2, 2017, p.~5.

\bibitem{kim2016accurate}
J.~Kim, J.~Kwon~Lee, and K.~Mu~Lee, ``Accurate image super-resolution using
  very deep convolutional networks,'' in \emph{Proceedings of the IEEE
  conference on computer vision and pattern recognition}, 2016, pp. 1646--1654.

\bibitem{lim2017enhanced}
B.~Lim, S.~Son, H.~Kim, S.~Nah, and K.~M. Lee, ``Enhanced deep residual
  networks for single image super-resolution,'' in \emph{The IEEE conference on
  computer vision and pattern recognition (CVPR) workshops}, vol.~1, no.~2,
  2017, p.~4.

\bibitem{ke2017srn}
W.~Ke, J.~Chen, J.~Jiao, G.~Zhao, and Q.~Ye, ``Srn: Side-output residual
  network for object symmetry detection in the wild,'' \emph{arXiv preprint
  arXiv:1703.02243}, 2017.

\bibitem{kendall2017uncertainties}
A.~Kendall and Y.~Gal, ``What uncertainties do we need in bayesian deep
  learning for computer vision?'' in \emph{Advances in neural information
  processing systems}, 2017, pp. 5574--5584.

\bibitem{zhu2017deep}
L.~Zhu and N.~Laptev, ``Deep and confident prediction for time series at
  uber,'' in \emph{2017 IEEE International Conference on Data Mining Workshops
  (ICDMW)}.\hskip 1em plus 0.5em minus 0.4em\relax IEEE, 2017, pp. 103--110.

\bibitem{devries2018learning}
T.~DeVries and G.~W. Taylor, ``Learning confidence for out-of-distribution
  detection in neural networks,'' \emph{arXiv preprint arXiv:1802.04865}, 2018.

\bibitem{sindagi2019pushing}
V.~A. Sindagi, R.~Yasarla, and V.~M. Patel, ``Pushing the frontiers of
  unconstrained crowd counting: New dataset and benchmark method,'' in
  \emph{Proceedings of the IEEE International Conference on Computer Vision},
  2019, pp. 1221--1231.

\bibitem{he2016deep}
K.~He, X.~Zhang, S.~Ren, and J.~Sun, ``Deep residual learning for image
  recognition,'' in \emph{Proceedings of the IEEE conference on computer vision
  and pattern recognition}, 2016, pp. 770--778.

\bibitem{wang2019learning}
Q.~Wang, J.~Gao, W.~Lin, and Y.~Yuan, ``Learning from synthetic data for crowd
  counting in the wild,'' \emph{arXiv preprint arXiv:1903.03303}, 2019.

\bibitem{zhu2017unpaired}
J.-Y. Zhu, T.~Park, P.~Isola, and A.~A. Efros, ``Unpaired image-to-image
  translation using cycle-consistent adversarial networks,'' in
  \emph{Proceedings of the IEEE international conference on computer vision},
  2017, pp. 2223--2232.

\bibitem{loy2013crowd}
C.~C. Loy, K.~Chen, S.~Gong, and T.~Xiang, ``Crowd counting and profiling:
  Methodology and evaluation,'' in \emph{Modeling, Simulation and Visual
  Analysis of Crowds}.\hskip 1em plus 0.5em minus 0.4em\relax Springer, 2013,
  pp. 347--382.

\bibitem{li2008estimating}
M.~Li, Z.~Zhang, K.~Huang, and T.~Tan, ``Estimating the number of people in
  crowded scenes by mid based foreground segmentation and head-shoulder
  detection,'' in \emph{Pattern Recognition, 2008. ICPR 2008. 19th
  International Conference on}.\hskip 1em plus 0.5em minus 0.4em\relax IEEE,
  2008, pp. 1--4.

\bibitem{ryan2009crowd}
D.~Ryan, S.~Denman, C.~Fookes, and S.~Sridharan, ``Crowd counting using
  multiple local features,'' in \emph{Digital Image Computing: Techniques and
  Applications, 2009. DICTA'09.}\hskip 1em plus 0.5em minus 0.4em\relax IEEE,
  2009, pp. 81--88.

\bibitem{chen2012feature}
K.~Chen, C.~C. Loy, S.~Gong, and T.~Xiang, ``Feature mining for localised crowd
  counting.'' in \emph{European Conference on Computer Vision}, 2012.

\bibitem{pham2015count}
V.-Q. Pham, T.~Kozakaya, O.~Yamaguchi, and R.~Okada, ``Count forest: Co-voting
  uncertain number of targets using random forest for crowd density
  estimation,'' in \emph{Proceedings of the IEEE International Conference on
  Computer Vision}, 2015, pp. 3253--3261.

\bibitem{xu2016crowd}
B.~Xu and G.~Qiu, ``Crowd density estimation based on rich features and random
  projection forest,'' in \emph{2016 IEEE Winter Conference on Applications of
  Computer Vision (WACV)}.\hskip 1em plus 0.5em minus 0.4em\relax IEEE, 2016,
  pp. 1--8.

\bibitem{wang2015deep}
C.~Wang, H.~Zhang, L.~Yang, S.~Liu, and X.~Cao, ``Deep people counting in
  extremely dense crowds,'' in \emph{Proceedings of the 23rd ACM international
  conference on Multimedia}.\hskip 1em plus 0.5em minus 0.4em\relax ACM, 2015,
  pp. 1299--1302.

\bibitem{arteta2016counting}
C.~Arteta, V.~Lempitsky, and A.~Zisserman, ``Counting in the wild,'' in
  \emph{European Conference on Computer Vision}.\hskip 1em plus 0.5em minus
  0.4em\relax Springer, 2016, pp. 483--498.

\bibitem{boominathan2016crowdnet}
L.~Boominathan, S.~S. Kruthiventi, and R.~V. Babu, ``Crowdnet: A deep
  convolutional network for dense crowd counting,'' in \emph{Proceedings of the
  2016 ACM on Multimedia Conference}.\hskip 1em plus 0.5em minus 0.4em\relax
  ACM, 2016, pp. 640--644.

\bibitem{wang2018defense}
Z.~Wang, Z.~Xiao, K.~Xie, Q.~Qiu, X.~Zhen, and X.~Cao, ``In defense of
  single-column networks for crowd counting,'' \emph{arXiv preprint
  arXiv:1808.06133}, 2018.

\bibitem{onoro2018learning}
D.~O{\~{n}}oro{-}Rubio, R.~J. L{\'{o}}pez{-}Sastre, and M.~Niepert, ``Learning
  short-cut connections for object counting,'' in \emph{British Machine Vision
  Conference 2018, {BMVC} 2018, Northumbria University, Newcastle, UK,
  September 3-6, 2018}, 2018.

\bibitem{sindagi2017survey}
V.~A. Sindagi and V.~M. Patel, ``A survey of recent advances in cnn-based
  single image crowd counting and density estimation,'' \emph{Pattern
  Recognition Letters}, 2017.

\bibitem{liu2018leveraging}
X.~Liu, J.~van~de Weijer, and A.~D. Bagdanov, ``Leveraging unlabeled data for
  crowd counting by learning to rank,'' in \emph{The IEEE Conference on
  Computer Vision and Pattern Recognition (CVPR)}, June 2018.

\bibitem{liu2018adcrowdnet}
N.~Liu, Y.~Long, C.~Zou, Q.~Niu, L.~Pan, and H.~Wu, ``Adcrowdnet: An
  attention-injective deformable convolutional network for crowd
  understanding,'' \emph{arXiv preprint arXiv:1811.11968}, 2018.

\bibitem{wan2019residual}
J.~Wan, W.~Luo, B.~Wu, A.~B. Chan, and W.~Liu, ``Residual regression with
  semantic prior for crowd counting,'' in \emph{Proceedings of the IEEE
  Conference on Computer Vision and Pattern Recognition}, 2019, pp. 4036--4045.

\bibitem{zhao2019leveraging}
M.~Zhao, J.~Zhang, C.~Zhang, and W.~Zhang, ``Leveraging heterogeneous auxiliary
  tasks to assist crowd counting,'' in \emph{Proceedings of the IEEE Conference
  on Computer Vision and Pattern Recognition}, 2019, pp. 12\,736--12\,745.

\bibitem{sindagi2019inverse}
V.~A. Sindagi and V.~M. Patel, ``Inverse attention guided deep crowd counting
  network,'' \emph{arXiv preprint}, 2019.

\bibitem{sindagi2019multi}
------, ``Multi-level bottom-top and top-bottom feature fusion for crowd
  counting,'' in \emph{Proceedings of the IEEE International Conference on
  Computer Vision}, 2019, pp. 1002--1012.

\bibitem{jiang2019crowd}
X.~Jiang, Z.~Xiao, B.~Zhang, X.~Zhen, X.~Cao, D.~Doermann, and L.~Shao, ``Crowd
  counting and density estimation by trellis encoder-decoder network,''
  \emph{arXiv preprint arXiv:1903.00853}, 2019.

\bibitem{shi2019revisiting}
M.~Shi, Z.~Yang, C.~Xu, and Q.~Chen, ``Revisiting perspective information for
  efficient crowd counting,'' in \emph{Proceedings of the IEEE Conference on
  Computer Vision and Pattern Recognition}, 2019, pp. 7279--7288.

\bibitem{liu2019context}
W.~Liu, M.~Salzmann, and P.~Fua, ``Context-aware crowd counting,'' in
  \emph{Proceedings of the IEEE Conference on Computer Vision and Pattern
  Recognition}, 2019, pp. 5099--5108.

\bibitem{zhang2019wide}
Q.~Zhang and A.~B. Chan, ``Wide-area crowd counting via ground-plane density
  maps and multi-view fusion cnns,'' in \emph{Proceedings of the IEEE
  Conference on Computer Vision and Pattern Recognition}, 2019, pp. 8297--8306.

\bibitem{wan2019adaptive}
J.~Wan and A.~Chan, ``Adaptive density map generation for crowd counting,'' in
  \emph{Proceedings of the IEEE International Conference on Computer Vision},
  2019, pp. 1130--1139.

\bibitem{sam2019locate}
D.~B. Sam, S.~V. Peri, A.~Kamath, R.~V. Babu \emph{et~al.}, ``Locate, size and
  count: Accurately resolving people in dense crowds via detection,''
  \emph{arXiv preprint arXiv:1906.07538}, 2019.

\bibitem{ma2019bayesian}
Z.~Ma, X.~Wei, X.~Hong, and Y.~Gong, ``Bayesian loss for crowd count estimation
  with point supervision,'' in \emph{Proceedings of the IEEE International
  Conference on Computer Vision}, 2019, pp. 6142--6151.

\bibitem{zhang20203d}
Q.~Zhang and A.~B. Chan, ``3d crowd counting via multi-view fusion with 3d
  gaussian kernels,'' \emph{arXiv preprint arXiv:2003.08162}.

\bibitem{gao2020cnn}
G.~Gao, J.~Gao, Q.~Liu, Q.~Wang, and Y.~Wang, ``Cnn-based density estimation
  and crowd counting: A survey,'' \emph{arXiv preprint arXiv:2003.12783}, 2020.

\bibitem{ranjan2016hyperface}
R.~{Ranjan}, V.~M. {Patel}, and R.~{Chellappa}, ``Hyperface: A deep multi-task
  learning framework for face detection, landmark localization, pose
  estimation, and gender recognition,'' \emph{IEEE Transactions on Pattern
  Analysis and Machine Intelligence}, vol.~41, no.~1, pp. 121--135, Jan 2019.

\bibitem{Yasarla_2019_CVPR}
R.~Yasarla and V.~M. Patel, ``Uncertainty guided multi-scale residual
  learning-using a cycle spinning cnn for single image de-raining,'' in
  \emph{The IEEE Conference on Computer Vision and Pattern Recognition (CVPR)},
  June 2019.

\bibitem{li2018csrnet}
Y.~Li, X.~Zhang, and D.~Chen, ``Csrnet: Dilated convolutional neural networks
  for understanding the highly congested scenes,'' in \emph{Proceedings of the
  IEEE Conference on Computer Vision and Pattern Recognition}, 2018, pp.
  1091--1100.

\bibitem{liu2019crowd}
L.~Liu, Z.~Qiu, G.~Li, S.~Liu, W.~Ouyang, and L.~Lin, ``Crowd counting with
  deep structured scale integration network,'' in \emph{Proceedings of the IEEE
  International Conference on Computer Vision}, 2019, pp. 1774--1783.

\end{thebibliography}
%
%

%

\begin{IEEEbiography}[{\includegraphics[width=1in,height=1.45in,clip,keepaspectratio]{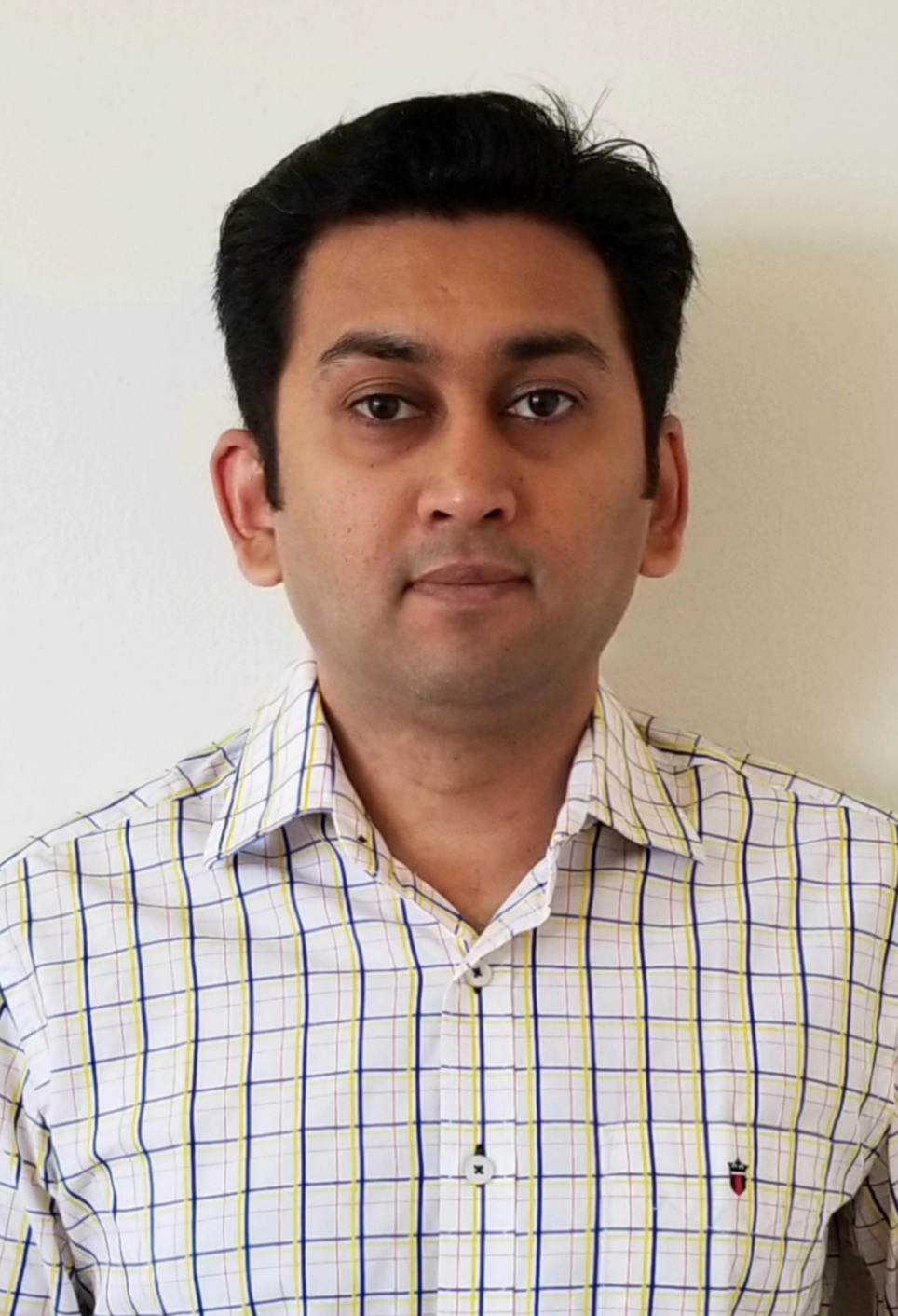}}]{Vishwanath A Sindagi }[S'16] is a PhD student in the Dept. Of Electrical \& Computer Engineering at The Johns Hopkins University.  Prior to joining Johns Hopkins, he worked for Samsung R\&D Institute-Bangalore. He graduated from IIIT-Bangalore with a Master's degree in Information Technology. His research interests include deep learning based crowd analytics, object detection, applications of generative modeling, domain adaptation and low-level vision.
\end{IEEEbiography}

\begin{IEEEbiography}[{\includegraphics[width=1in,height=1.45in,clip,keepaspectratio]{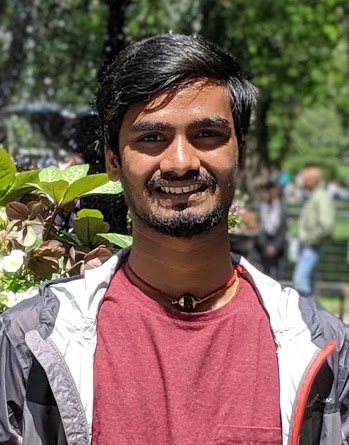}}]{Rajeev Yasarla }[S'18] is a PhD student in the Dept. Of Electrical \& Computer Engineering at The Johns Hopkins University.  Prior to joining Johns Hopkins, he worked for Qualcomm India. He graduated from IIT-Madras with a Master's degree in Communication Engineering. His research interests interests lie in the image restoration and object detection tasks. Currently, he is  working on deraining, deblurring and dehazing tasks.
\end{IEEEbiography}

\begin{IEEEbiography}[{\includegraphics[width=1in,height=1.25in,clip,keepaspectratio]{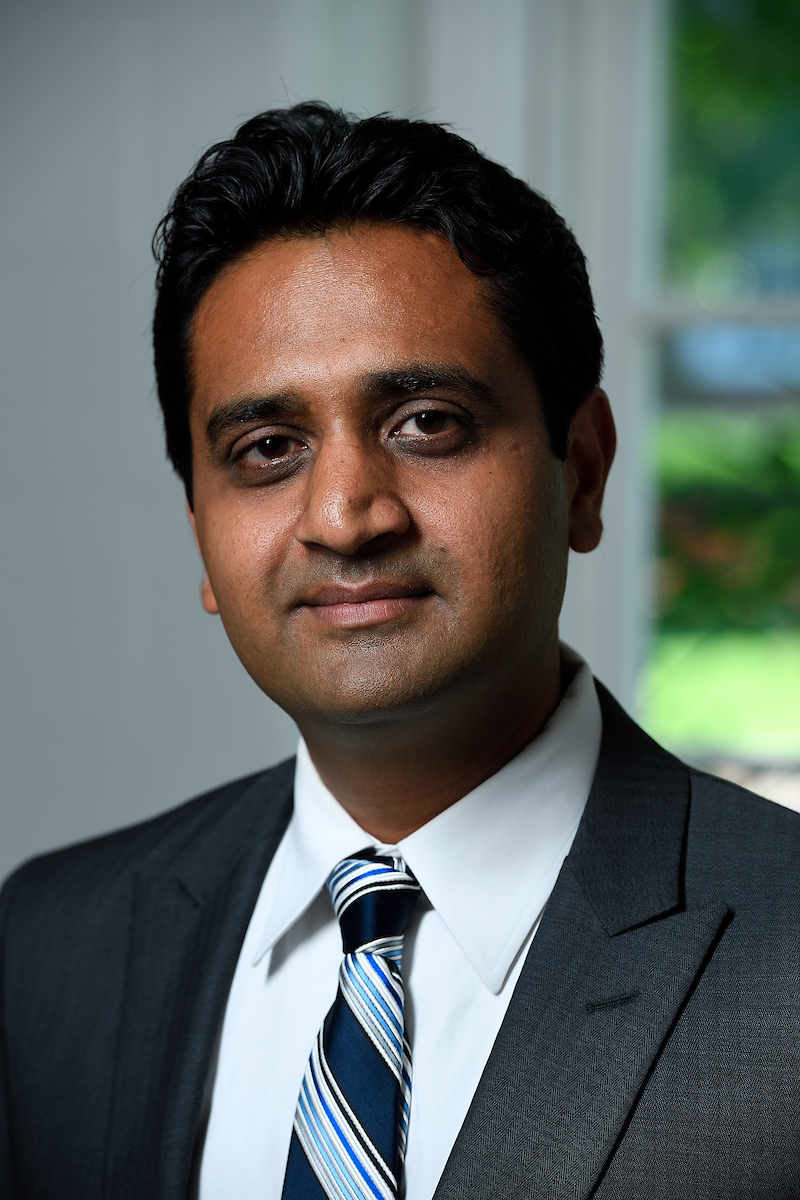}}]{Vishal M. Patel}[SM’15] is an Assistant Professor in the Department of Electrical and Computer Engineering (ECE) at Johns Hopkins University. Prior to joining Hopkins, he was an A. Walter Tyson Assistant Professor in the Department of ECE at Rutgers University and a member of the research faculty at the University of Maryland Institute for Advanced Computer Studies (UMIACS).  His current research interests include signal processing, computer vision, and pattern recognition with applications in biometrics and imaging.  He has received a number of awards including the 2016 ONR Young Investigator Award, the 2016 Jimmy Lin Award for Invention, A. Walter Tyson Assistant Professorship Award, Best Paper Award at IEEE AVSS 2017, Best Paper Award at IEEE BTAS 2015, Honorable Mention Paper Award at IAPR ICB 2018, two Best Student Paper Awards at IAPR ICPR 2018, and Best Poster Awards at BTAS 2015 and 2016. He is an Associate Editor of the IEEE Signal Processing Magazine, IEEE Biometrics Compendium, Pattern Recognition Journal, and serves on the Information Forensics and Security Technical Committee of the IEEE Signal Processing Society.  He is serving as the  Vice  President (Conferences)  of  the  IEEE  Biometrics  Council.  He is a member of Eta Kappa Nu, Pi Mu Epsilon, and Phi Beta Kappa.
\end{IEEEbiography}

%






\end{document}